%% file: main.tex
\documentclass[letterpaper, 10 pt, conference]{ieeeconf}
\IEEEoverridecommandlockouts                              

\usepackage{cite}
\usepackage{amsmath}
\usepackage{graphicx}
\usepackage{float} 
\usepackage{subfigure}
\usepackage{overpic}
\usepackage{wrapfig}
\usepackage{balance} 
\usepackage{ulem}
\usepackage{color}
\usepackage[colorlinks,linkcolor=red]{hyperref} 
\usepackage{multirow}
\usepackage{mathtools}
\usepackage{booktabs} 
\usepackage{newtxmath}
\usepackage[linesnumbered, ruled]{algorithm2e}
\graphicspath{{figures/}}

\title{\LARGE \bf
Neural Surface Reconstruction and Rendering \\for LiDAR-Visual Systems
}

\author{
    Jianheng Liu, Chunran Zheng, Yunfei Wan, Bowen Wang, Yixi Cai and Fu Zhang
}%

\begin{document}

\maketitle
\thispagestyle{empty}
\pagestyle{empty}


\begin{abstract}

This paper presents a unified surface reconstruction and rendering framework for LiDAR-visual systems, integrating Neural Radiance Fields (NeRF) and Neural Distance Fields (NDF) to recover both appearance and structural information from posed images and point clouds. 
We address the structural visible gap between NeRF and NDF by utilizing a visible-aware occupancy map to classify space into the free, occupied, visible unknown, and background regions. This classification facilitates the recovery of a complete appearance and structure of the scene.  
We unify the training of the NDF and NeRF using a spatial-varying scale SDF-to-density transformation for levels of detail for both structure and appearance.
The proposed method leverages the learned NDF for structure-aware NeRF training by an adaptive sphere tracing sampling strategy for accurate structure rendering. 
In return, NeRF further refines structural in recovering missing or fuzzy structures in the NDF. 
Extensive experiments demonstrate the superior quality and versatility of the proposed method across various scenarios.
To benefit the community, the codes will be released at \url{https://github.com/hku-mars/M2Mapping}.
\end{abstract}

\vspace{-10pt}

\section{Introduction}

3D reconstruction and novel-view synthesis (NVS) are fundamental tasks in computer vision and robotics\cite{wang2022survey,makoviychuk2021isaac}. 
The widespread availability of cameras and low-cost LiDAR sensors on robots provides rich multimodal data for 3D reconstruction and NVS.
LiDAR-visual SLAM technology are widely used in 3D reconstruction tasks like LVI-SAM \cite{lvisam2021shan}, R3LIVE \cite{lin2022r}, and FAST-LIVO \cite{zheng2024fast}. 
However, these methods can only obtain colorized raw point maps, so the map resolution, density and accuracy are fundamentally limited by the LiDAR sensors. In practical digital-twin applications \cite{straub2019replica}, high-quality watertight surface reconstructions along with photorealistic rendering are crucially important.

\begin{figure}
    \centering
    
    \subfigure{
      \centering
      \includegraphics[width=0.23\textwidth]{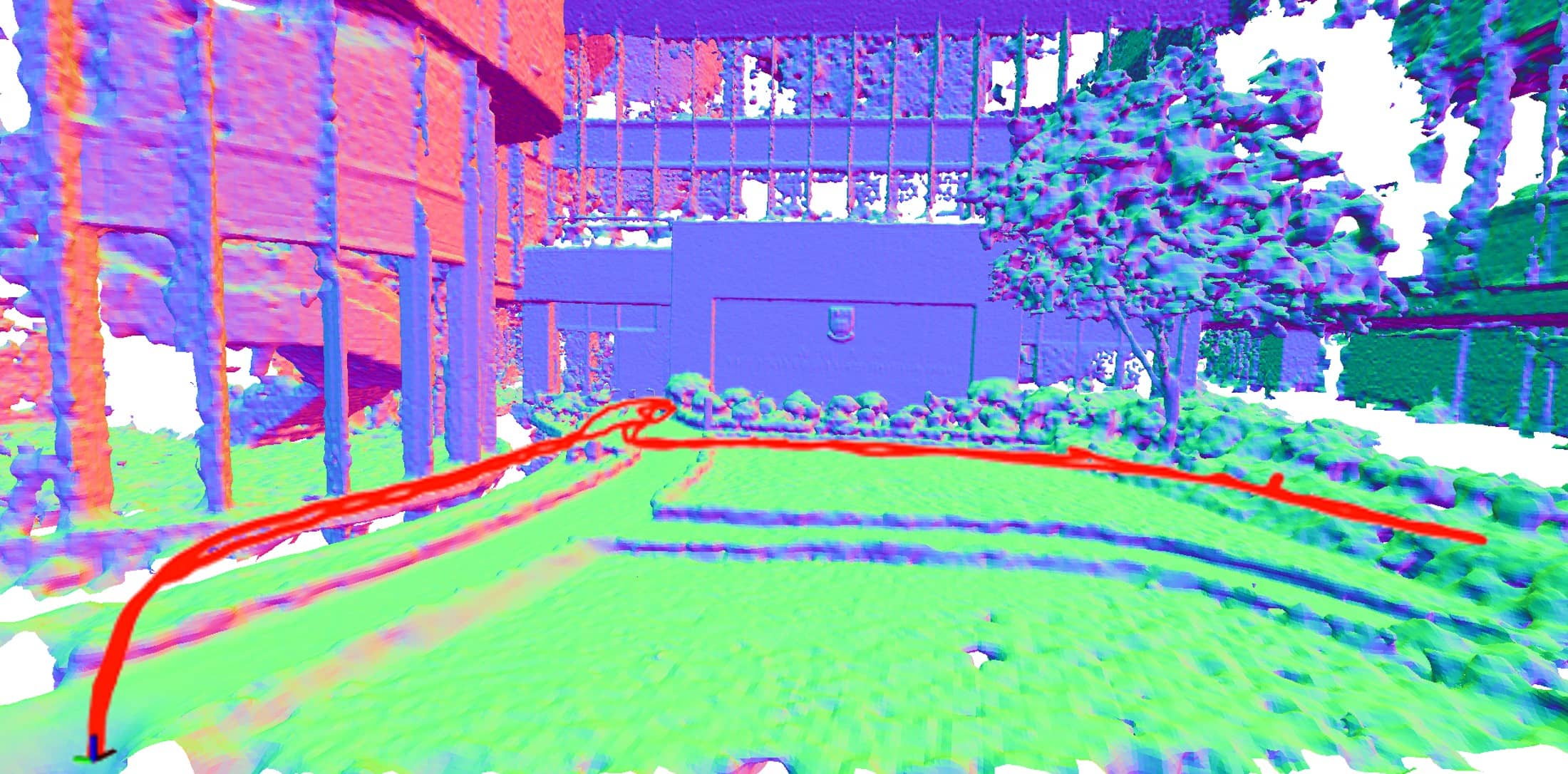}
  }
  \hspace{-12pt}
  \subfigure{
    \centering
    \includegraphics[width=0.23\textwidth]{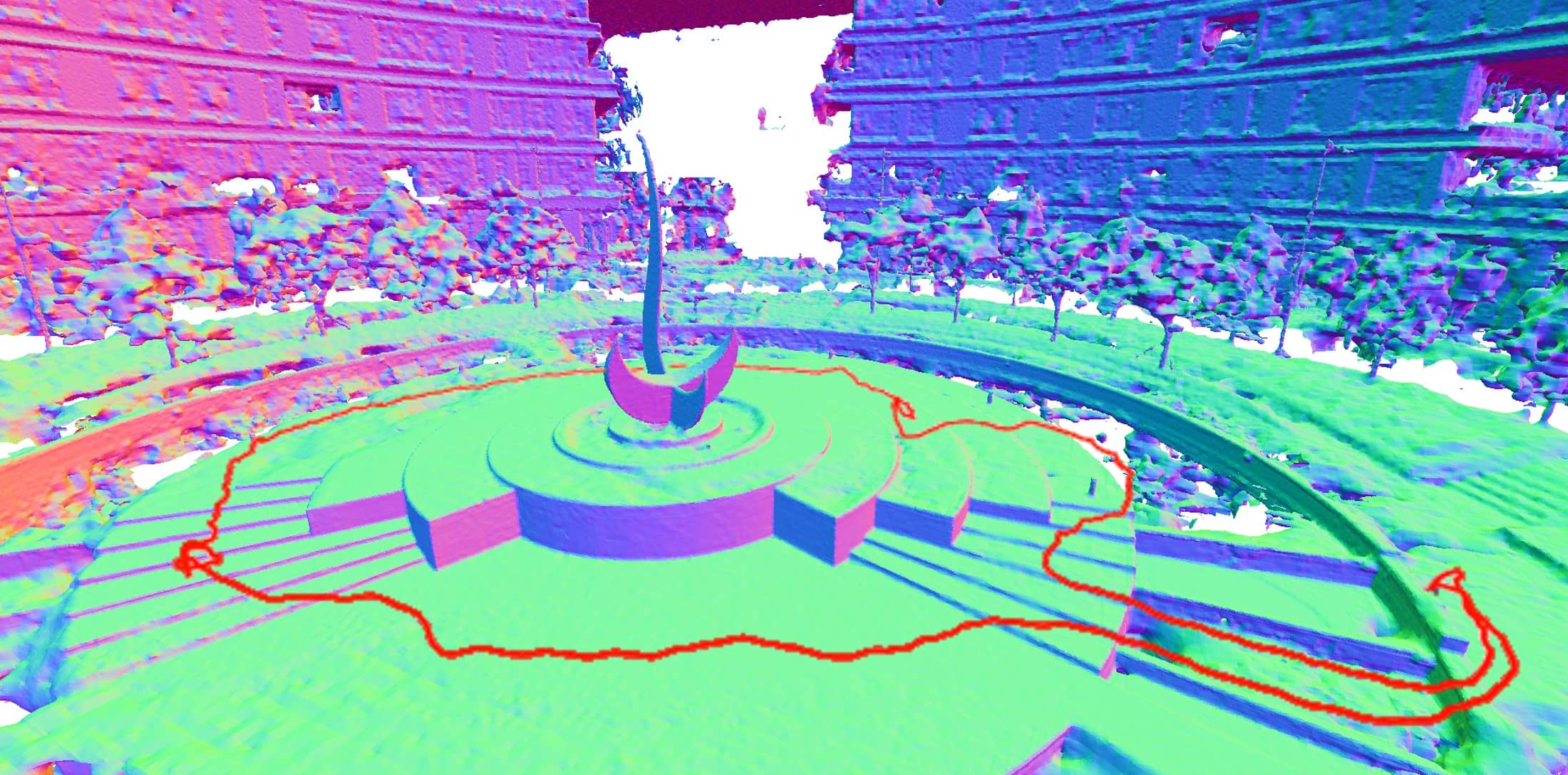}
    }

    \vspace{-10pt}

    \setcounter{subfigure}{0}
    \subfigure[Campus (FF)]{
      \centering
      \includegraphics[width=0.23\textwidth]{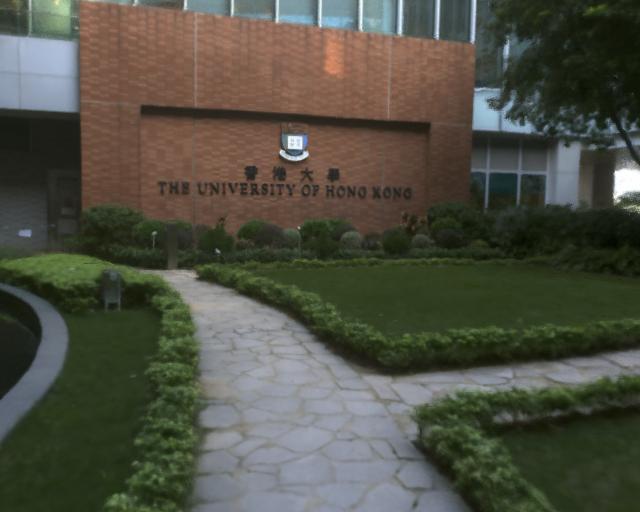}
  }
  \hspace{-12pt}
  \subfigure[Sculture (OC)]{
    \centering
    \includegraphics[width=0.23\textwidth]{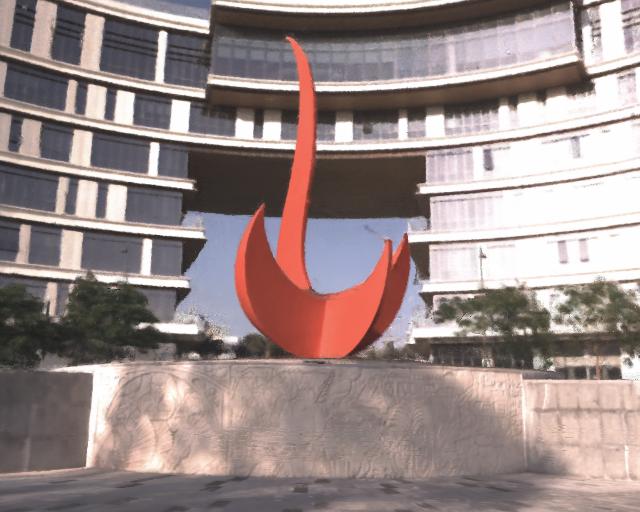}
    }

  \vspace{-6pt}

  \subfigure{
    \centering
    \includegraphics[width=0.23\textwidth]{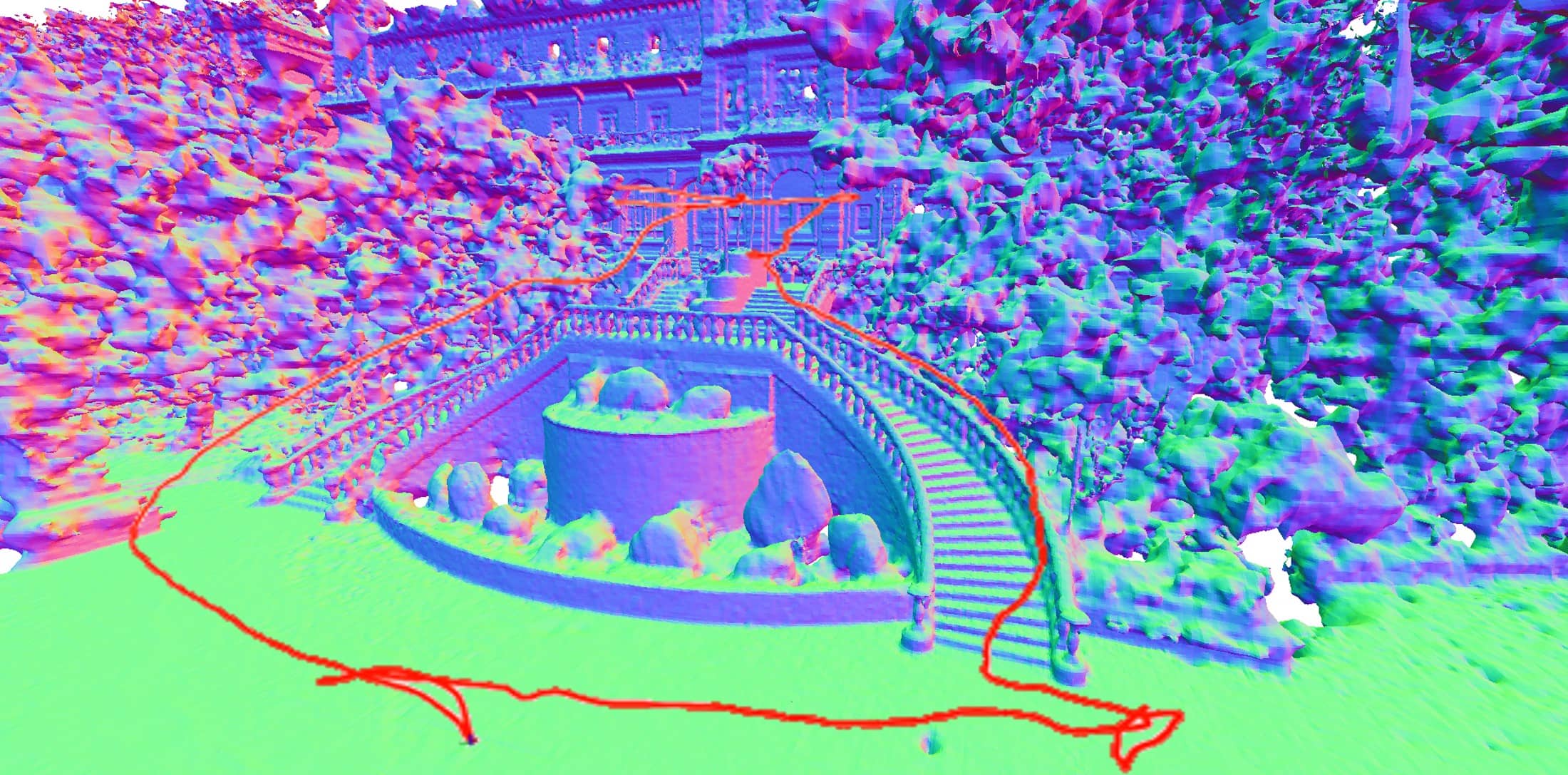}
}
\hspace{-12pt}
\subfigure{
  \centering
  \includegraphics[width=0.23\textwidth]{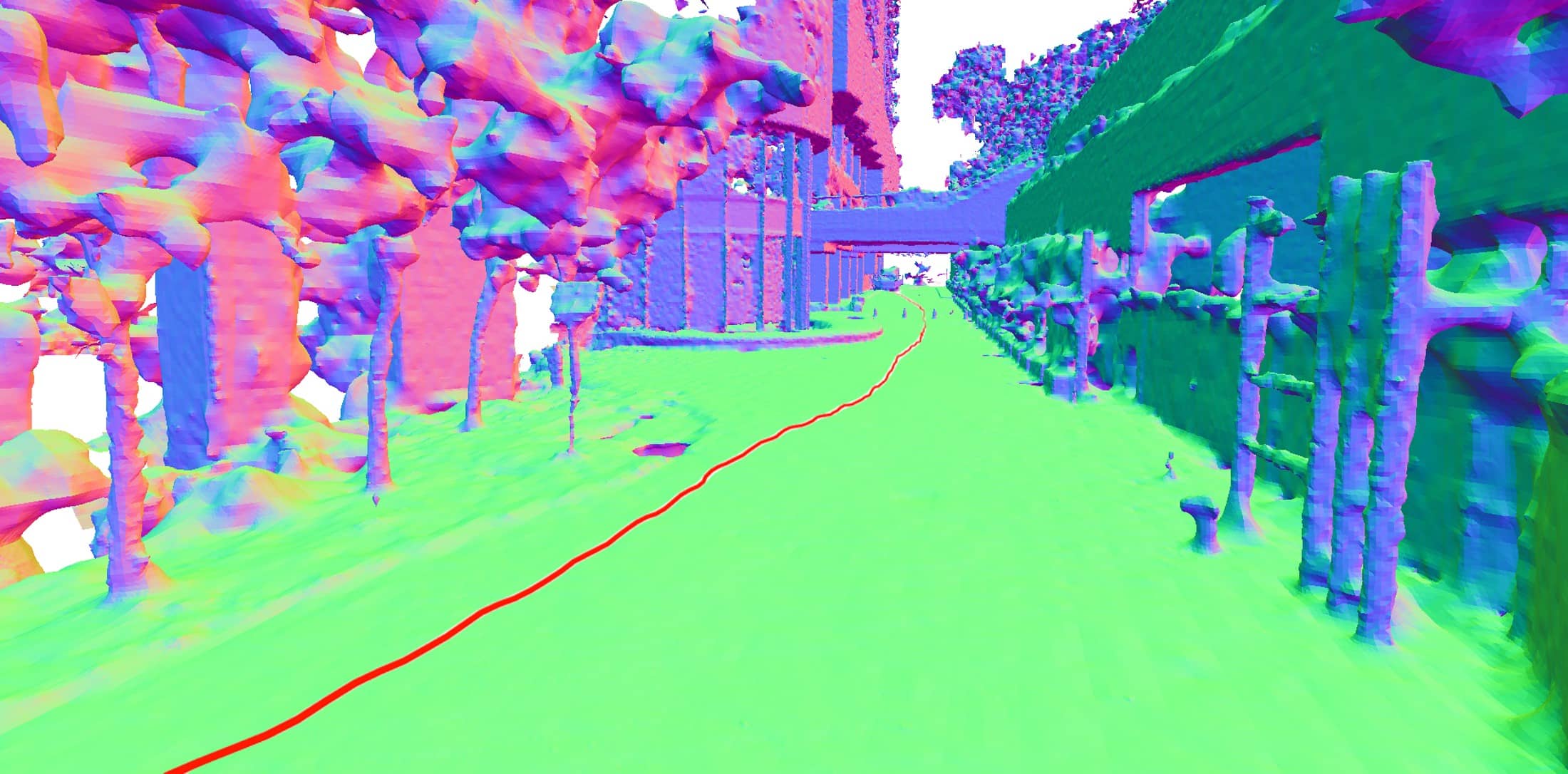}
  }

  \vspace{-10pt}

  \subfigure[Culture (Free)]{
      \centering
      \includegraphics[width=0.23\textwidth]{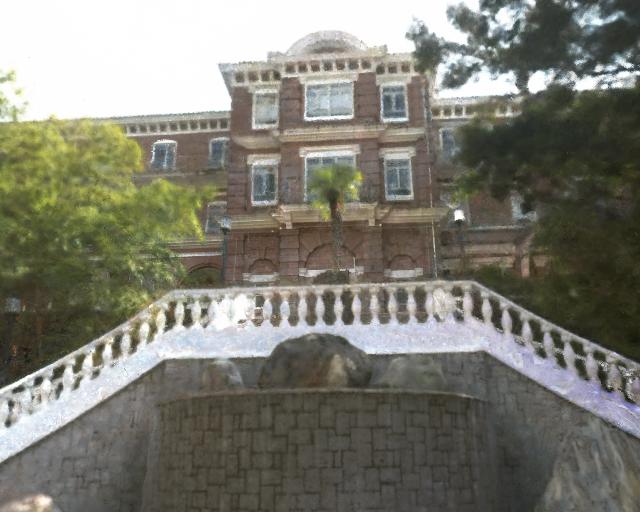}
  }  
    \hspace{-12pt}
    \subfigure[Drive (Free)]{
        \centering
        \includegraphics[width=0.23\textwidth]{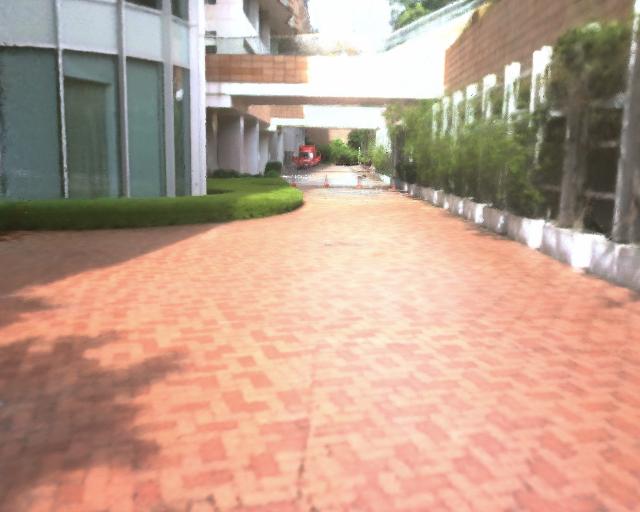}
    }
    \vspace{-2pt}
\caption{We show our surface reconstruction results (color indicates the normal direction) and rendering results collected by real-world LiDAR-visual sensor systems under different trajectories (red lines): forward-facing (FF), object-centric (OC), and free-view (Free).}
\label{fig:fast_livo_inter}
\vspace{-20pt}
\end{figure}

Recovering high-quality surface reconstructions or photorealistic rendering in the wild presents significant challenges.
Explicit surface reconstruction methods, such as direct meshing \cite{lin2023immesh}, often perform poorly for noisy or misaligned LiDAR point cloud data and lead to artifacts.
In contrast, implicit surface reconstruction methods estimate implicit functions, such as Poisson functions\cite{vizzo2021poisson} or signed distance fields (SDFs)\cite{oleynikova2017voxblox,vizzo2022vdbfusion}, which define watertight manifold surfaces at their zero-level sets, offering a more reliable approach to surface reconstruction.
The neural distance field (NDF) \cite{chibane2020neural,ortiz2022isdf,zhong2022shine} enhanced by neural networks, further improves the continuity of distance fields through gradient regularization \cite{gropp2020implicit}, demonstrating significant potential for capturing high-granularity details in surface reconstruction.

For scenes with accurate geometry, surface rendering can efficiently render solid objects using only the intersection points of pixel rays and surfaces.
However, obtaining high-quality geometry is often challenging in practice, and the rendering quality is heavily influenced by the accuracy of the underlying geometry\cite{lin2023immesh}.
In cases where the geometry is incomplete or imprecise (e.g., due to the sparse and noisy LiDAR point measurements), surface rendering methods could degrade considerably, making volumetric rendering methods more appropriate \cite{wang2023adaptive}.
Neural Radiance Fields (NeRFs) have emerged as a powerful 3D representation method for photorealistic novel-view synthesis and geometry reconstruction. 
NeRFs encode a scene's volumetric density and appearance as a function of 3D spatial coordinates and viewing direction, generating high-quality images via volume rendering. 
However, the ambiguous density field of NeRF may exhibit inaccuracies and visual artifacts when extrapolating views.
Recent works \cite{yariv2021volume,wang2021neus} demonstrate that detailed implicit surfaces can be learned via NeRF by converting NDFs to density fields.
However, this approach requires multi-view images and extensive training for each location in construction, making it less suitable for free-view trajectories, where the region of interest around the trajectory often suffers from limited multi-view images.
A low-cost LiDAR, which provides direct sample points on the surface, offers a promising solution to reduce the reliance on extensive multi-view images and enable free-view trajectories, as shown in Fig.~\ref{fig:fast_livo_inter}.

This work aims to recover both the appearance and structural information of unbounded scenes using posed images and low-cost LiDAR data from any casual trajectories, which are typical situations in practical 3D reconstruction data collection.
We propose a unified surface reconstruction and rendering framework for LiDAR-visual systems, capable of recovering both the NeRF and NDF.
This approach combines NDF and NeRF to provide better structure and appearance for scenes.
We directly apply signed ray distance in 3D space to efficiently approximate an NDF.
Photorealistic images are rendered by integrating the density field from the NDF with the radiance field.
The NDF enforces geometric consistency within NeRF and guides NeRF's samples to concentrate on surfaces using sphere tracing. 
NeRF, in turn, leverages photometric errors from rendering to refine and complete fuzzy or incomplete structures within the NDF.
The main contributions of this paper are as follows:
\begin{enumerate}
\item An open-source neural rendering framework for LiDAR-visual systems achieving high-granularity surface reconstructions and photorealistic rendering using a neural signed distance field and neural radiance field.
\item A visible-aware prior occupancy map constructed from point clouds for efficient and complete rendering of unbounded scenes.
\item A neural signed distance field with spatial-varying scale leverages sphere tracing to enable structure-aware volume rendering for neural radiance field.
\end{enumerate}
We perform extensive experiments to validate our insights and demonstrate the superiority of the proposed method across various scenarios for general types of trajectories.

\section{Related Works}

Neural Radiance Fields (NeRF) \cite{mildenhall2021nerf} is a groundbreaking method for novel-view synthesis that employs a neural network to model a scene's volumetric density and radiance.
However, NeRF encounters significant challenges when representing surfaces, as it lacks a defined density level set for surface representations.
Recent works\cite{yariv2021volume,wang2021neus,oechsle2021unisurf} employ SDF as the structure field in volume rendering, enabling surface reconstructions from multi-view images.
Recovering surfaces from images is resource-intensive while doing so from point clouds is much easier.
Surface reconstructions from point clouds are predominantly achieved using implicit representations like Poisson functions \cite{kazhdan2006poisson} or signed distance fields \cite{oleynikova2017voxblox,vizzo2022vdbfusion}.
Recent advancements in neural implicit representations\cite{ortiz2022isdf,zhong2022shine,liu2023towards} model SDFs using neural networks, providing high-granularity and complete surface representations.

Point clouds, which provide direct structural information, have been effectively used to constrain the structure field in NeRF by aligning the rendering depth with depth measurements\cite{deng2022depth}. 
RGBD cameras that provide aligned color and depth information are well practiced in neural radiance fields \cite{sucar2021imap,jiang2023h} but limited in indoor scene settings.
For outdoor scenes, LiDAR providing more accurate measurements is more favorable than RGBD cameras and extends NeRF's applicability from room-scale to urban-scale scenes \cite{rematas2022urf}.
To address the sparsity of LiDAR point clouds, existing works \cite{ziyang2023snerf,tao2024silvr} use networks to densify LiDAR into depth images for NeRF training.
While these above works still limit LiDAR usage for deep supervision in 2D image space to regularize the structure field. 
In this work, we directly apply structure constraints in 3D space by forming a continuous NDF from point clouds, and conduct a structure-aware NeRF rendering based on the NDF to avoid the limitations posed by LiDAR sparsity.

Unlike NeRF, 3D Gaussian Splatting (3DGS) \cite{kerbl20233d} explicitly stores appearance within Gaussian points initialized by structure-from-motion points. 
Its high efficiency and high-quality rendering establish it as a new benchmark in novel-view synthesis.
However, the discrete representation of 3DGS makes it difficult to represent manifold surfaces, which fails to apply direct structure regularization in 3D space.
Therefore, depth supervision \cite{GSSLAM2024} is more commonly used to regularize 3DGS' geometry.
LiDAR point clouds are well-suit for initializing 3DGS\cite{hong2024liv}, but 3DGS lacks a viable pipeline to refine the structure in reverse.
Therein, we adhere to the volume rendering of SDF pipeline \cite{yariv2021volume} to enable surface and appearance reconstructions within a unified framework and leverage both point clouds and images to supervise the structure field.

\begin{figure*}[!t]
    \centering
    \includegraphics[width=0.98\textwidth]{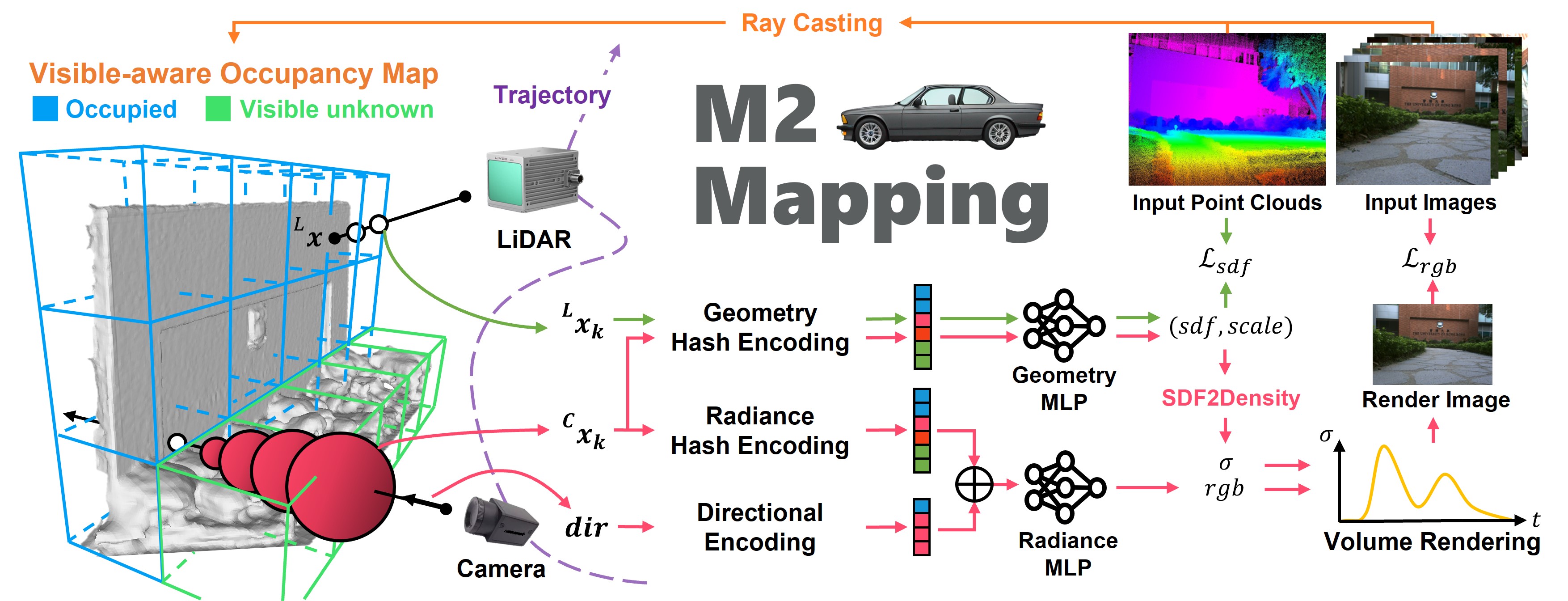}
    \caption{The overall pipeline of the proposed multimodal mapping framework, named M2Mapping. Given a series of posed images and LiDAR point clouds, we first construct the visible-aware occupancy map via ray casting. The neural distance field is trained using the ray distance value from the point cloud. The neural distance field guides the structure-aware sampling process of the neural radiance field and predicts the density of each point. The sample points and direction are encoded as features, and the MLP forwards the concatenated features to infer color. The volume rendering accumulates densities and color for novel-view synthesis. ($\bigoplus$ denotes the concatenation operation)}
    \label{fig:pipeline}
    \vspace{-16pt}
\end{figure*}

\section{Methodology}

Given a series of posed images and LiDAR point clouds, we aim to recover both the surface and the appearance of the scene. 
The overall pipeline is shown in Fig.~\ref{fig:pipeline}.
We first leverage ray casting to construct a visible-aware occupancy map from the LiDAR point cloud and images to classify the space where is need to be encoded (Sec.~\ref{sec:occ_map}).
We utilize LiDAR point clouds to supervise an NDF to recover the structure of the scene (Sec.~\ref{sec:geo_sup}).
The NDF serves as the structure field for NeRF rendering, and the NeRF reversely refines the NDF through multi-view photometric errors (Sec.~\ref{sec:app_sup}).
In this section, we elaborate on how our approach enables both the NDF and the NeRF to be aware of scenes' structures (Sec.~\ref{sec:scale} and Sec.~\ref{sec:st_sampling}) and the training strategies to deal with common issues shown in real-world applications (Sec.~\ref{sec:training}).

\subsection{Visible-aware Occupancy Map}\label{sec:occ_map}
\begin{figure}[htb]
    \centering
    \includegraphics[width=0.48\textwidth]{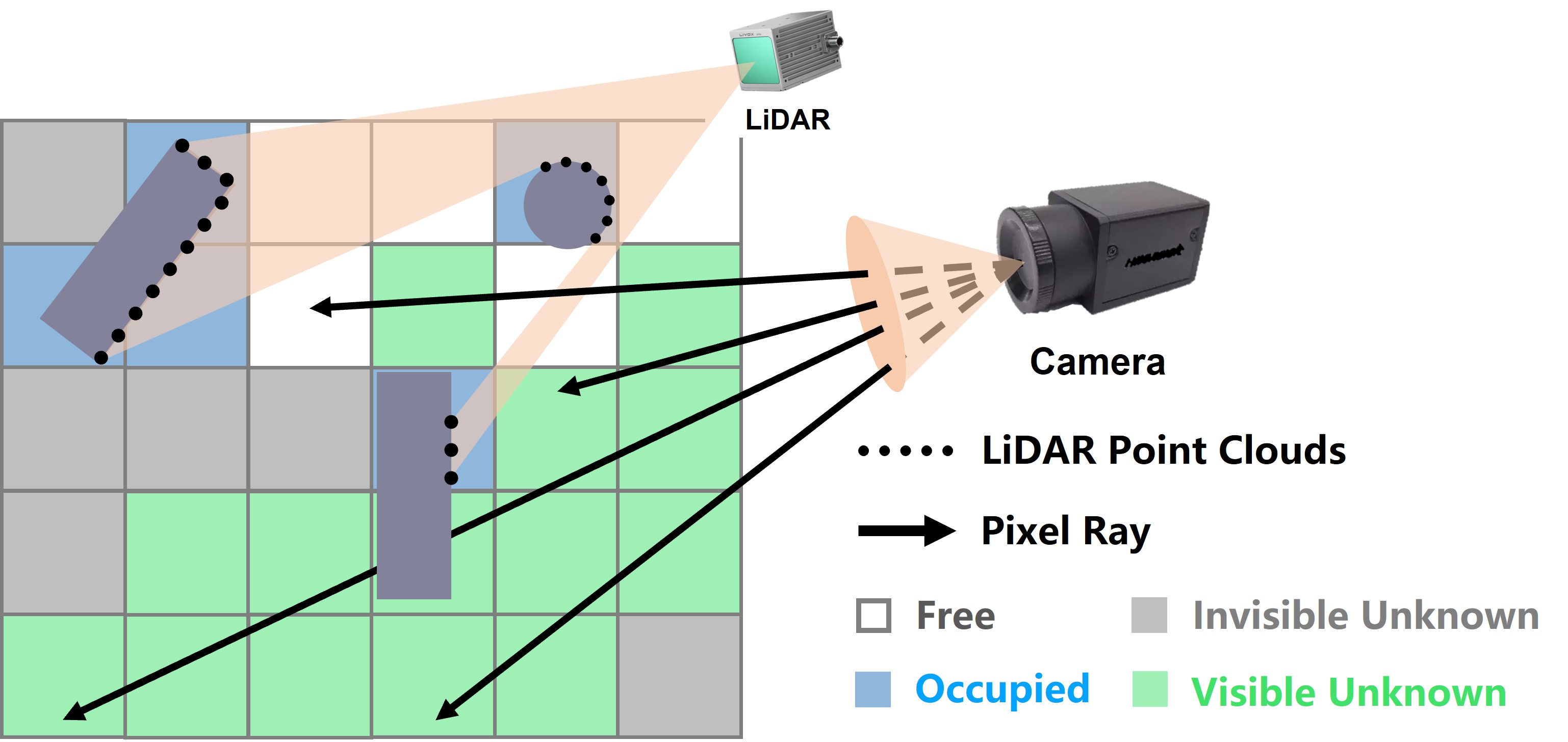}
    \caption{{Illustrations of our visible-aware occupancy map. The LiDAR observations derive a standard occupancy grid map (OGM) with states: free, occupied, and unknown. We further classify the unknown states into visible unknown and invisible unknown via images' pixel ray casting. Note that the visible unknown grids arise from unknown grids which is visible from the camera without any occupied grid in between. }}
    \label{fig:visoccmap}
    \vspace{-16pt}
\end{figure}

We would like to enable NeRF training for generalized LiDAR-visual systems.
LiDAR point clouds provide strong and accurate information of the environment, such as occupied, free, and unknown spaces. An occupancy map can be easily attained through ray casting to represent such information \cite{ren2023rog}, which can serve as an auxiliary acceleration data structure for volume rendering to skip free space.
However, due to the different field of views between LiDAR and cameras and the sparsity of the LiDAR point cloud, much space is visible to the camera but not to the LiDAR.
Only using occupied information from LiDAR for NeRF training may lead to imcomplete rendering results, therefore, we need to identify the space in need to be encoded for efficiency and complete rendering.
To address this, we adopt an occupancy map and categorize the occupancy grid states as free, occupied, visible unknown, and invisible unknown, where the visible unknown and invisible unknown grids are separated from the unknown grids based on whether they are visible in the input image. 
As illustrated in Fig.~\ref{fig:visoccmap}, we cast images pixels' rays in the built occupancy map and mark the traversed unknown grids before the first arrived occupied grids or map boundary as the visible unknown grids.
We encode the neural fields within a visible-aware occupancy map that only includes the occupied and visible unknown grids to reduce the training workload.

\subsection{Geometry supervision}\label{sec:geo_sup}

The neural signed distance field represents a scene as a distance field that maps 3D points $\boldsymbol{x} \in \mathbb{R}^3$ to signed distance value $s \in \mathbb{R}$.
We leverage hash encoding\cite{muller2022instant} and tiny multi-layer perceptions (MLPs) to form a geometry neural network to encode a scalable signed distance field: $(\hat{s},\hat{\beta}) = f_{\mathcal{S}}(\boldsymbol{x})$, where $\hat{\beta} \in \mathbb{R}$ is our proposed spatial-varying scale (Sec.~\ref{sec:scale}).
Most existing volume rendering methods for SDFs \cite{yariv2021volume,wang2021neus} train the NDF via pure images.
In contrast, we supervise the NDF using direct 3D LiDAR points.

Given a LiDAR point ${}^{L}\boldsymbol{x} = {}^{L}\boldsymbol{o} + t\ {}^{L}\boldsymbol{d}$, where ${}^{L}\boldsymbol{o}$ is the LiDAR origin and ${}^{L}\boldsymbol{d}$ is the LiDAR point's ray direction, we uniformly sample points from the LiDAR origin to the point along the ray.
For one sample point ${}^{L}\boldsymbol{x}_k = {}^{L}\boldsymbol{o} + t_k{}^{L}\boldsymbol{d}$, its ray distance value $s_k = t - t_k$ as the supervision ground truth value is transformed to the occupancy $o_k = \Phi(-s_k,\hat{\beta}_k)$, where $\Phi(v,h) = \left(1+\exp\left(-v/h\right)\right)^{-1}$ is the sigmoid function, and $\hat{\beta}_k$ is the predicted scale factor, which is indicates the standard deviation in the Logistic function controlling the sharpness of the transition.
The binary cross-entropy loss \cite{zhong2022shine} is employed to approximate a signed distance field:
\begin{equation}
    \mathcal{L}_{sdf} = -\frac{1}{N_L}\sum_{k} {o}_{i}\log(\hat{o}_{k}) + (1-{o}_{k})\log(1-\hat{o}_{k}),
    \label{eq:sdf_loss}
\end{equation}
where $\hat{o}_k = \Phi(-\hat{s}_k,\hat{\beta}_k)$ is the geometry neural network predicted occupancy value, $(\hat{s}_k,\hat{\beta}_k) = f_{\mathcal{S}}({}^{L}\boldsymbol{x}_k)$, and $N_L$ is total number of points sampled on a ray. 

\subsection{Appearance supervision} \label{sec:app_sup}



The neural radiance field maps 3D points $\boldsymbol{x} \in \mathbb{R}^3$ and viewing directions $\boldsymbol{d} \in \mathbb{R}^3$ to view-dependent colors $\boldsymbol{c} \in \mathbb{R}^3$.
We leverage hash encoding \cite{muller2022instant} for multi-resolution feature encoding and spherical harmonics encoding\cite{verbin2022ref} for directional encoding, and these two encodings are concatenated and fed into a radiance MLP to predict view-dependent color: $\hat{\boldsymbol{c}}(\boldsymbol{x},\boldsymbol{d}) = f_{\mathcal{C}}(\boldsymbol{x},\boldsymbol{d})$.

Given a pixel ray direction ${}^{C}\boldsymbol{d}$, we sample points $\{{}^{C}\boldsymbol{x}_k = {}^{C}\boldsymbol{o} + t_k{}^{C}\boldsymbol{d}\}$ from the camera origin ${}^{C}\boldsymbol{o}$ along the ray with our proposed structure-aware sampling (Sec.~\ref{sec:st_sampling}) and the volume rendering gets a pixel color $\hat{\boldsymbol{C}}$ via blending the samples' colors and densities:
\begin{equation}
    \hat{\boldsymbol{C}} =\sum_{k} \exp \left(-\sum_{j<k} \hat{\sigma}_j \delta_j\right)\left(1-\exp \left(-\hat{\sigma}_k \delta_k\right)\right) \hat{\boldsymbol{c}}_k,
    \label{eq:render}
\end{equation}
where $\delta_k = t_{k+1} - t_{k}$ is the distance between two adjacent samples along the ray, and the density $\hat{\sigma}_k$ is transformed from the signed distance value $(\hat{s}_k,\hat{\beta}_k) = f_{\mathcal{S}}(\boldsymbol{x}_k)$ to enable volume rendering of SDF\cite{wang2022hf}:
\begin{equation}
    \begin{aligned}
    \hat{\sigma}_k  &=\max\left(-\frac{\Phi\left(-\hat{s}_k,\hat{\beta}_k\right)}{\hat{\beta}_k} \frac{\partial\hat{s}_k}{\partial{}^{C}\boldsymbol{x}_k}\cdot {}^{C}\boldsymbol{d}, 0 \right),
    \end{aligned}
    \label{eq:sdf_density}
\end{equation}
where $\hat{M}_k = \frac{\partial\hat{s}_k}{\partial{}^{C}\boldsymbol{x}_k}\cdot {}^{C}\boldsymbol{d}$ is the gradient of the signed distance value along the ray direction, so-called slope.
The volume rendering formulation (Eq.~\ref{eq:render}) can be further abbreviated to $\hat{\boldsymbol{C}} =\sum_{k} T_k\alpha_k\hat{\boldsymbol{c}}_k$, where $\alpha_k = 1-\exp \left(-\hat{\sigma}_k \delta_k\right)$ is the opacity and $T_k = \prod_{j<k}\left(\exp \left(-\hat{\sigma}_j \delta_j\right)\right)$ is the transmittance.

To encode the scene's appearance, a photometric loss between the source pixel color $\boldsymbol{C}_i$ and rendered pixel color $\hat{\boldsymbol{C}}_i$ is applied:
\begin{equation}
    \mathcal{L}_{rgb} = \frac{1}{N_C}\sum_{i} \left \| \boldsymbol{C}_i - \hat{\boldsymbol{C}}_i \right \|^2,
\end{equation}
$N_C$ is total number of pixels sampled during the training.
The photometric error indirectly adjusts the neural distance field by backpropagation via volume rendering (Eq.~\ref{eq:render}) and SDF-to-density transformation (Eq.~\ref{eq:sdf_density}).

\subsubsection{Spatial-varying Scale}\label{sec:scale}

The sigmoid function $\Phi(\cdot)$ introduces a scale factor $\beta$ to control both the sharpness of the NDF and NeRF, where the smaller scale indicates the sharper transition.
In the past works, the scale factor is treated as a global scale either using truncated distance\cite{jiang2023h}, a single learnable scale\cite{wang2021neus} or calculated scale \cite{yariv2021volume}, which is not well adaptive to various granularity structures, such as tiny or fuzzy structures.
We extend the NDF to predict both the SDF values and scales at a location: $(\hat{s},\hat{\beta}) = f_{\mathcal{S}}(\boldsymbol{x})$.
The spatial-varying learnable scale turns the neural signed distance field into a stochastic signed distance field, where the scale shows the confidence of the neural networks' prediction.
For geometry supervision, a smaller scale makes the structure loss (Eq.~\ref{eq:sdf_loss}) more sensitive to noise, which enables the networks to adaptively adjust the sharpness for different granularities geometry.
For appearance supervision, a smaller scale returns a higher density (Eq.~\ref{eq:sdf_density}), which emphasizes the surface points to act like surface rendering.
In reverse, a larger scale considers more points, which is more suitable for fuzzy structures with high uncertainty in geometry.

\subsubsection{Structure-aware sampling}\label{sec:st_sampling}

The volume rendering sampling focuses on identifying the most critical areas for final rendering. 
While the photometric-oriented NeRF training often renders artifacts in the air that overfit the images, we would like to conduct an accurate structure rendering based on our learned NDF.
Sphere tracing \cite{ban2023automatic} is designed for finding the surface of a signed distance field.
This method naturally distributes more samples near a surface and fewer samples when is away.
Therefore, we propose to treat each step point in the sphere tracing as a sample thAat makes the sampling aware to structure information from the NDF for a geometry-consistent rendering, as shown in Fig.~\ref{fig:sampler_qual}.
The detailed algorithm of structure-aware sampling could be found in the supplementary materials\footnote{The supplementary materials could be found in \url{https://github.com/hku-mars/M2Mapping/blob/main/supplement.pdf}}.

\begin{figure*}[!t]
    \centering
    \setcounter{subfigure}{0}
    \subfigure[VDB-Fusion]{
    \centering
    \includegraphics[width=0.15\textwidth]{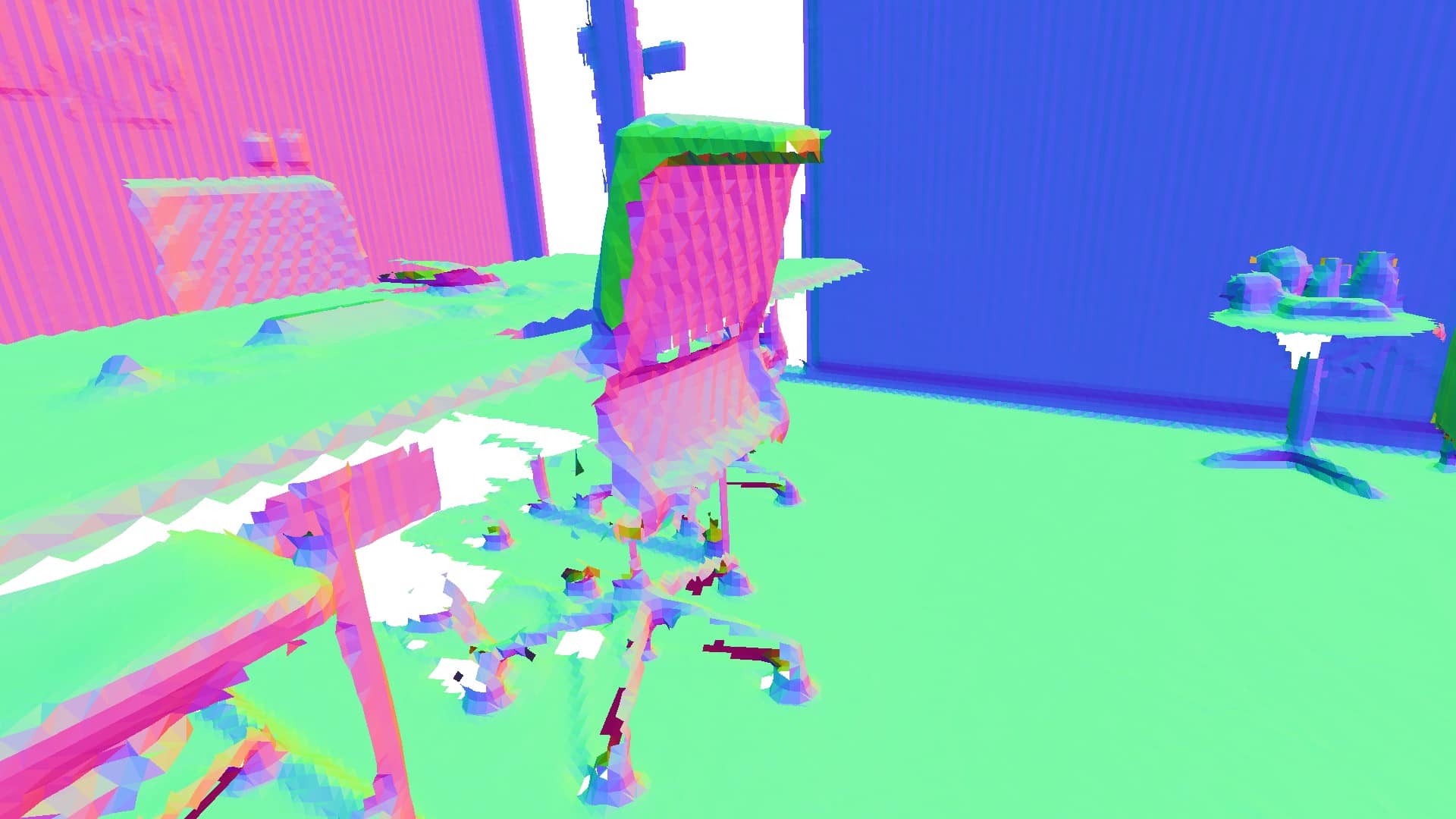}
    }
    \hspace{-11pt}
    \subfigure[iSDF]{
        \centering
        \includegraphics[width=0.15\textwidth]{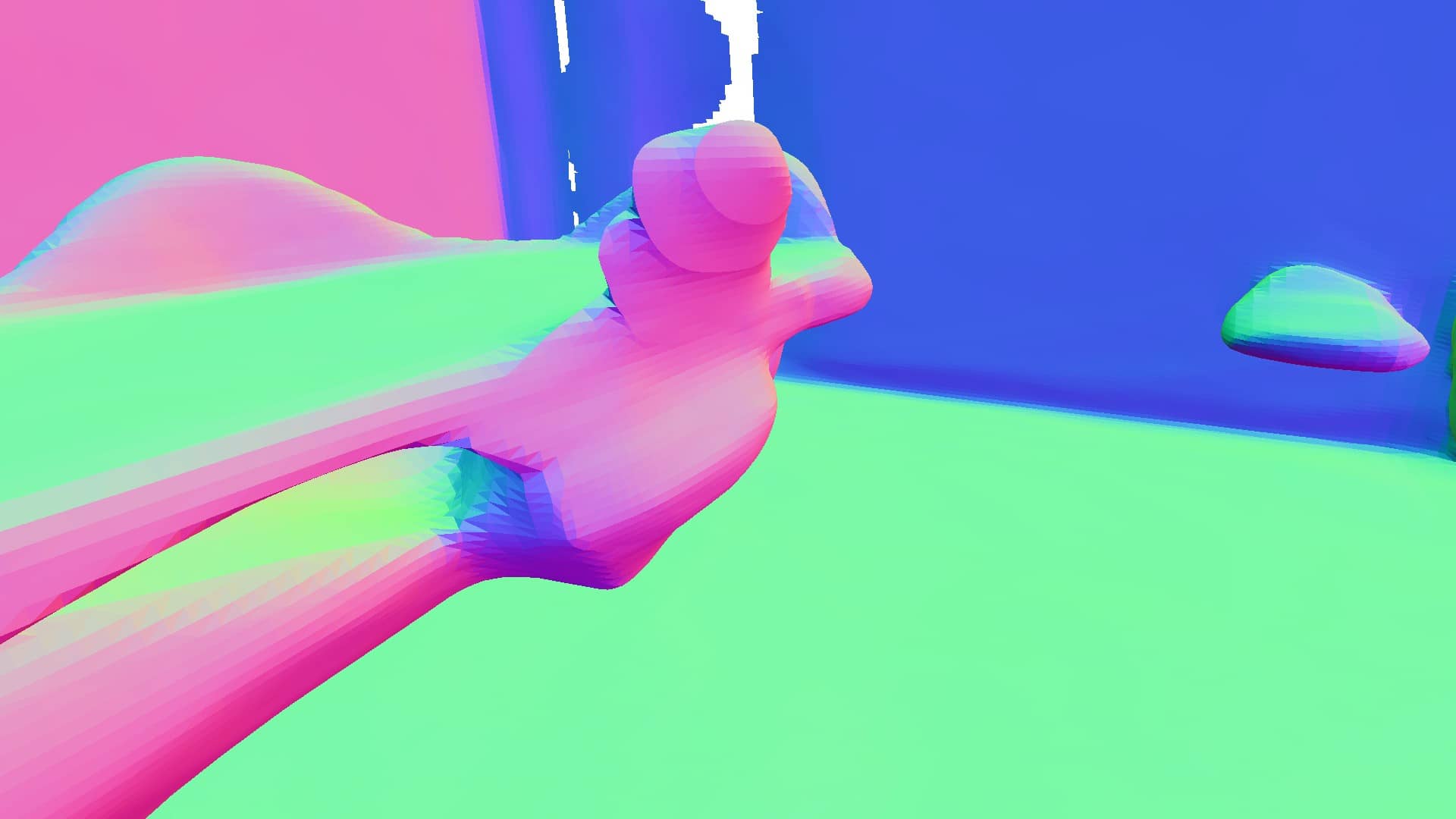}
    }
    \hspace{-11pt}
    \subfigure[SHINE-Mapping]{
    \centering
    \includegraphics[width=0.15\textwidth]{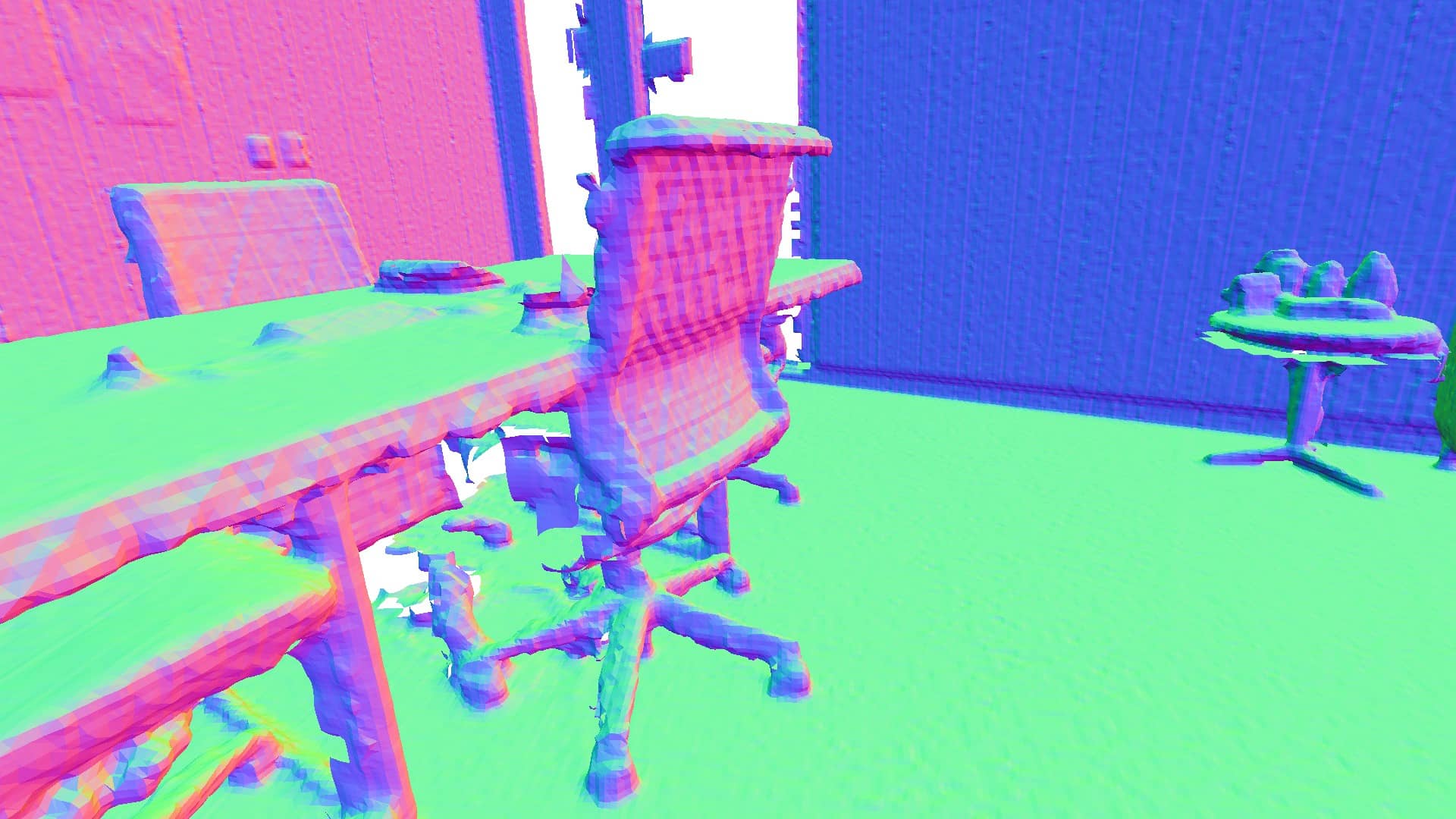}
    }
    \vline
    \setcounter{subfigure}{0}
    \subfigure[InstantNGP]{
    \centering
    \includegraphics[width=0.15\textwidth]{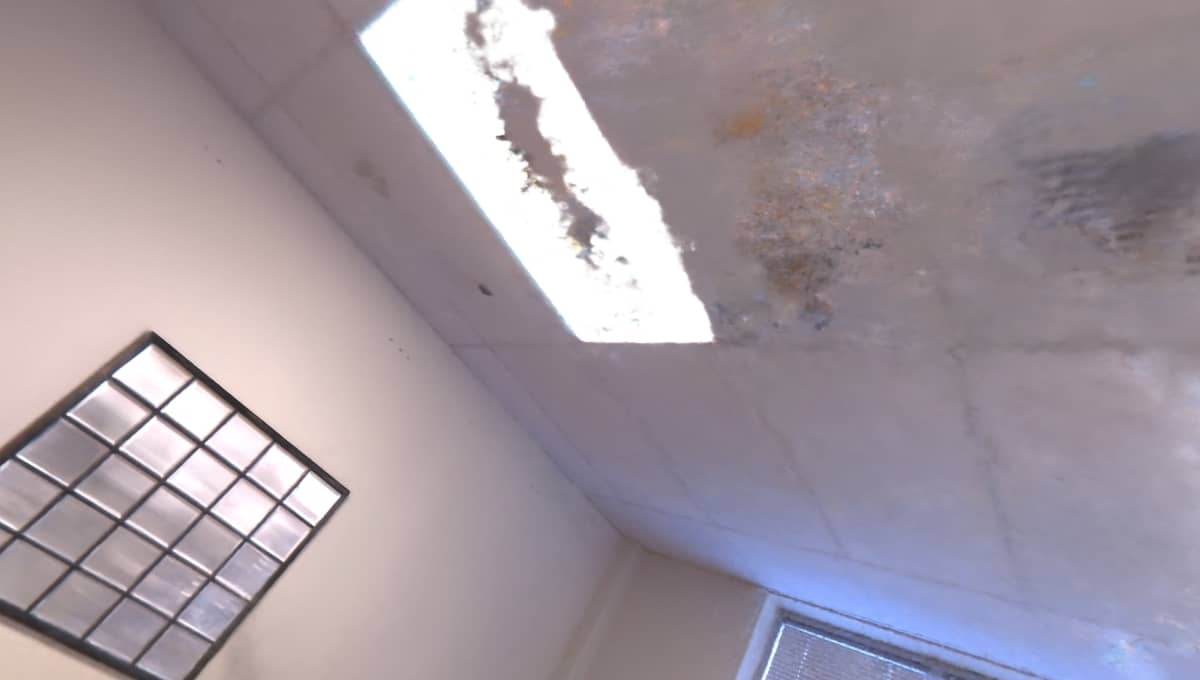}
    }
    \hspace{-11pt}
    \subfigure[3DGS$^{\dagger}$]{
    \centering
    \includegraphics[width=0.15\textwidth]{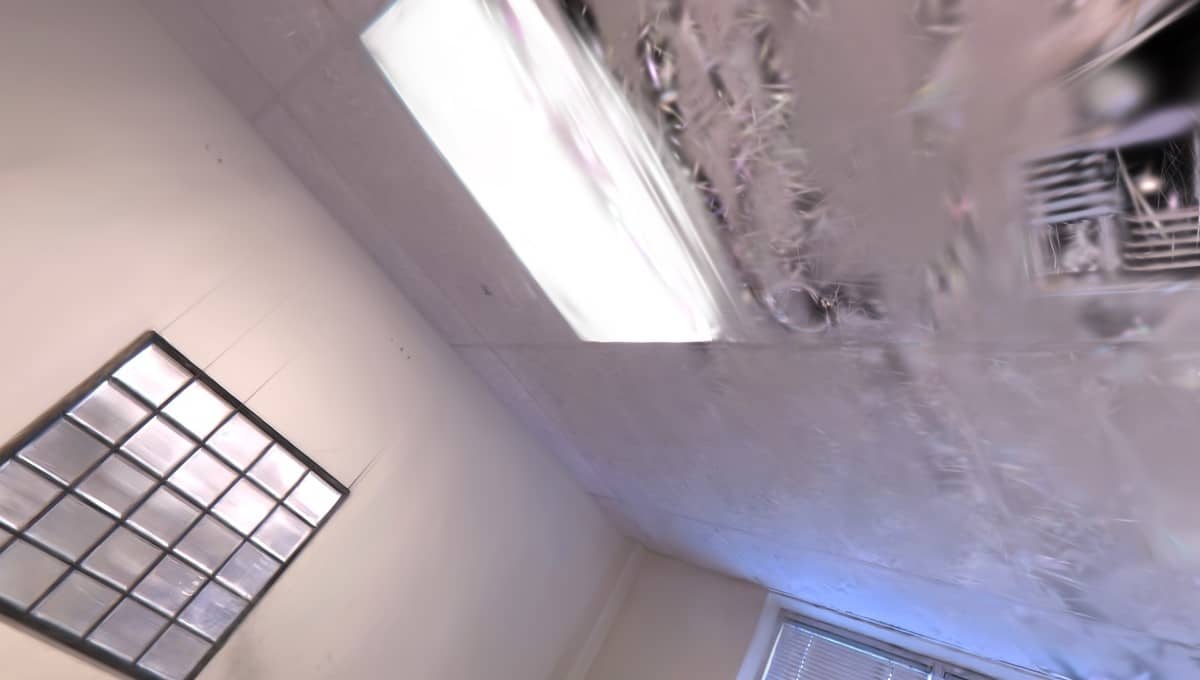}
    }
    \hspace{-11pt}
    \subfigure[MonoGS]{
    \centering
    \includegraphics[width=0.15\textwidth]{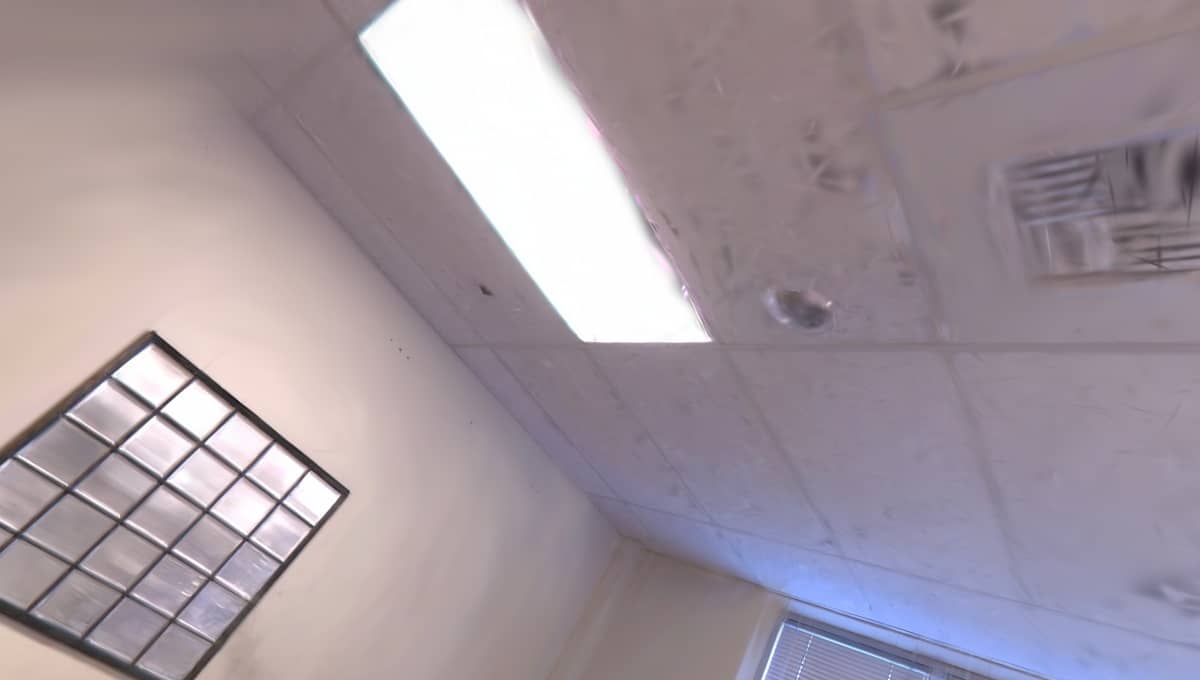}
    }

    \vspace{-6pt}
    
    \setcounter{subfigure}{2}
    \subfigure[H2Mapping]{
        \centering
        \includegraphics[width=0.15\textwidth]{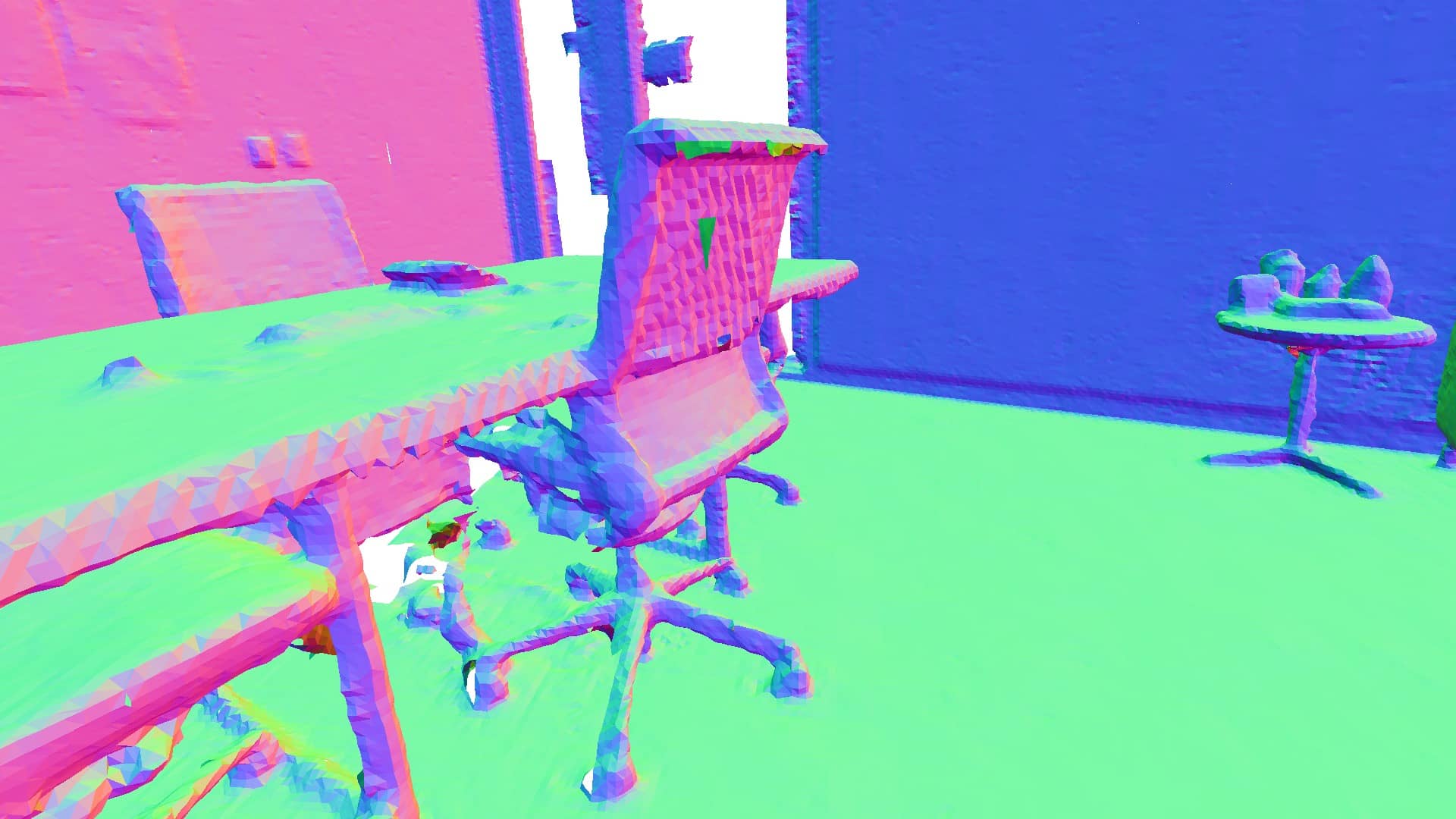}
    }
    \hspace{-11pt}
    \subfigure[Ours]{
        \centering
        \includegraphics[width=0.15\textwidth]{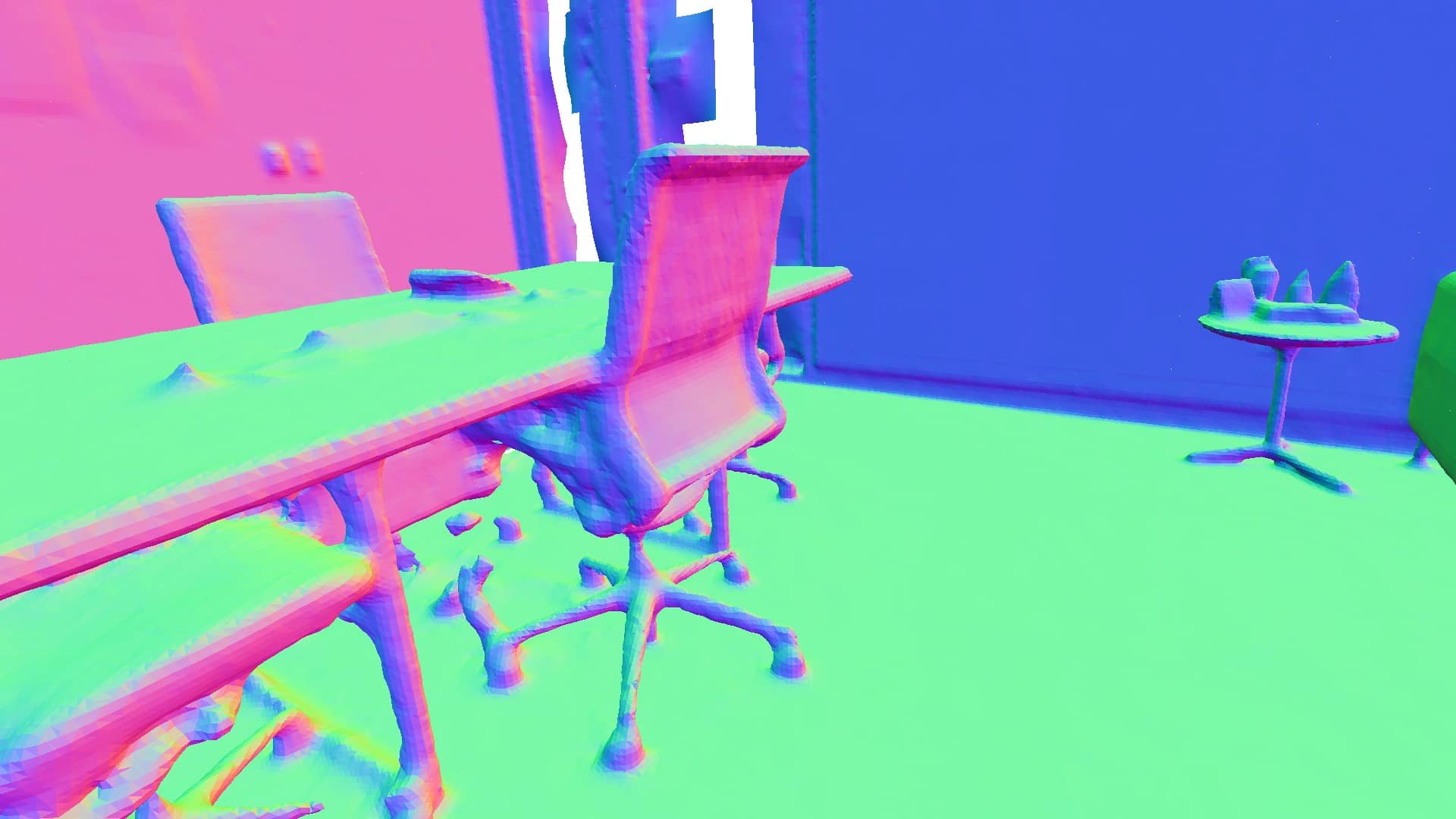}
    }
    \hspace{-11pt}
    \subfigure[Ground Truth]{
        \centering
        \includegraphics[width=0.15\textwidth]{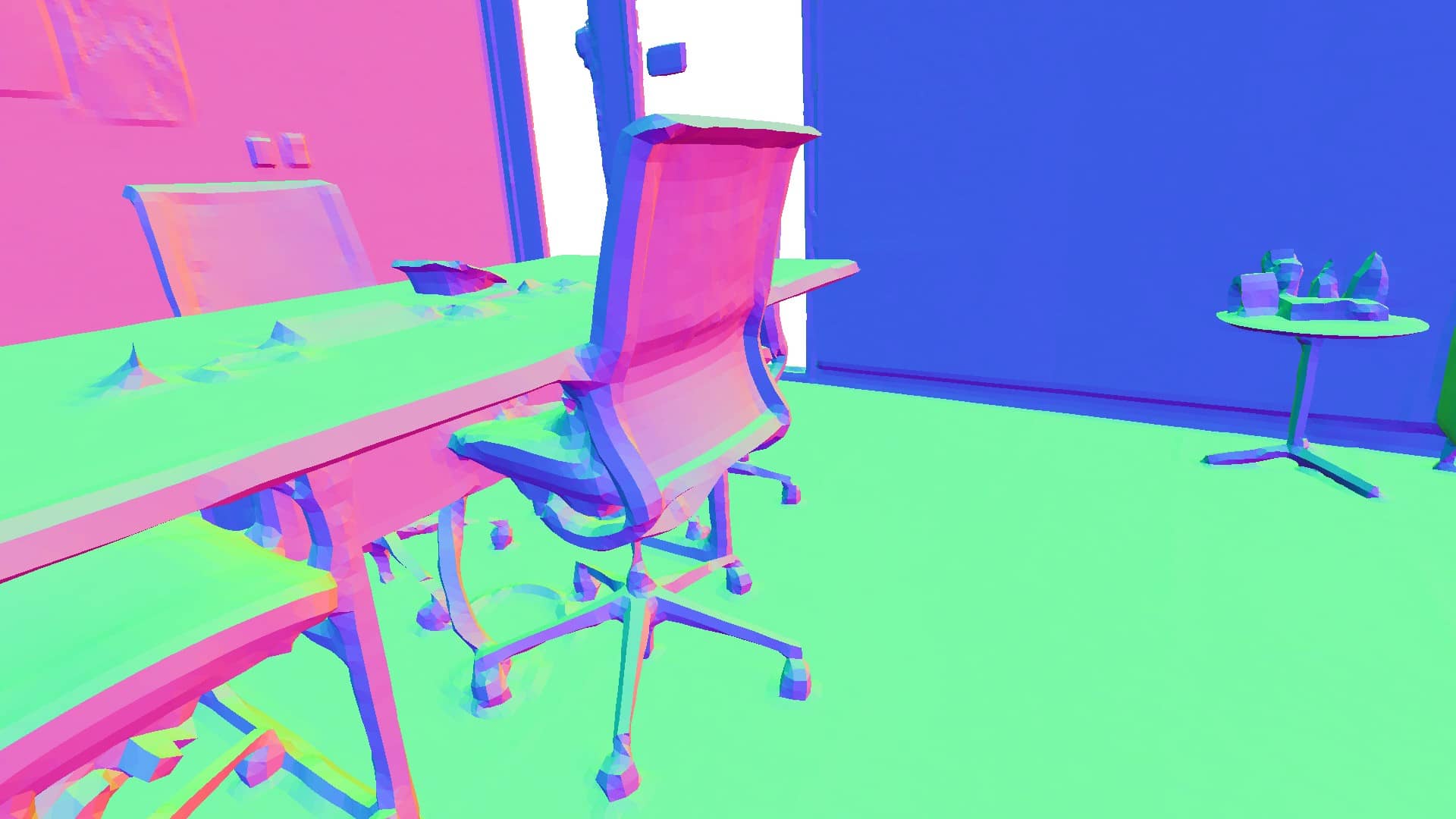}
    }
    \vline
    \setcounter{subfigure}{3}
    \subfigure[H2Mapping]{
    \centering
    \includegraphics[width=0.15\textwidth]{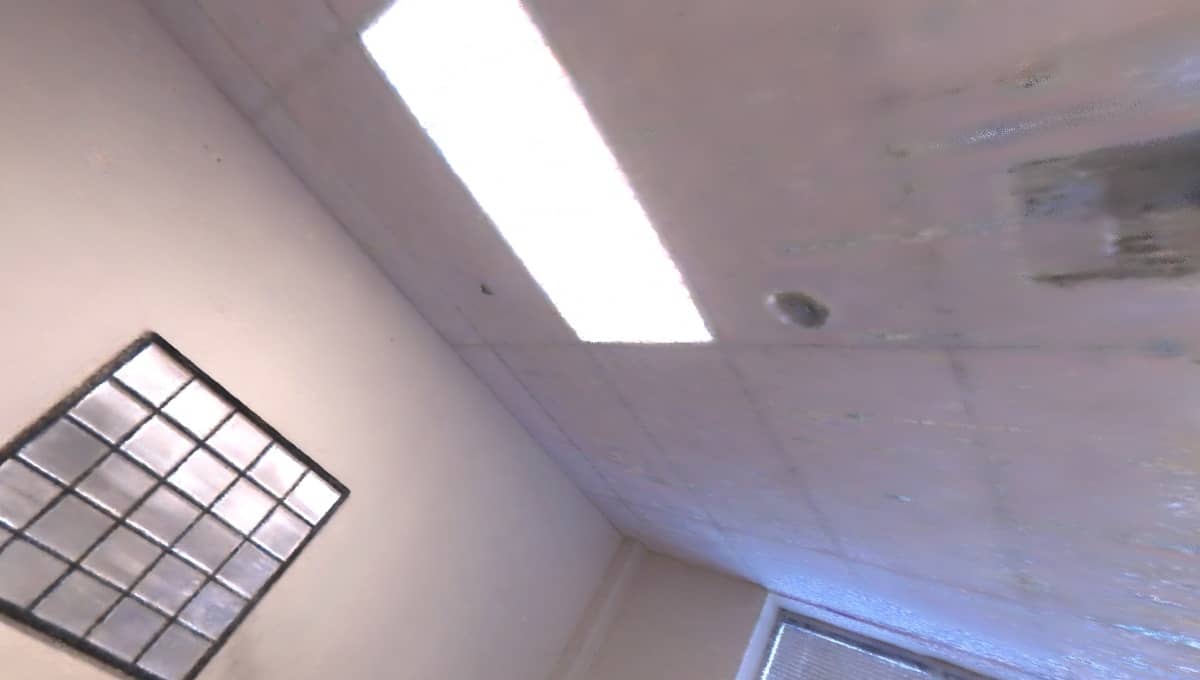}
    }
    \hspace{-11pt}
    \subfigure[Ours]{
    \centering
    \includegraphics[width=0.15\textwidth]{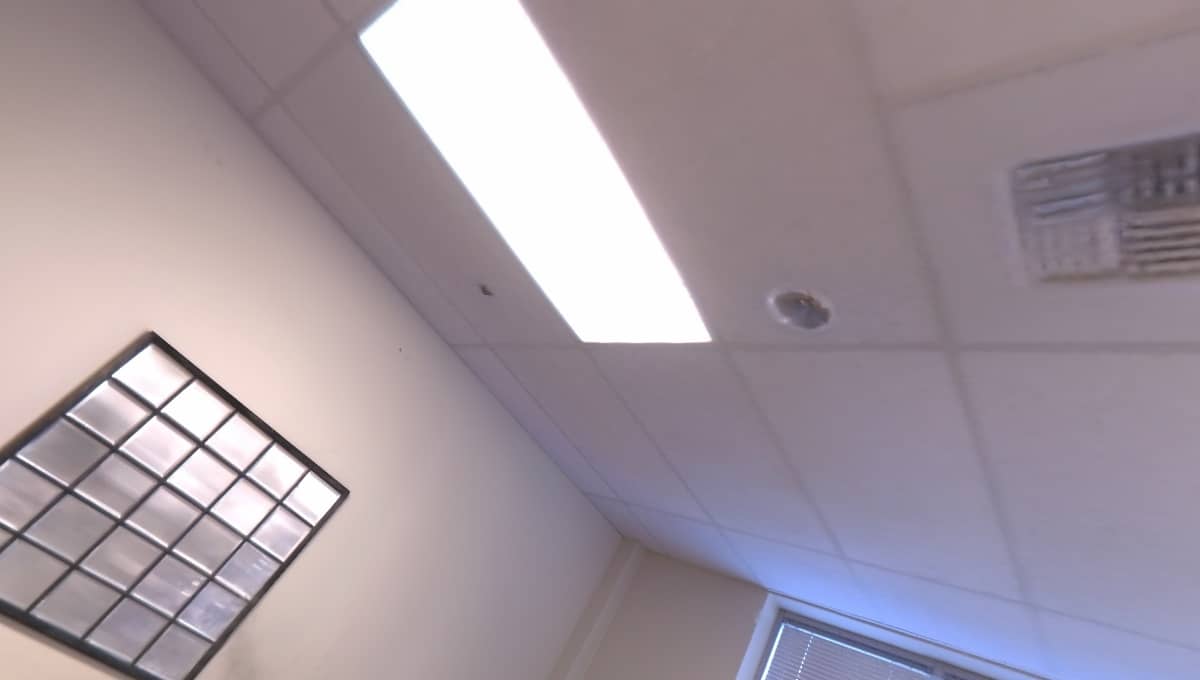}
    }
    \hspace{-11pt}
    \subfigure[Ground Truth]{
    \centering
    \includegraphics[width=0.15\textwidth]{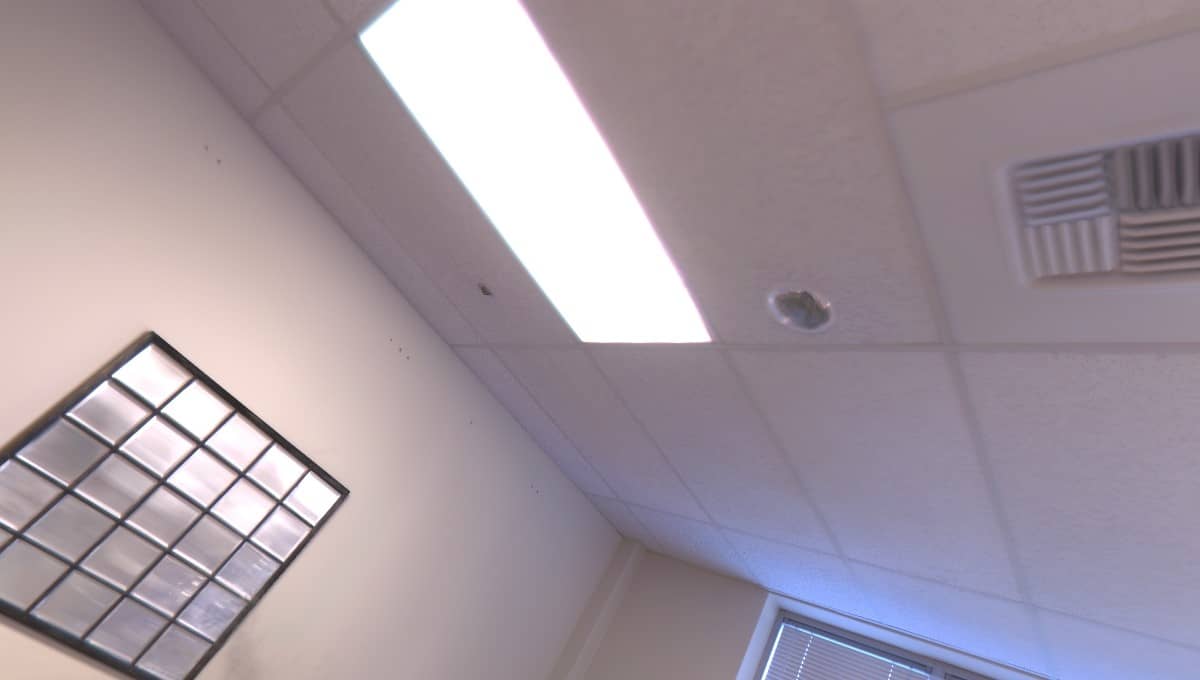}
    }
    
    \caption{
    (Left) Reconstructed mesh on the Replica dataset's office-2 and colors indicate the direction of the surface normal. 
    (Right) Extrapolation rendering results on the Replica dataset's room-0.
    Our method can capture more precise geometric details in slim objects and retain both the structure and texture in extrapolation rendering.}
    \label{fig:replica_qual}
\end{figure*}

\subsection{Training}\label{sec:training}


We further regularize the NDF with Eikonal\cite{gropp2020implicit} and curvature \cite{yang2023steik} loss, and the details can be found in supplementary materials.
The overall training loss is defined as follows:
\begin{equation}
    \begin{aligned}
    \mathcal{L} = \mathcal{L}_{sdf} +\lambda_{rgb} \mathcal{L}_{rgb} + \lambda_{eik} \mathcal{L}_{eik} + \lambda_{curv} \mathcal{L}_{curv},
    \end{aligned}
\end{equation}
where $\lambda_{rgb}, \lambda_{eik}$ and $\lambda_{curv}$ are the weights for the photometric, Eikonal, and curvature losses, respectively.

\section{Experiments}


In this section, we conduct extensive experiments to evaluate our proposed system.
Due to space limitations, the implementation details and part of the experiments including ablation study are illustrated in the supplementary materials.

\subsection{Experiments Settings}
\subsubsection{Baselines}
We compare our method with other state-of-the-art approaches.
For structure-oriented methods, we include the voxel-based method VDBFusion \cite{vizzo2022vdbfusion}, and the neural-network-based methods iSDF \cite{ortiz2022isdf} and SHINE-Mapping \cite{zhong2022shine}.
For appearance-oriented methods, we include the NeRF-based method InstantNGP \cite{muller2022instant}.
For depth-aided appearance-oriented methods, we include 3D Gaussian Splatting \cite{kerbl20233d} initialized with dense point clouds, denoted as 3DGS$^{\dagger}$, and RGBD-based methods H2Mapping \cite{jiang2023h} and MonoGS \cite{GSSLAM2024}.

\subsubsection{Metrics}
To evaluate the quality of the geometry, we recover the implicit model to a triangular mesh using marching cubes \cite{lorensen1987marching} and calculate the Chamfer Distance (C-L1, cm) and F-Score ($<2$ cm, \%) with the ground truth mesh.
For rendering quality evaluation, we use the Structural Similarity Index (SSIM), Peak Signal-to-Noise Ratio (PSNR), and Learned Perceptual Image Patch Similarity (LPIPS) to compare the rendered image with the ground truth image.

\begin{table*}[!th]
    \centering
    \caption{Quantitative surface reconstruction, interpolation(I) and extrapolation(E) rendering results on the Replica dataset. }
    \label{tab:replica_quan_inter}
    \resizebox{1.0\textwidth}{!}{
    \begin{tabular}{ccccccccccc}
        \toprule
        \textbf{Metrics} & \textbf{Methods} & \textbf{Office-0} & \textbf{Office-1} & \textbf{Office-2} & \textbf{Office-3} & \textbf{Office-4} & \textbf{Room-0} & \textbf{Room-1} & \textbf{Room-2} & \textbf{Avg.} \\
        \midrule

        \multirow{5}*{C-L1[cm]$\downarrow$}  

        & {VDBFusion} 
        & 0.618 & 0.595 & 0.627 & 0.685 & 0.633 & 0.637 & 0.558 & 0.656 & 0.626
        \\
        & {iSDF} 
        & 2.286 & 4.032 & 3.144 & 2.352 & 2.120 & 1.760 & 1.712 & 2.604 & 2.501
        \\
        & {SHINE-Mapping}
        & 0.753 & 0.663 & 0.851 & 0.965 & 0.770 & 0.650 & 0.844 & 0.826 & 0.790
        \\
        & {H2Mapping} 
        & \underline{0.557} & \underline{0.529} & \underline{0.585} & \underline{0.644} & \underline{0.616} & \underline{0.568} & \underline{0.523} & \underline{0.616} & \underline{0.580}
        \\
        & {Ours} 
        & \textbf{0.494} & \textbf{0.476} & \textbf{0.501} & \textbf{0.517} & \textbf{0.531} & \textbf{0.486} & \textbf{0.455} & \textbf{0.532} & \textbf{0.499}
        \\
        \midrule

        \multirow{5}*{F-Score[\%]$\uparrow$}  

        & {VDBFusion} 
        & 97.129 & 97.107 & 97.007 & 96.221 & 96.993 & 97.871 & 98.311 & 96.117 & 97.095
        \\
        & {iSDF} 
        & 75.829 & 55.429 & 71.232 & 77.274 & 81.281 & 78.253 & 84.801 & 76.815 & 75.114
        \\
        & {SHINE-Mapping}
        & 94.397 & 95.606 & 92.150 & 87.855 & 94.427 & 97.010 & 91.307 & 92.601 & 93.169
        \\
        & {H2Mapping} 
        & \underline{98.430} & \underline{98.279} & \underline{97.935} & \underline{97.077} & \underline{97.391} & \underline{98.850} & \underline{99.003} & \underline{96.904} & \underline{97.983}
        \\
        & {Ours}
        & \textbf{98.970} & \textbf{98.771} & \textbf{98.657} & \textbf{98.415} & \textbf{98.158} & \textbf{99.238} & \textbf{99.496} & \textbf{97.677} & \textbf{98.673}
        \\
        \midrule
        \midrule

        \multirow{5}*{SSIM(I)$\uparrow$}  

        & {InstantNGP} 
        & 0.981 &  {0.980} & 0.963 & 0.960 & 0.966 & 0.964 & 0.964 & 0.967 & 0.968
        \\

        & {3DGS$^{\dagger}$} 
        & \textbf{0.987} & {0.980} & \textbf{0.980} & \textbf{0.976} & \textbf{0.980} & \textbf{0.977} & \textbf{0.982} & \textbf{0.980} & \textbf{0.980}
        \\

        & {H2Mapping} 
        & 0.963 & 0.960 & 0.931 & 0.929 & 0.941 & 0.914 & 0.929 & 0.935 & 0.938
        \\
        
        & {MonoGS} 
        & \underline{0.985} & \underline{{0.982}} & \underline{0.973} & \underline{0.970} & {0.976} & \underline{0.971} & \underline{0.976} & \underline{0.977}  & {0.975}
        \\
        
        & {Ours}
        & \underline{0.985} & \textbf{0.983} & \underline{0.973} & \underline{0.970} & \underline{0.977} & 0.968 & 0.975 & \underline{0.977} & \underline{0.976}
        \\
        \midrule

        \multirow{5}*{PSNR(I)$\uparrow$}  

        & {InstantNGP} 
        & 43.538 & \underline{{44.340}} & \underline{38.273} & {37.603} & {39.772} & {37.926} & 38.859 & 39.568 & 39.984
        \\

        & {3DGS$^{\dagger}$} 
        & \underline{{43.932}} & 43.237 & \underline{{39.182}} & \underline{{38.564}} & \textbf{41.366} & \textbf{39.081} & \textbf{41.288} & \textbf{41.431} & \underline{{41.010}}
        \\

        & {H2Mapping} 
        & 38.307 & 38.705 & 32.748 & 33.021 & 34.308 & 31.660 & 33.466 & 32.809 & 34.378
        \\

        & {MonoGS} 
        & {43.648} & {43.690} & {37.695} & {37.539} & {40.224} & {37.779} & {39.563} & {40.134} & {40.034}
        \\

        & {Ours}
        & \textbf{44.369} & \textbf{44.935} & \textbf{39.652} & \textbf{38.874} & \underline{41.318} & \underline{38.541} & \underline{40.775} & \underline{40.705} & \textbf{41.146}
        \\
        \midrule

        \multirow{5}*{LPIPS(I)$\downarrow$}  

        & {InstantNGP} 
        & 0.046 & 0.069 & 0.096 & 0.087 & 0.089 & 0.079 & 0.094 & 0.094 & 0.082
        \\
        
        & {3DGS$^{\dagger}$} 
        & \underline{0.040} & 0.075 & {0.069} & {0.068} & {0.065} & \underline{0.060} & {0.057} & 0.075 & {0.064}
        \\

        & {H2Mapping} 
        & 0.109 & 0.163 & 0.186 & 0.170 & 0.154 & 0.177 & 0.167 & 0.177 & 0.163
        \\
        
        & {MonoGS} 
        & \textbf{0.031}  &  \underline{{0.048}} & \underline{{0.061}} & \underline{{0.057}} & \underline{{0.052}} & \textbf{0.055} & \textbf{0.050} & \underline{{0.056}} & \underline{{0.051}}
        \\

        & {Ours}
        & \textbf{0.031} & \textbf{0.044} & \textbf{0.055} & \textbf{0.050} & \textbf{0.050} & 0.062 & \underline{0.056} & \textbf{0.053} & \textbf{0.050}
        \\

        \midrule
        \midrule
        
      \multirow{5}*{SSIM(E)$\uparrow$}  
      
      & {InstantNGP} 
      & 0.972 & \underline{0.961} & 0.934 & 0.938 &  \underline{0.952} & \underline{0.918} & 0.936 & 0.941 & 0.944
      \\

      & {3DGS$^{\dagger}$} 
      & 0.936 & 0.897 & 0.924 & 0.917 & 0.925 & 0.881 & 0.915 & 0.919 & 0.914
      \\

      & {H2Mapping} 
      & 0.957 & 0.955 & 0.932 & 0.925 & 0.937 & 0.866 & 0.916 & 0.917 & 0.926
      \\
      
      & {MonoGS} 
      & \underline{0.974} & \underline{0.961} & \underline{0.945} & \underline{0.942} &  {0.950} & {0.912}& \underline{0.942} & \underline{0.946} & \underline{0.947}
      \\

      & {Ours}
      & \textbf{0.980} & \textbf{0.976} & \textbf{0.960} & \textbf{0.964} & \textbf{0.970} & \textbf{0.955} & \textbf{0.963} & \textbf{0.965} & \textbf{0.967}
      \\
      \midrule

      \multirow{5}*{PSNR(E)$\uparrow$}  

      & {InstantNGP} 
      & \underline{39.874} & \underline{39.120} & 31.274 & \underline{32.135} & \underline{34.458} & \underline{32.587} & \underline{33.024} & \underline{32.266} & \underline{34.341}
      \\
      
      & {3DGS$^{\dagger}$} 
      & 31.220 & 29.959 & 27.411 & 26.442 & 28.324 & 27.541 & 28.429 & 27.139 & 28.307
      \\

      & {H2Mapping} 
      & 36.740 & 37.841 & \underline{31.427} & 31.144 & 31.988 & 28.815 & 31.192 & 30.603 & 32.468
      \\

      & {MonoGS} 
      & 39.197 & 38.818 & 29.740 & 29.664 & 31.632 & 29.949 & 31.126 & 30.621 & 32.593
      \\

      & {Ours}
      & \textbf{41.965} & \textbf{42.215} & \textbf{35.056} & \textbf{37.465} & \textbf{38.667} & \textbf{36.427} & \textbf{37.294} & \textbf{36.722} & \textbf{38.226}
      \\
      \midrule

      \multirow{5}*{LPIPS(E)$\downarrow$}  
      
      & {InstantNGP} 
      & 0.085 & 0.117 & 0.176  &  0.185 & 0.149 & 0.180 & 0.162 & 0.156 & 0.151
      \\
      & {3DGS$^{\dagger}$} 
      & 0.117 & 0.177 & 0.167 & 0.181 & 0.155 & 0.204 & 0.170 & 0.174 & 0.168
      \\

      & {H2Mapping} 
      & 0.119 & 0.163 & 0.195 & 0.221 & 0.166 & 0.250 & 0.188 & 0.199 & 0.188
      \\

      & {MonoGS} 
      & \underline{0.061} &  \underline{0.104} &  \underline{0.136} & \underline{0.137} & \underline{0.108} & \underline{0.170} & \underline{0.123} & \underline{0.117} & \underline{0.120}
      \\

      & {Ours}
      & \textbf{0.046} & \textbf{0.062} & \textbf{0.084} & \textbf{0.081} & \textbf{0.066} & \textbf{0.114} & \textbf{0.084} & \textbf{0.087} & \textbf{0.078}
      \\
        \bottomrule
    \end{tabular}
    }
\end{table*}

\begin{figure*}[!t]
    \centering
    
    \subfigure{
        \centering
        \includegraphics[width=0.3405\textwidth]{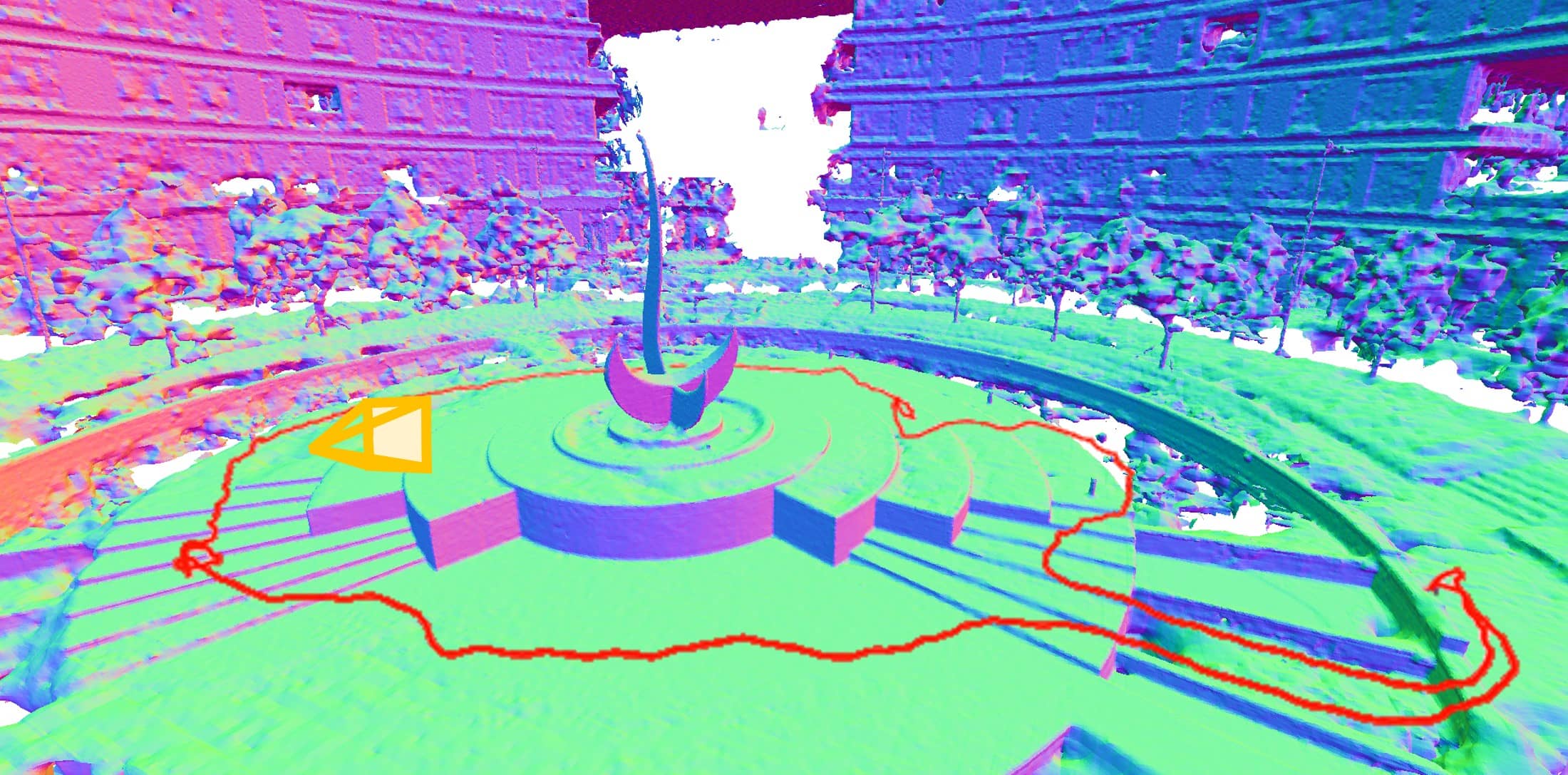}
    }
    \hspace{-12pt}
    \subfigure{
    \centering
    \includegraphics[width=0.21\textwidth]{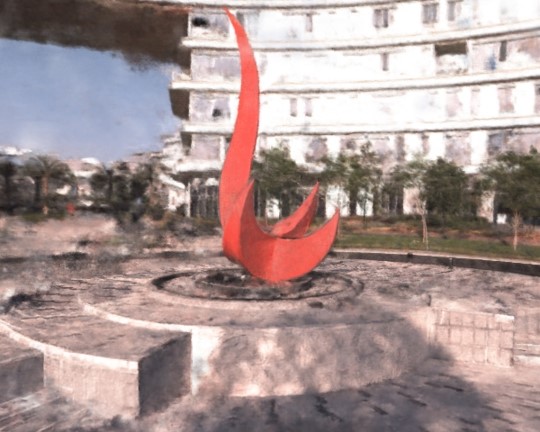}
    }
    \hspace{-12pt}
    \subfigure{
    \centering
    \includegraphics[width=0.21\textwidth]{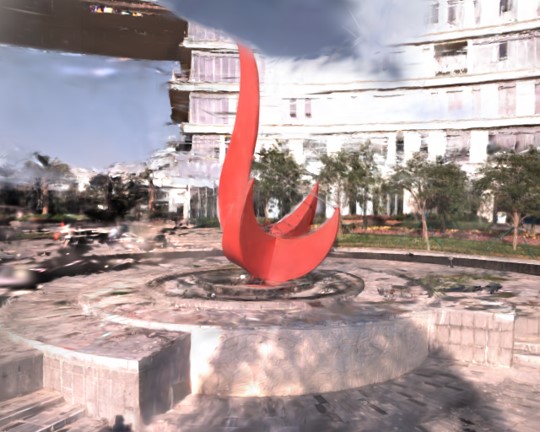}
    }
    \hspace{-12pt}
    \subfigure{
    \centering
    \includegraphics[width=0.21\textwidth]{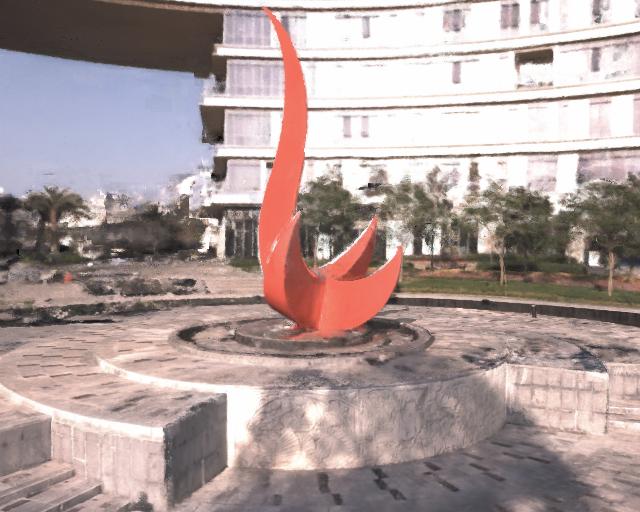}
    }

    \vspace{-8pt}
    
    \setcounter{subfigure}{0}
    \subfigure[Mesh (Ours)]{
        \centering
        \includegraphics[width=0.3405\textwidth]{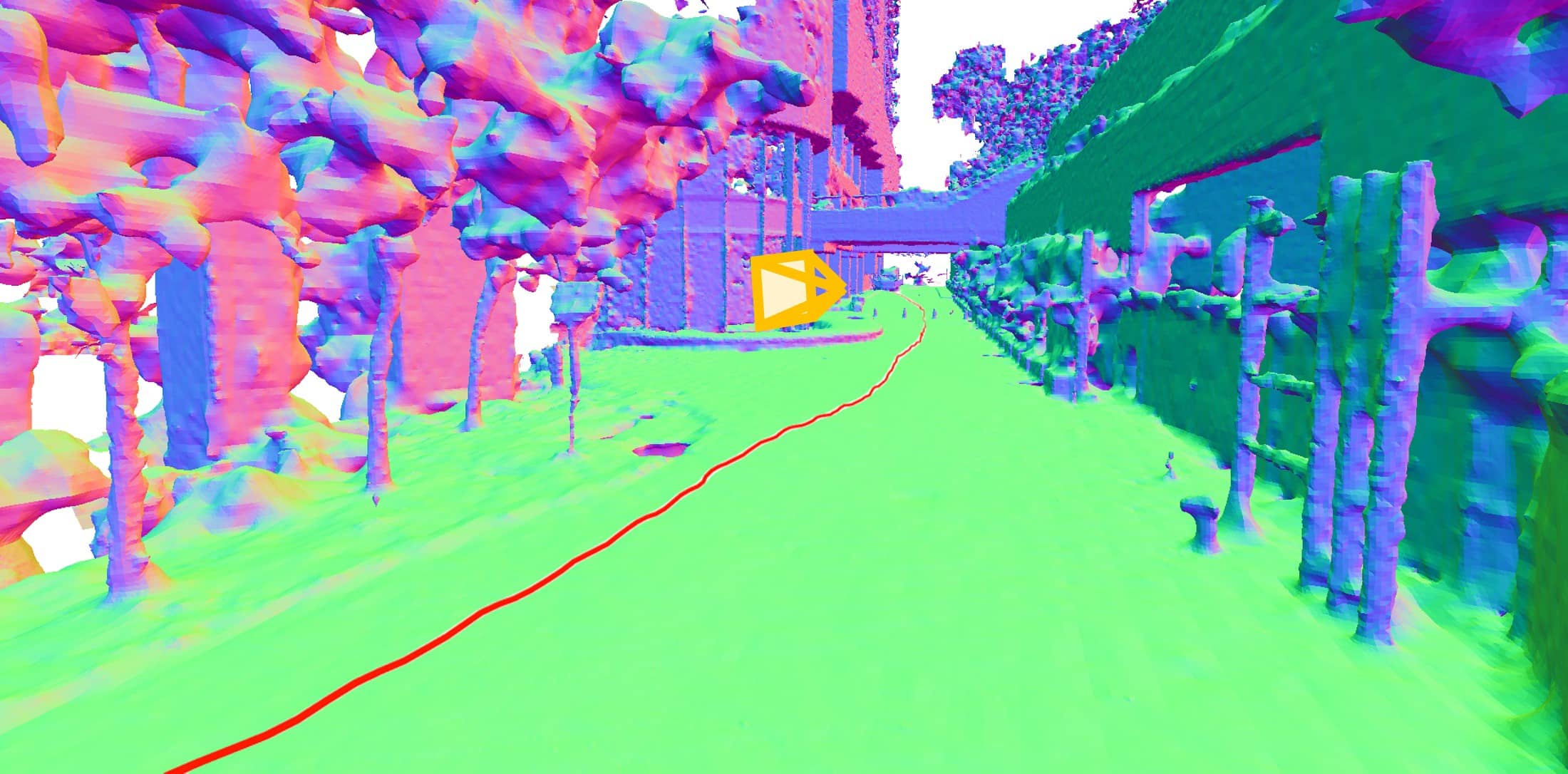}
    }
    \hspace{-12pt}
    \subfigure[InstantNGP]{
    \centering
    \includegraphics[width=0.21\textwidth]{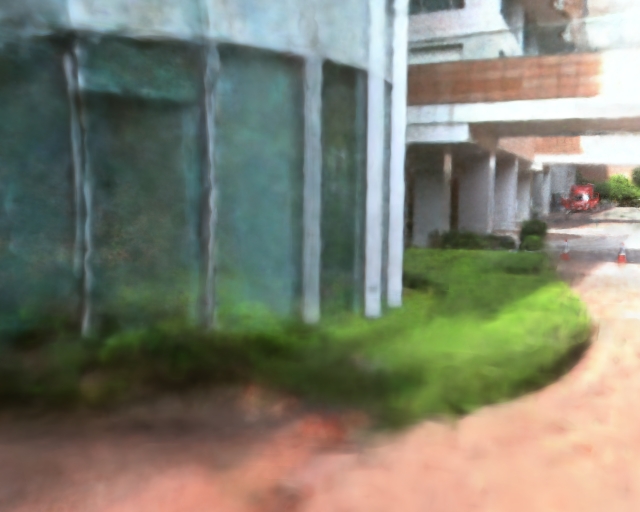}
    }
    \hspace{-12pt}
    \subfigure[3DGS$^{\dagger}$]{
    \centering
    \includegraphics[width=0.21\textwidth]{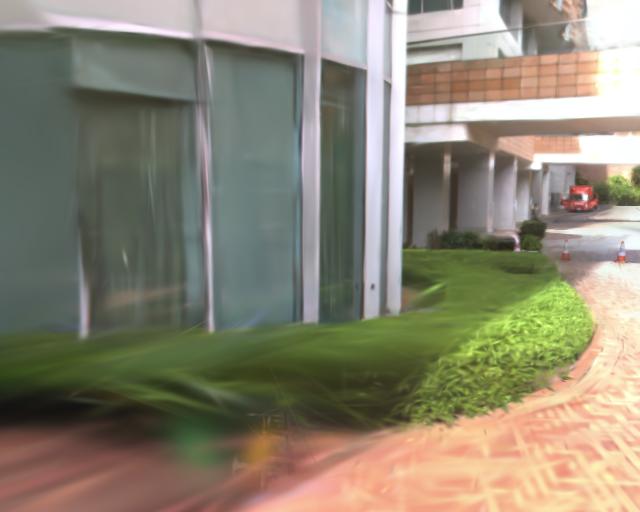}
    }
    \hspace{-12pt}
    \subfigure[Ours]{
    \centering
    \includegraphics[width=0.21\textwidth]{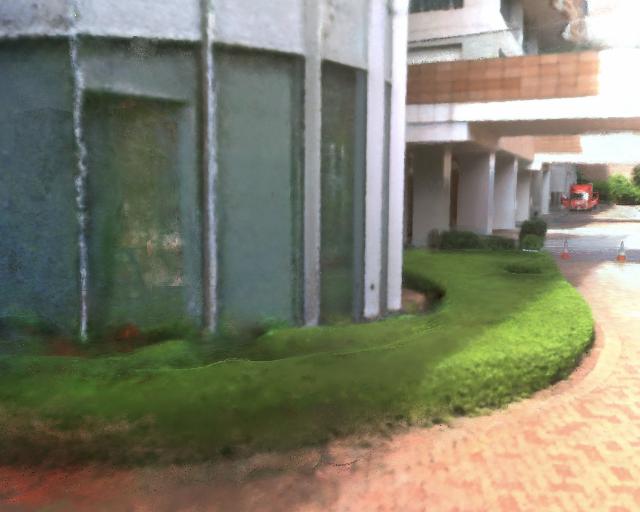}
    }
    
    \caption{
    The qualitative results of FAST-LIVO2 datasets (scenes from top to bottom are Sculpture and Drive). We show our surface reconstruction results on the left, and the red line indicates the training path and the orange cameras indicates the extrapolation views for the right side's rendering results.}
    \label{fig:fast_livo2_qual}
    \vspace{-8pt}
\end{figure*}

\subsection{Replica Dataset}

The Replica dataset \cite{straub2019replica} provides room-scale indoor simulation RGBD sensor data.
We emphasize the issues of extrapolation rendering consistency by uniformly sampling positions and orientations in each scene to generate the extrapolation dataset from Replica.
The compared surface reconstruction, interpolation, and extrapolation rendering results are shown in Tab.~\ref{tab:replica_quan_inter}.
As illustrated in Fig.~\ref{fig:replica_qual} (Left), we capture more precise geometric details in slim objects while maintaining smoothness.
For novel-view synthesis, the proposed structure-aware rendering method yields competitive results in interpolation rendering and achieves a significant improvement in extrapolation rendering compared to other methods.
As depicted in Fig.~\ref{fig:replica_qual} (Right), our method retains both the structure and texture in views of the ceiling, which were not seen with a vertical view in the training dataset.

\begin{table}[h]
    \centering
    \caption{Quantitative results on the FAST-LIVO2 dataset.}
    \label{tab:fast_livo_quan}
    \resizebox{0.48\textwidth}{!}{
    \begin{tabular}{cccccccccc}
        \toprule
        \textbf{Metrics} & \textbf{Methods} & \textbf{Campus} & \textbf{Sculpture} & \textbf{Culture} & \textbf{Drive} & \textbf{Avg.} \\
        \midrule
  
        \multirow{3}*{SSIM$\uparrow$}  
        & {InstantNGP} 
        & 0.789 & 0.698 & 0.670 & 0.697 & 0.714
        \\
        
        & {3DGS$^{\dagger}$} 
        & \textbf{0.849} & \textbf{{0.769}} & \underline{{0.726}} & \textbf{0.778} & \textbf{0.780}
        \\

        & {Ours}
        & \underline{0.834} & \underline{0.729} & \textbf{0.727} & \underline{0.764} & \underline{0.764}
        \\
        \midrule
  
        \multirow{3}*{PSNR$\uparrow$}  
        & {InstantNGP} 
        & 28.880 & 22.356 & 21.563 & 24.145 & 24.236
        \\
  
        & {3DGS$^{\dagger}$} 
        & \textbf{31.310} & \textbf{{24.128}} & \underline{21.764} & \underline{25.837} & \underline{25.760}
        \\
        
        & {Ours}
        & \underline{30.681} & \underline{23.453} & \textbf{24.695} & \textbf{25.941} & \textbf{26.193}
        \\
        \midrule
  
        \multirow{3}*{LPIPS$\downarrow$}  
        
        & {InstantNGP} 
        & 0.255 & 0.376 & 0.428 & 0.416 & 0.369
        \\

        & {3DGS$^{\dagger}$} 
        & \textbf{0.182} & \textbf{{0.266}} & \underline{{{0.361}}} & \underline{0.296} & \textbf{0.276}
        \\
        & {Ours}
        & \underline{0.210} & \underline{0.321} & \textbf{0.350} & \textbf{0.293} & \underline{0.293}
        \\
        \bottomrule
    \end{tabular}
    }
    \vspace{-8pt}
\end{table}

\subsection{FAST-LIVO2 Dataset}

\begin{figure}
    \centering
    \subfigure{
        \centering
        \includegraphics[width=0.15\textwidth]{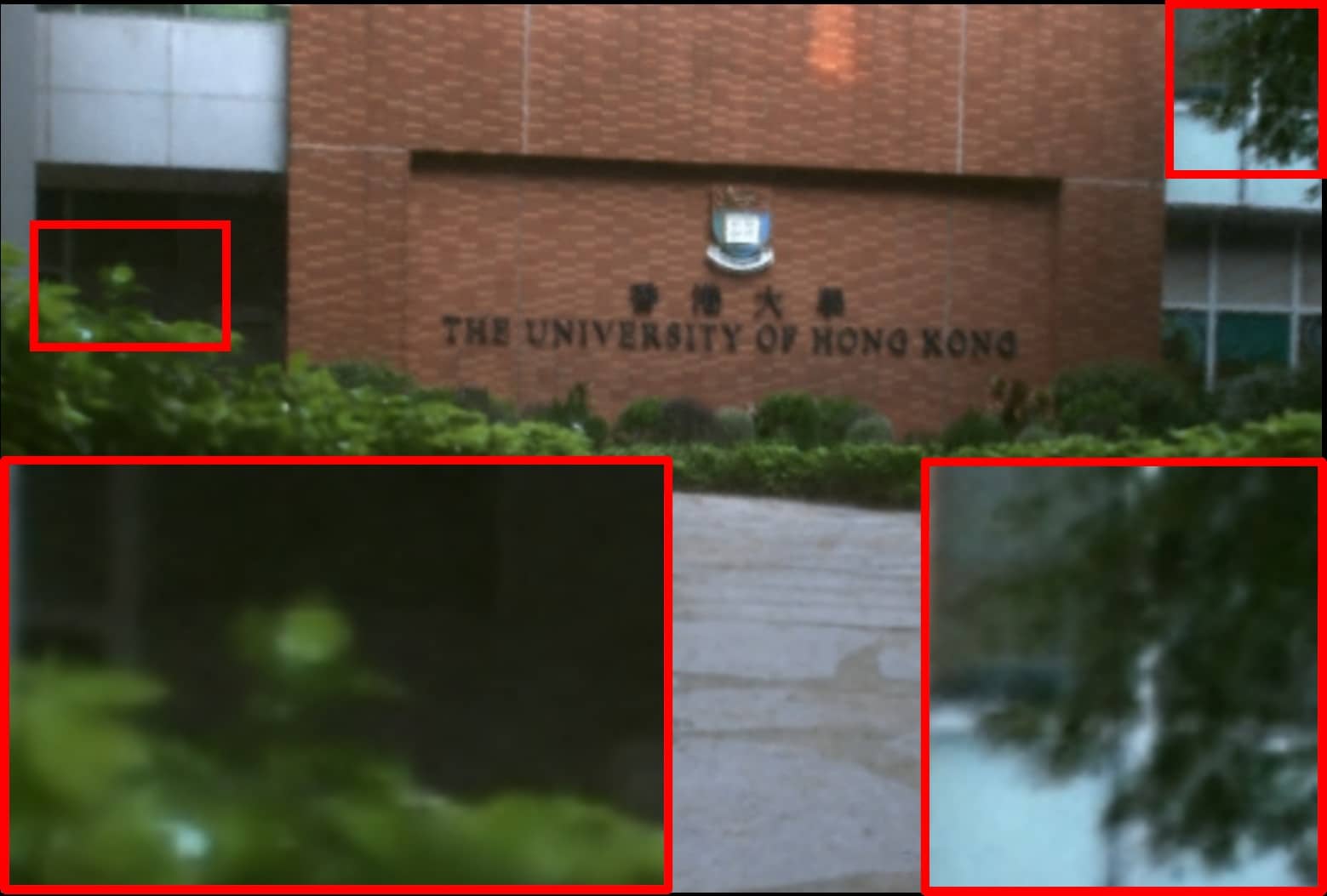}
    }
    \hspace{-11pt}
    \subfigure{
    \centering
    \includegraphics[width=0.15\textwidth]{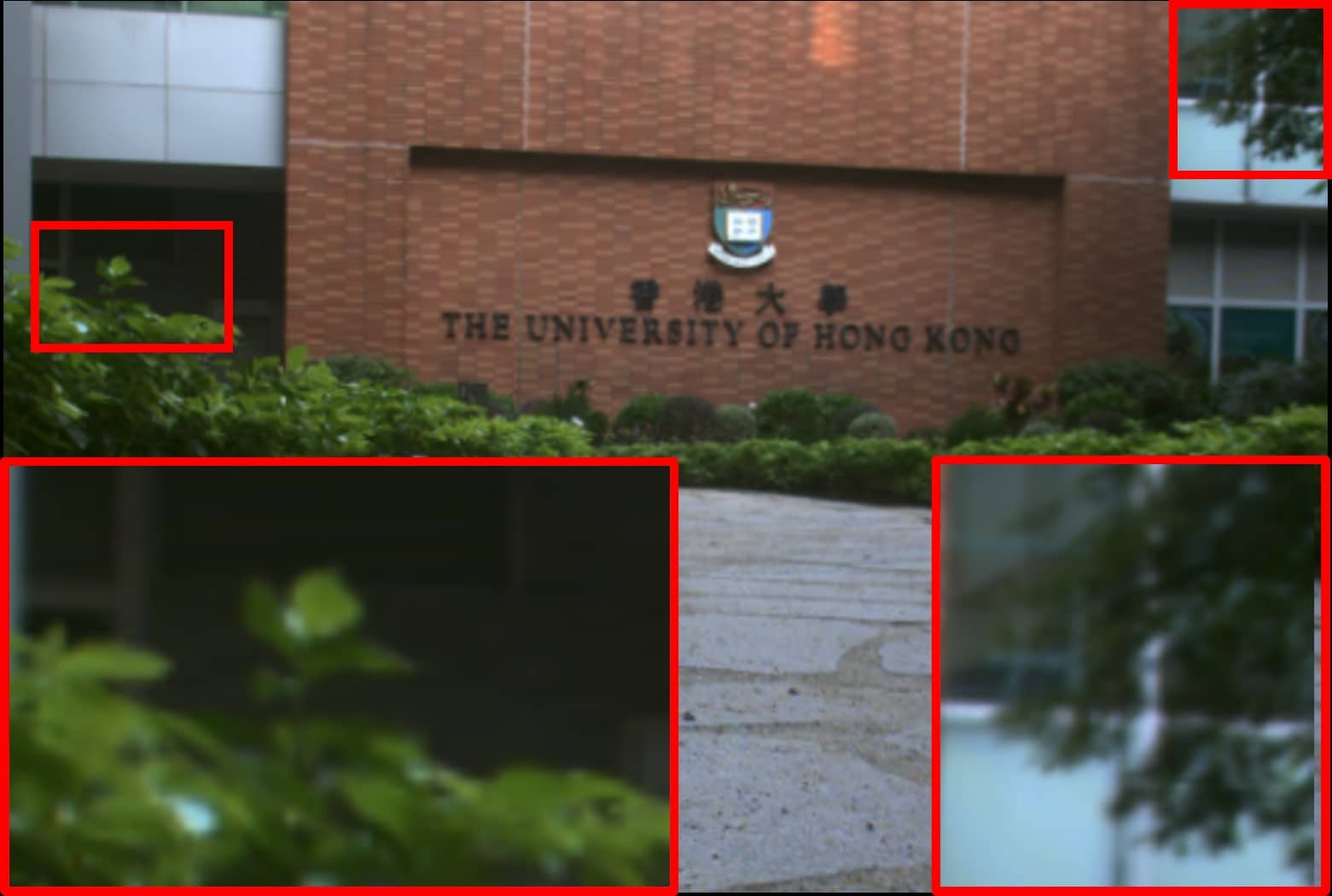}
    }
    \hspace{-11pt}
    \subfigure{
    \centering
    \includegraphics[width=0.15\textwidth]{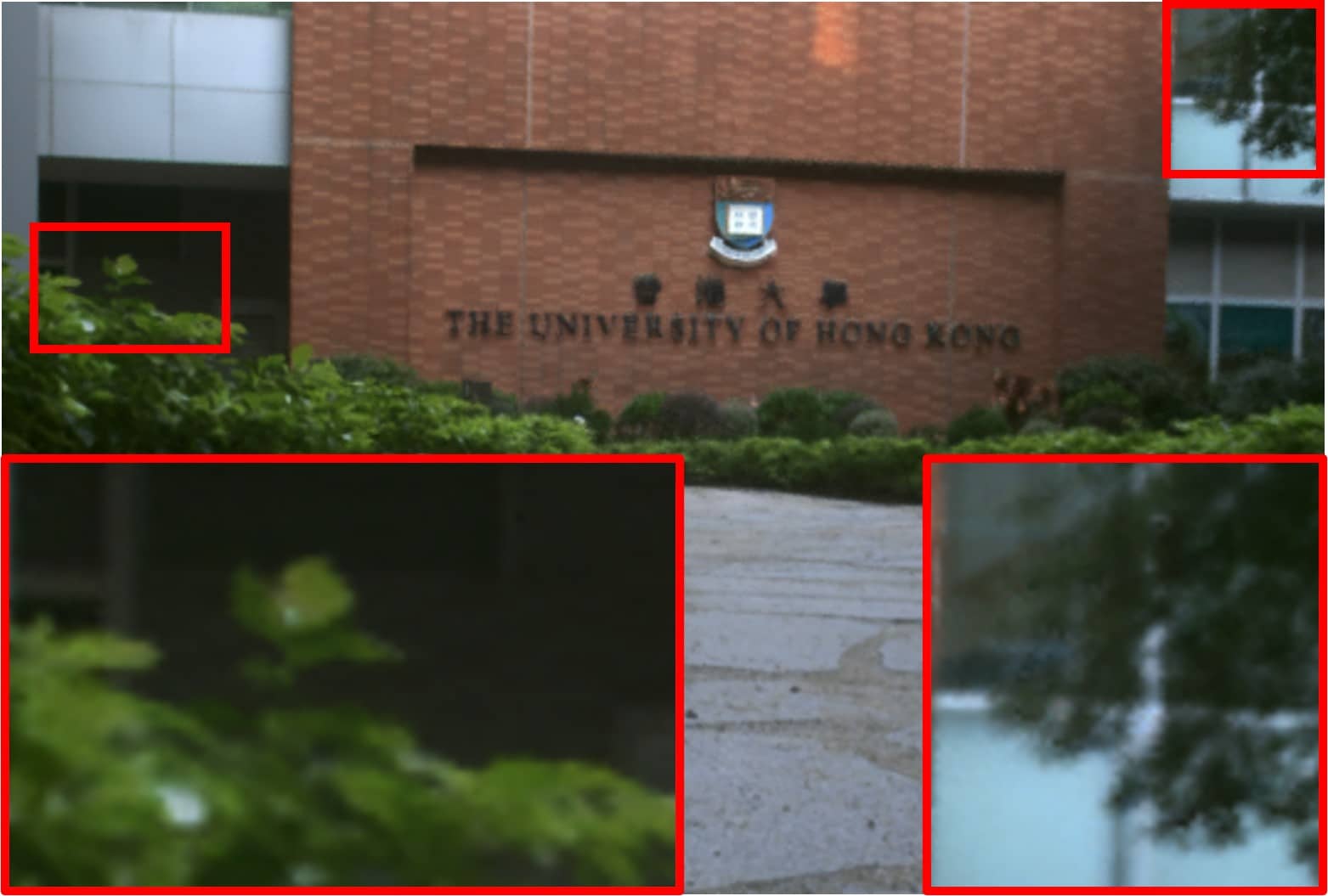}
    }

    \vspace{-8pt}
    
    \setcounter{subfigure}{0}
    \subfigure[InstantNGP]{
        \centering
        \includegraphics[width=0.15\textwidth]{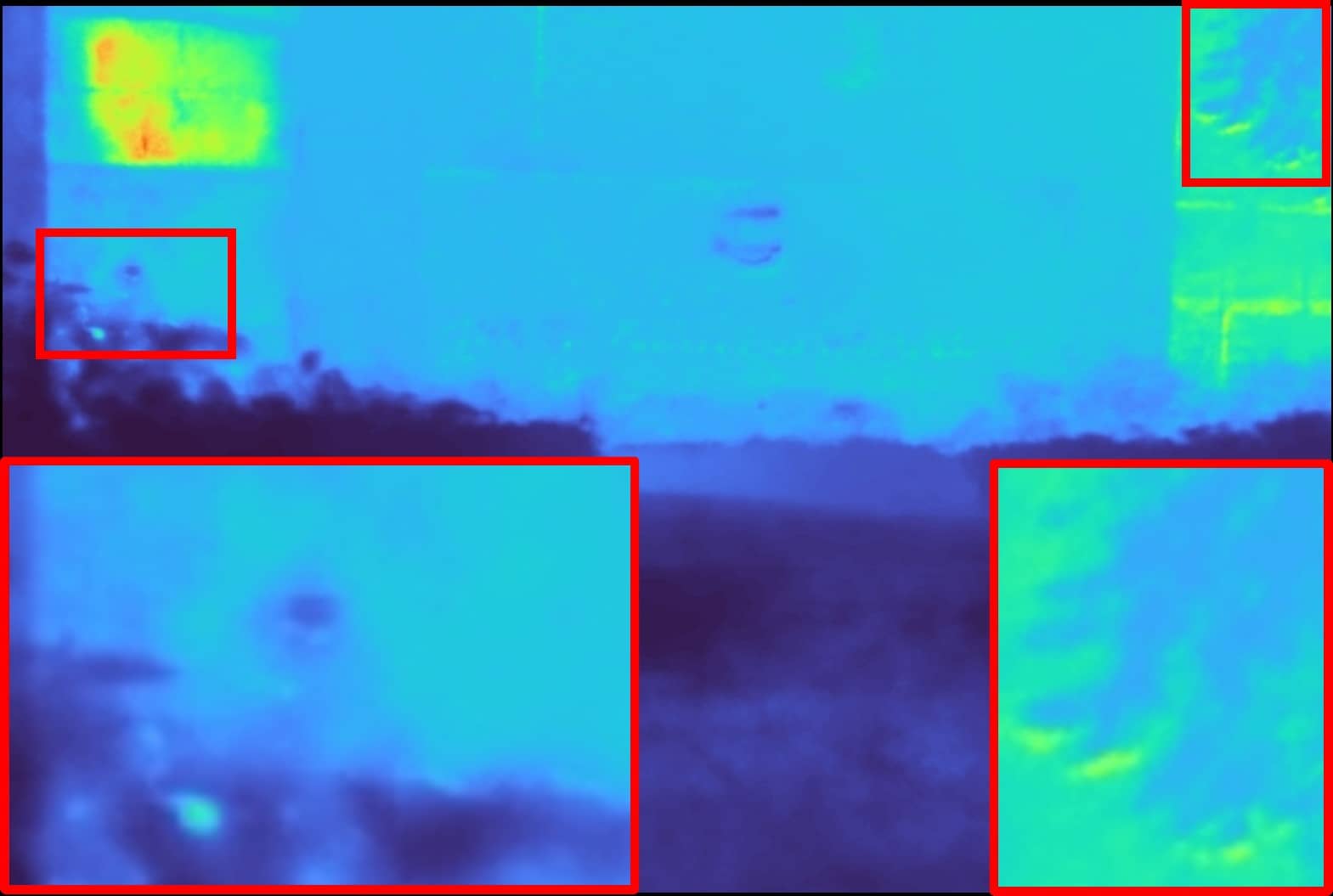}
    }
    \hspace{-11pt}
    \subfigure[3DGS$^{\dagger}$]{
    \centering
    \includegraphics[width=0.15\textwidth]{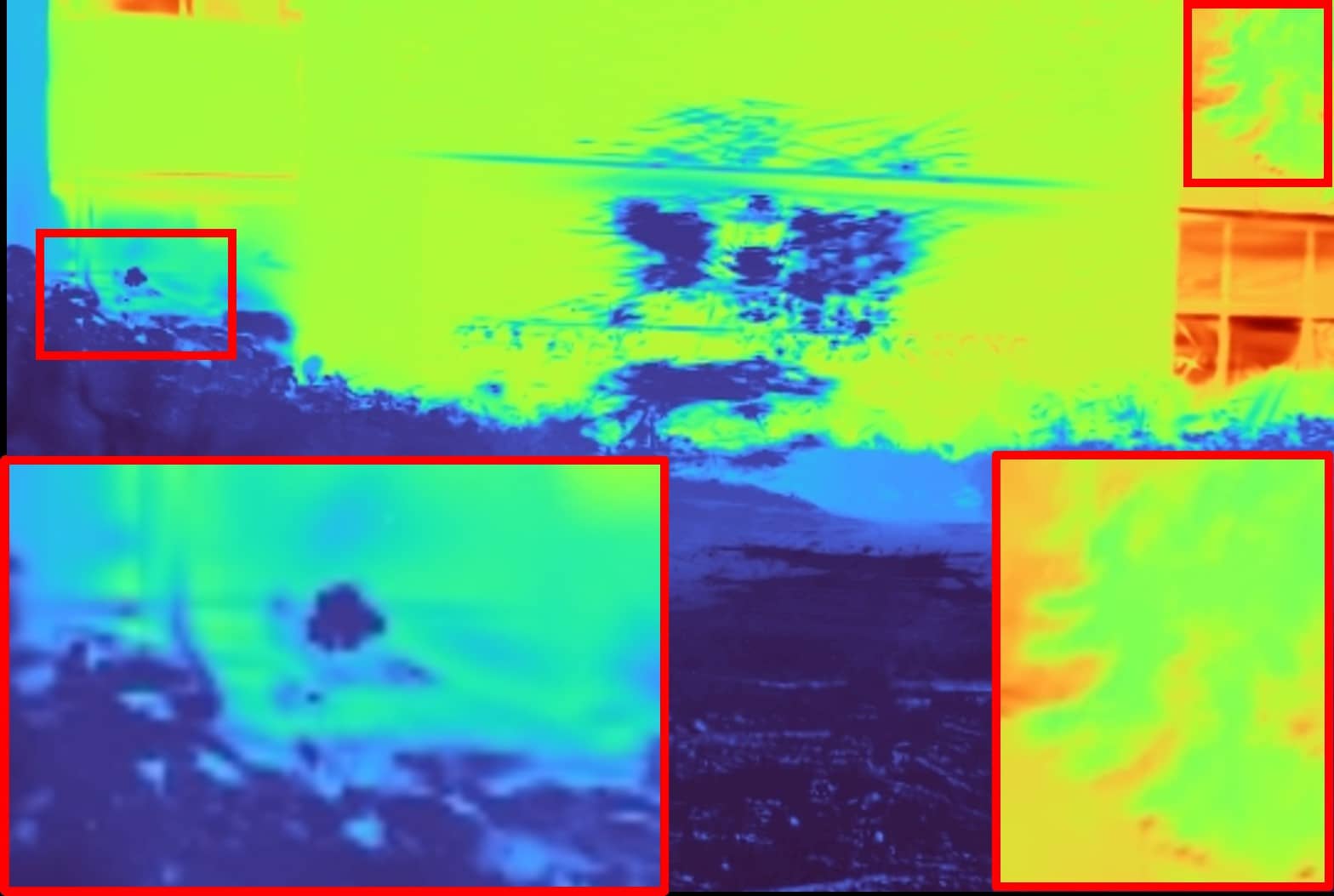}
    }
    \hspace{-11pt}
    \subfigure[Ours]{
    \centering
    \includegraphics[width=0.15\textwidth]{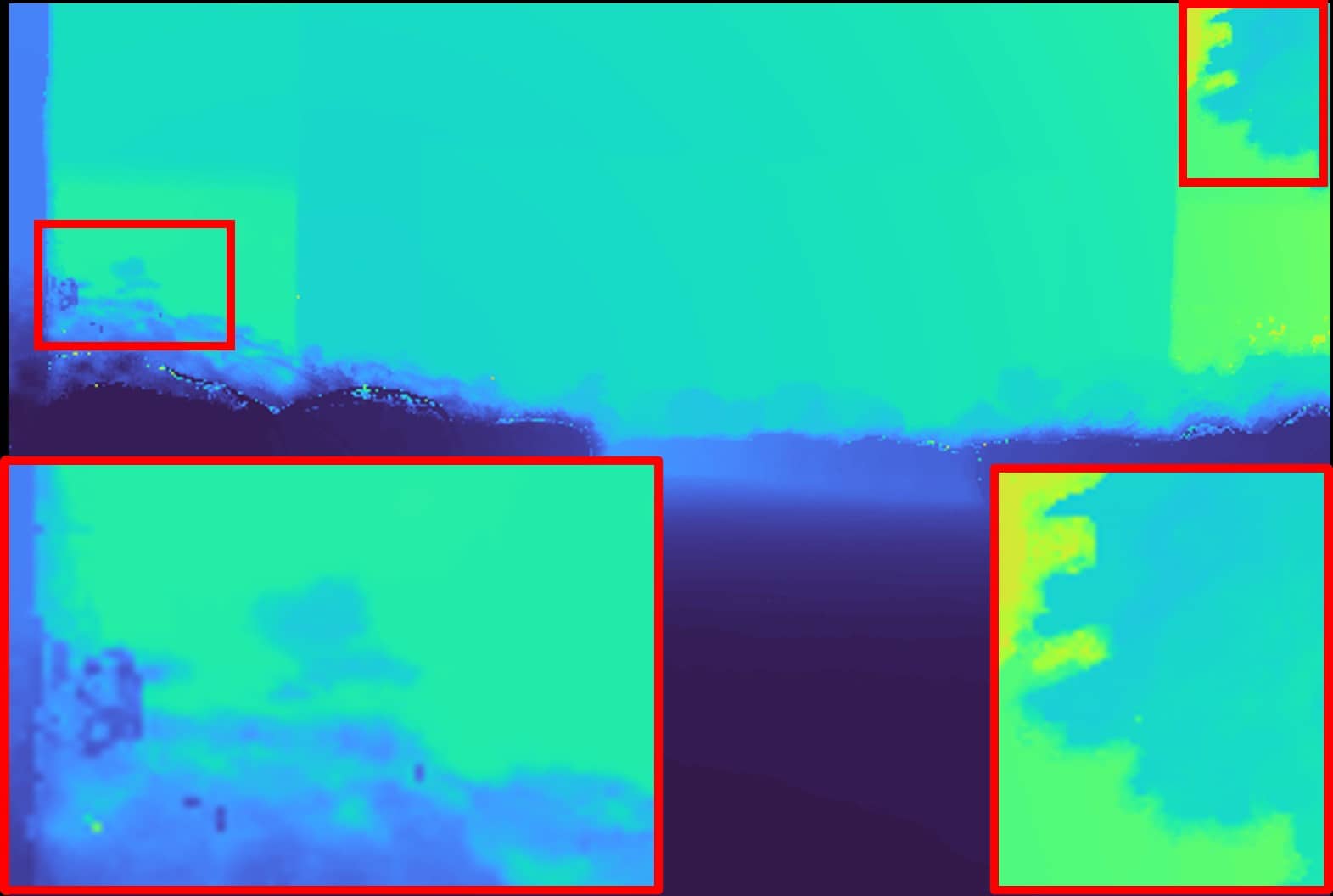}
    }

    \caption{
    Rendering results between different baselines on the FAST-LIVO dataset's Campus scene.}
    \label{fig:sampler_qual}
    \vspace{-12pt}
\end{figure}

For real-world datasets, we evaluate three types of trajectory datasets collected with a camera and a LiDAR from the FAST-LIVO2 datasets \cite{zheng2024fast}: forward-facing (Campus), object-centric (Sculpture), and free-view (Culture, Drive) trajectories, as shown in Fig.~\ref{fig:fast_livo_inter}.
We use the localization results from FAST-LIVO2 as ground truth poses and use a train-test split for dataset interpolation evaluation, where every 8th photo is used for the test set and the rest for the training set \cite{kerbl20233d}.
The quantitative results are shown in Tab.~\ref{tab:fast_livo_quan}. 
RGBD-based methods \cite{jiang2023h, GSSLAM2024} are not included as they are not compatible with LiDAR-visual systems. 
Our method demonstrates competitive rendering performance with the state-of-the-art method 3DGS \cite{kerbl20233d} and outperforms others in complex free-view trajectory scenes.
The interpolation qualitative results are shown in Fig.~\ref{fig:sampler_qual} (Left).
Our method captures precise geometric details in fuzzy objects (leaf) and avoids overfitting occurring in InstantNGP and 3DGS$^{\dagger}$ (middle wall's depth).
We further present our complete and detailed surface reconstruction results and extrapolation rendering results on the FAST-LIVO2 datasets, as shown in Fig~\ref{fig:fast_livo_inter} and Fig.~\ref{fig:fast_livo2_qual}.
In extrapolating views, all methods show some degree of degradation in rendering quality, where our method renders more consistent results than other methods, especially in the free-trajectory-type scene, where the camera moves freely, and lacks co-visible views.

\section{Conclusion}

This study presents a NeRF-based Surface Reconstruction and Rendering for LiDAR-visual systems. 
It bridges the gap between NDF and NeRF, harnessing the strengths of both to provide high-quality structure and appearance for a scene. 
The framework leverages structural information from LiDAR point clouds to build a visible-aware occupancy map prior for efficiency and a scalable NDF for high-granularity surface reconstruction. 
The NDF enables a structure-aware sampling for NeRF to conduct accurate structure rendering. 
The NeRF also utilizes photometric errors from rendering to refine structures. 
Extensive experiments validate the effectiveness of the proposed method across various scenarios, demonstrating its potential for real-world applications in computer vision and robotics.




{
\bibliographystyle{IEEEtran}
\balance
\bibliography{reference}
}
\input{supply}
\end{document}

%% file: supply.tex
\clearpage
\setcounter{equation}{0}
\setcounter{figure}{0}
\setcounter{table}{0}
\setcounter{page}{1}
\setcounter{section}{0}%
\setcounter{subsection}{0}%
\setcounter{subsubsection}{0}%
\setcounter{paragraph}{0}%

\normalem 

\graphicspath{{figures/}}

\begin{center}
	\textbf{\large Supplementary Material for Neural Surface Reconstruction and Rendering for Lidar-Visual Systems}
\end{center}

\maketitle
\thispagestyle{empty}
\pagestyle{empty}


\section{Supplementary Details of Methodology}

\subsection{Structure-aware sampling}\label{sec:st_sampling}
\begin{figure}[htb]
    \centering
    \includegraphics[width=0.48\textwidth]{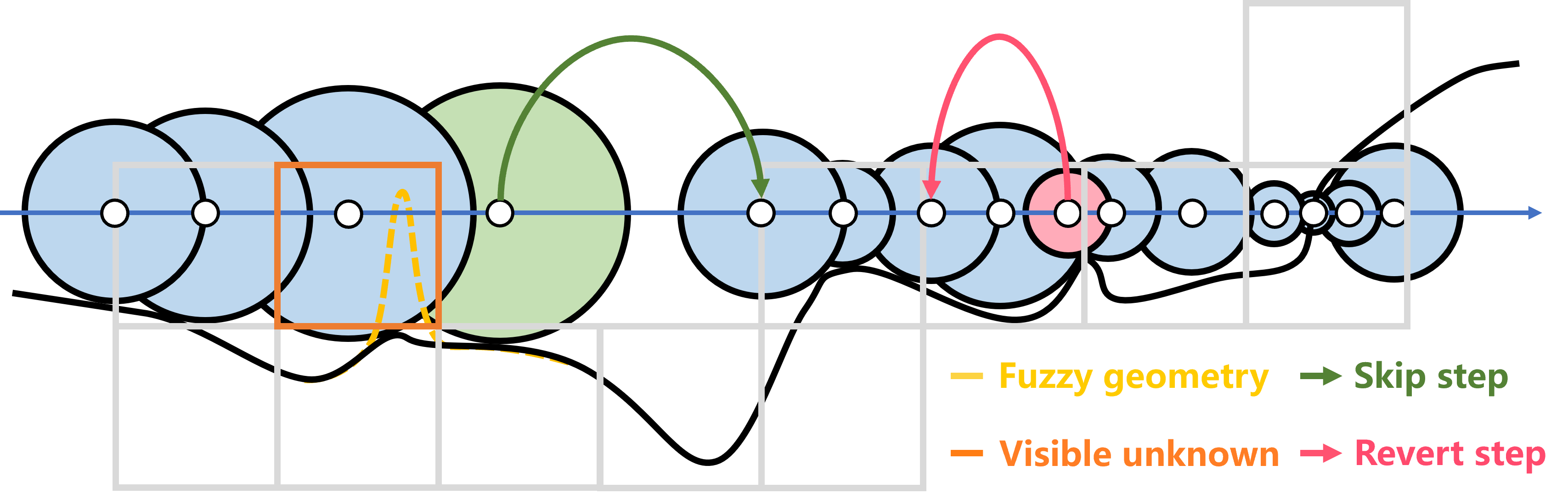}
    \caption{Illustrations of visible-aware occupancy map and structure-aware sampling based on adaptive sphere tracing. Each circle has a radius equal to the SDF value at the sample point.}
    \label{fig:st}
\end{figure}

The overall algorithm is shown in Alg.~\ref{alg:ast} and Fig.~\ref{fig:st}.
Given any desired rendering direction $\boldsymbol{d}$, the algorithm starts from the camera origin (Alg.~\ref{alg:ast}-\ref{alg:ast:init}), and iteratively steps forward (Alg.~\ref{alg:ast}-\ref{alg:ast:step}) along the ray direction according to its signed distance value. 
The gradient of the signed distance field along the ray direction ${M}$ is estimated using linear interpolation (Alg.\ref{alg:ast}-\ref{alg:ast:M}).
A filtered gradient $m$ is updated with a relaxation coefficient $\gamma = 0.7$ (Alg.\ref{alg:ast}-\ref{alg:ast:m}), which adaptively determines the next step size $\delta_i$ (Alg.~\ref{alg:ast}-\ref{alg:ast:delta}) for efficiency.
At every step, we ensure that the space between two adjacent steps is intersected (Alg.~\ref{alg:ast}-\ref{alg:ast:check}), wherein we take predicted scale $\beta_i$ into account to avoid triggering false revert steps in unconverged NDF.
To avoid poor local minima, we keep marching even behind surfaces (Alg.\ref{alg:ast}-\ref{alg:ast:step}), and until a ray's transmittance (Alg.~\ref{alg:ast}-\ref{alg:ast:T}) falls below a threshold $\epsilon_T = 0.001$.
The filtered slope $m$ is verified by sphere tracing, and smaller step sizes near surfaces yield a higher sampling frequency for more accurate slope estimations.
Therefore, we apply the estimated filtered slope $M$ in the SDF-to-density transition to avoid expensive analytical or numerical gradient calculations.
We further utilize the visible-aware occupancy map to skip the free space for efficiency, as shown in Fig.~\ref{fig:st}'s skip step.

\subsection{Losses}\label{sec:loss}

The Eikonal loss \cite{gropp2020implicit} and curvature loss \cite{yang2023steik} are further employed for both NeRF and NSDF samples to prevent the zero-everywhere and overfitting solutions for the SDF:
\begin{equation}
    \begin{aligned}
    \mathcal{L}_{eik} &= \frac{1}{N_i}\sum_i \left( \|\nabla f_{\mathcal{S}}(\boldsymbol{x}_i)\|_2 - 1 \right)^2, \\
    \mathcal{L}_{curv} &= \frac{1}{N_i}\sum_i  |\nabla^2 f_{\mathcal{S}}(\boldsymbol{x}_i)|  ,
    \end{aligned}
\end{equation}
where $\nabla f_{\mathcal{S}}(\boldsymbol{x}i)$ and $\nabla^2 f_{\mathcal{S}}(\boldsymbol{x}_i)$ are the gradient and Hessian of the SDF at $\boldsymbol{x}_i$ calculated by numerical differentiation with a progressively smaller step size \cite{li2023neuralangelo}.


The overall training loss is defined as follows:
\begin{equation}
    \begin{aligned}
    \mathcal{L} = \mathcal{L}_{sdf} +\lambda_{rgb} \mathcal{L}_{rgb} + \lambda_{eik} \mathcal{L}_{eik} + \lambda_{curv} \mathcal{L}_{curv},
    \end{aligned}
\end{equation}
where $\lambda_{eik} = 0.1$ and $\lambda_{curv} = 5\times10^{-4}$ are the weights for the Eikonal, and curvature losses, respectively. 
We schedule the $\lambda_{rgb}$ linearly increases from $10^{-4}$ to $10$ during the training to ensure that the NeRF learns on a well-structured NSDF to avoid local minima. 

\begin{algorithm}[h]
    \caption{Structure-aware Sampling}
    \label{alg:ast}
    \KwIn{Neural distance field $f_{\mathcal{S}}$, point on rendering ray $\boldsymbol{x}(t) = \boldsymbol{o} + t \boldsymbol{d}$, termination conditions $\epsilon_{T}$, relaxation coefficient $\gamma \in(0,1)$.}
    \KwOut{ray samples $\{t_i, m_i\}$.}
    \SetKwInput{Not}{Notation}
    \Not{Slope $M$, filtered slope $m$, transmittance $T$.} \label{alg:ast:init}

    $t_i:=0,\ (s_{i},\beta_{i}):=f_{\mathcal{S}}(\boldsymbol{x}(t_i)),\ m:=-1,\ T:=1$\\ \label{alg:ast:init}
    \While{$s_i > 0 \cup T > \epsilon_{T}$}{
        $\delta_i:= |s_i|*\frac{2}{1-m} \quad \triangleright \text { Adaptive step}$\\ \label{alg:ast:delta}
        $t_{i+1}:=t_{i} + \delta_i$\\ \label{alg:ast:step}
        $\left(s_{i+1},\beta_{i+1}\right):=f_{\mathcal{S}}\left(\boldsymbol{x}\left(t_{i+1}\right)\right)$\\
        \If{$\delta_i \leq |s_i|+|s_{i+1}| + 3\beta_{i}$\label{alg:ast:check} } 
        {
            $T:=T *\exp\left(-\sigma\left(s_{i},\beta_{i},m\right)\delta_i\right)$\\\label{alg:ast:T}
            $M:={\left(s_{i+1}-s_{i}\right)}/{\delta_i}$\\\label{alg:ast:M}
            $m:= \gamma m+\left(1-\gamma\right) M $\\\label{alg:ast:m}
            $\left(t_i, m_i\right):= \left(t_{i+1}, m\right) \quad \triangleright \text { Sample step}$\\ 
        }
        \Else{
            $m:=-1 \quad \triangleright \text { Revert step}$
        }
    }
\end{algorithm}

\subsection{Training}\label{sec:training}

\begin{figure*}[!t]
    \centering
    \subfigure{
        \centering
        \includegraphics[width=0.28\textwidth]{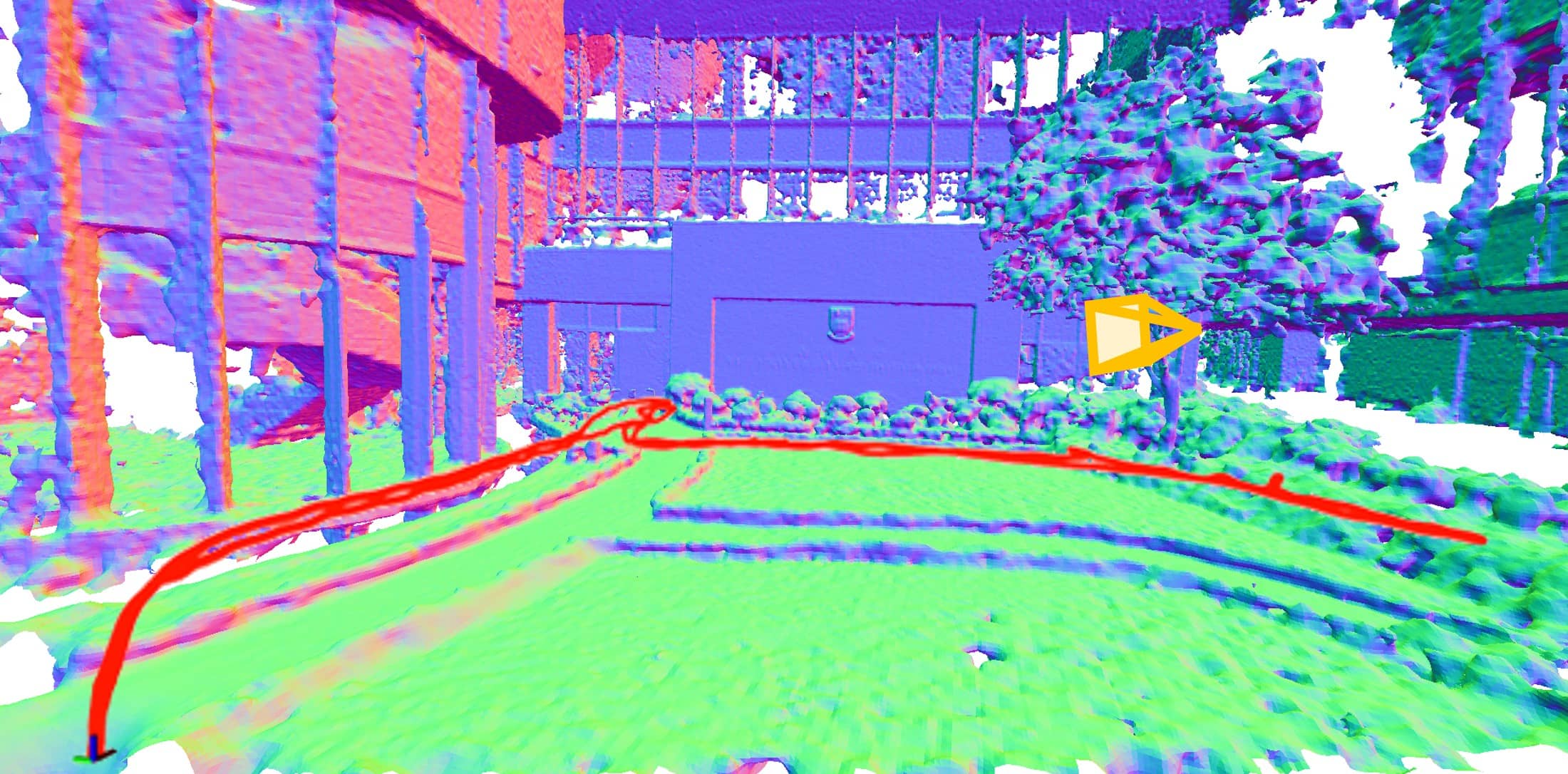}
    }
    \hspace{-12pt}
    \subfigure{
    \centering
    \includegraphics[width=0.173\textwidth]{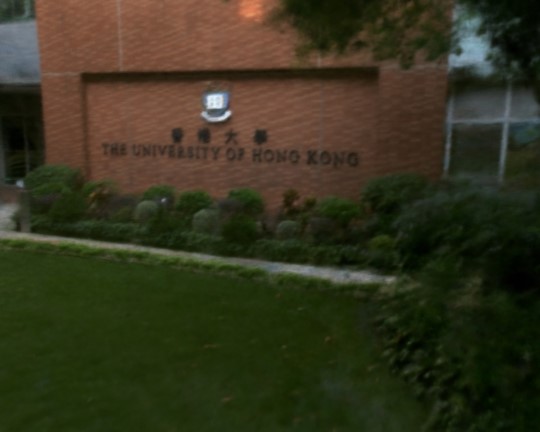}
    }
    \hspace{-12pt}
    \subfigure{
    \centering
    \includegraphics[width=0.173\textwidth]{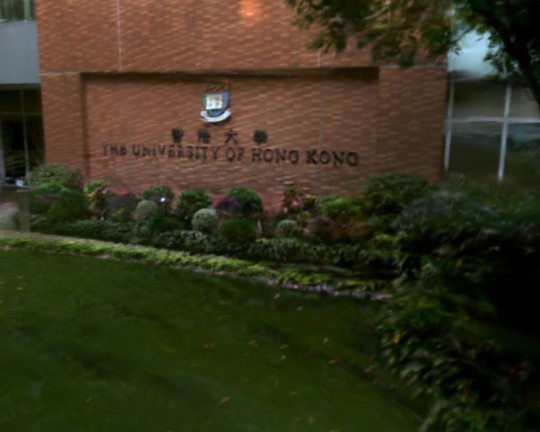}
    }
    \hspace{-12pt}
    \subfigure{
    \centering
    \includegraphics[width=0.173\textwidth]{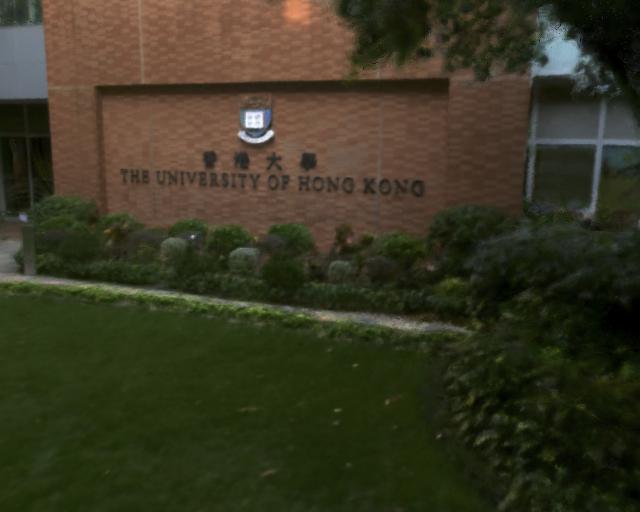}
    }
    \hspace{-12pt}
    \subfigure{
    \centering
    \includegraphics[width=0.173\textwidth]{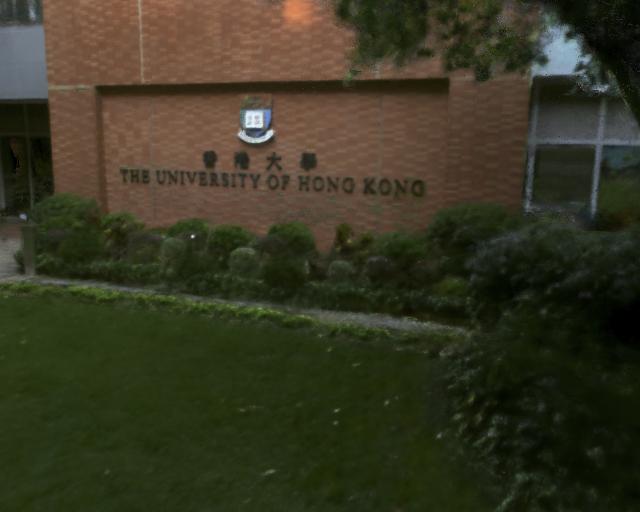}
    }

    \vspace{-8pt}
    
    \subfigure{
        \centering
        \includegraphics[width=0.28\textwidth]{figures/fast_livo2/mesh/red_sculture.jpeg}
    }
    \hspace{-12pt}
    \subfigure{
    \centering
    \includegraphics[width=0.173\textwidth]{figures/fast_livo2/extra/redsculture_ingp.jpeg}
    }
    \hspace{-12pt}
    \subfigure{
    \centering
    \includegraphics[width=0.173\textwidth]{figures/fast_livo2/extra/redsculture_3dgs.jpeg}
    }
    \hspace{-12pt}
    \subfigure{
    \centering
    \includegraphics[width=0.173\textwidth]{figures/fast_livo2/extra/redsculture_ours.jpeg}
    }
    \hspace{-12pt}
    \subfigure{
    \centering
    \includegraphics[width=0.173\textwidth]{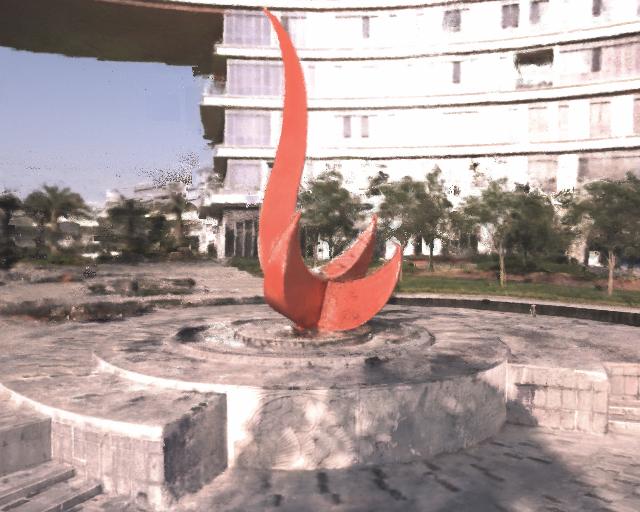}
    }

    \vspace{-8pt}
    
    \subfigure{
        \centering
        \includegraphics[width=0.28\textwidth]{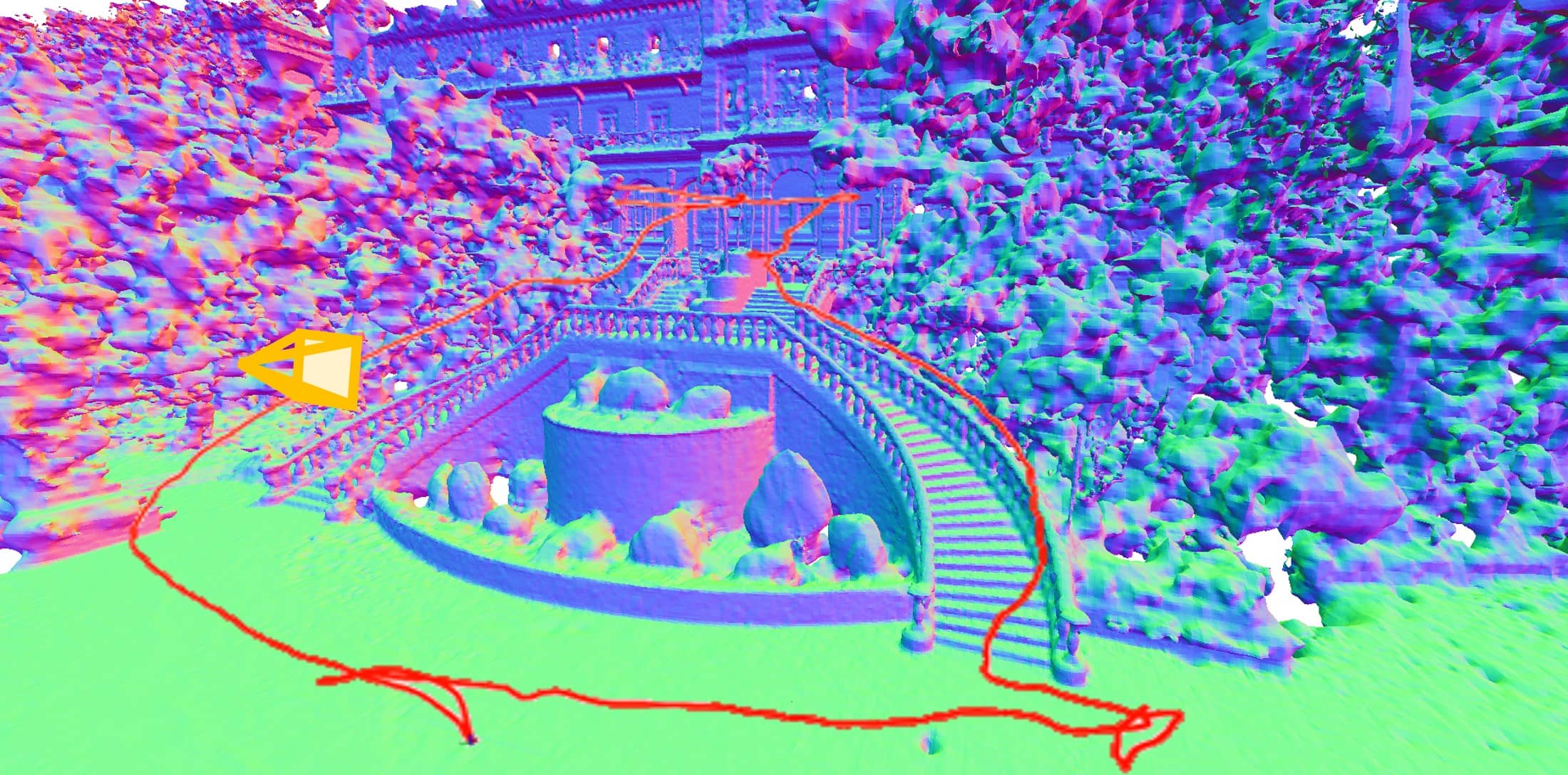}
    }
    \hspace{-12pt}
    \subfigure{
    \centering
    \includegraphics[width=0.173\textwidth]{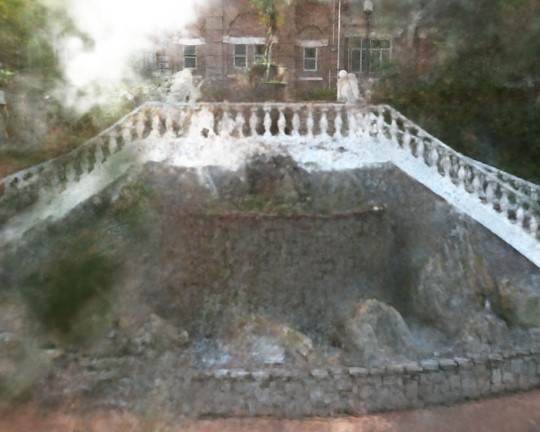}
    }
    \hspace{-12pt}
    \subfigure{
    \centering
    \includegraphics[width=0.173\textwidth]{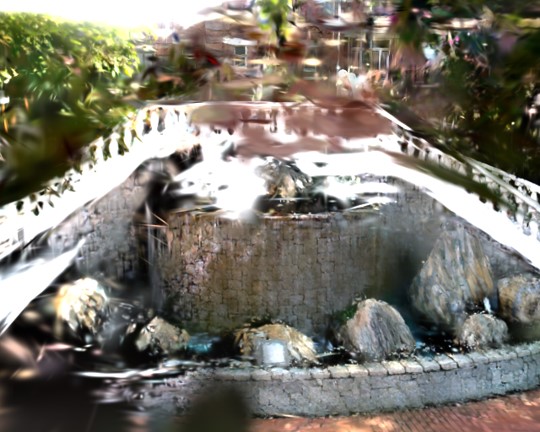}
    }
    \hspace{-12pt}
    \subfigure{
    \centering
    \includegraphics[width=0.173\textwidth]{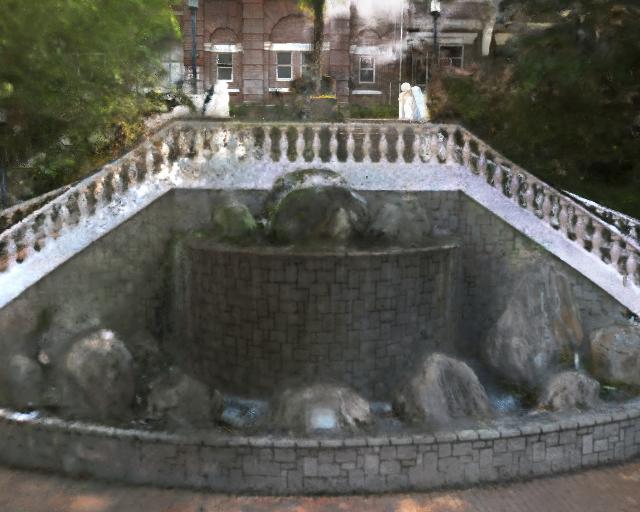}
    }
    \hspace{-12pt}
    \subfigure{
    \centering
    \includegraphics[width=0.173\textwidth]{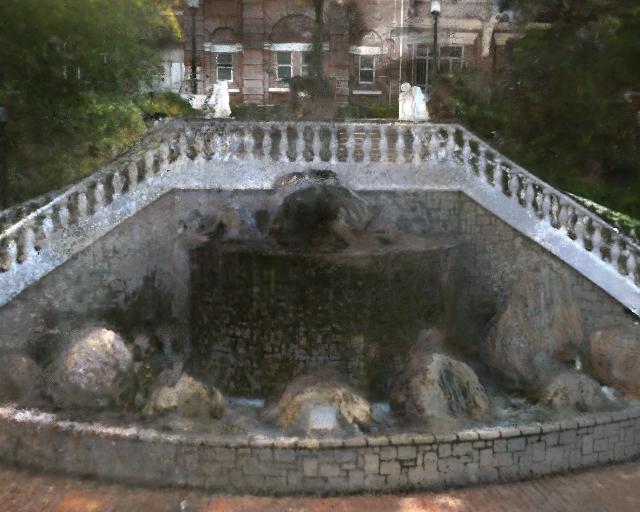}
    }

    \vspace{-8pt}
    
    \setcounter{subfigure}{0}
    \subfigure[Mesh (Ours)]{
        \centering
        \includegraphics[width=0.28\textwidth]{figures/fast_livo2/mesh/drive.jpg}
    }
    \hspace{-12pt}
    \subfigure[InstantNGP]{
    \centering
    \includegraphics[width=0.173\textwidth]{figures/fast_livo2/extra/drive_ingp.jpg}
    }
    \hspace{-12pt}
    \subfigure[3DGS$^{\dagger}$]{
    \centering
    \includegraphics[width=0.173\textwidth]{figures/fast_livo2/extra/drive_3dgs.jpg}
    }
    \hspace{-12pt}
    \subfigure[Ours]{
    \centering
    \includegraphics[width=0.173\textwidth]{figures/fast_livo2/extra/drive_ours.jpeg}
    }
    \hspace{-12pt}
    \subfigure[wo dir. scheduler]{
    \centering
    \includegraphics[width=0.173\textwidth]{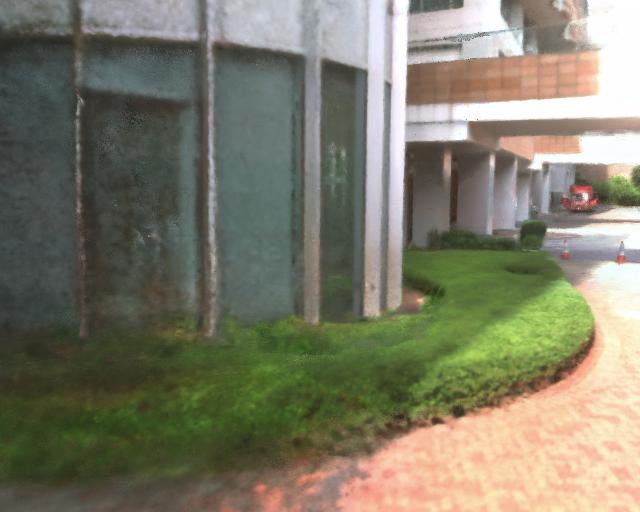}
    }
    
    \caption{
    The qualitative results of FAST-LIVO2 datasets (scenes from top to bottom are Campus, Sculpture, Culture, and Drive). We show our surface reconstruction results on the left, and the red line indicates the training path and the orange cameras indicates the extrapolation views for the right side's rendering results.}
    \label{fig:fast_livo2_qual}
\end{figure*}

\subsubsection{Outlier removal}\label{sec:outlier_removal}
The zero-level set of the NDF defines the fitting surface, and the inferred signed distance values of the input LiDAR points naturally indicate the reconstruction error. 
In dynamic scenes, points on dynamic objects are supervised with an SDF value of zero. However, LiDAR points on the static background traversing through these dynamic points produce a supervised SDF value greater than zero.
Given that dynamic points are typically sparse, this region is often dominated by the background points and has a SDF value greater than zero.
Based on this observation, we propose an outlier removal strategy that periodically infers the signed distance values of training LiDAR points and eliminates points whose predicted signed distance values are more than $\epsilon$ away from 0. 
This enables the NDF remains a static structure field and also helps to erase dynamic objects in rendering, as shown in Fig.~\ref{fig:outlier_removal}.

\subsubsection{Directional embedding scheduler}\label{sec:dir_scheduler}

To synthesize photorealistic novel-view images, the neural radiance field considers the view direction $\boldsymbol{d}$ to output the view-dependent color $\boldsymbol{c}$ at each position $\boldsymbol{x}$: $\boldsymbol{c} = f_{\mathcal{C}}(\boldsymbol{x},\boldsymbol{d})$, where the view direction is encoded using a 4-degree spherical harmonics encoding\cite{verbin2022ref}.
The tightly coupled position and view direction in training make the rendering quality in extrapolation views show a degree of color degeneration, especially at places that have only an image observations observing from similar directions, as shown in Fig.~\ref{fig:fast_livo2_qual} (d).
We consider that the surface's color is composed of view-independent diffuse color and view-dependent specular color.
We schedule the degree of directional embedding to learn the surface diffuse color at the degree of 0 (i.e., view-independent feature) in the beginning and gradually increase from 0 to 4 during training to learn high-frequency view-dependent specular color.

\subsubsection{Scene contraction for background rendering}
To address the infinitely far background space, we define a boundary space that extends outward from the occupied map for background color rendering.
For each ray, $n_b$ points are uniformly sampled in the boundary space and contracted \cite{barron2022mip} as follows:
\begin{equation}
    \begin{aligned}
        \operatorname{contract}(\boldsymbol{x})= \begin{cases}\boldsymbol{x} &,\ |\boldsymbol{x}| \leq 1 \\ 
        \left(1 + B\left(1-\frac{1}{|\boldsymbol{x}|}\right)\right)\left(\frac{\boldsymbol{x}}{|\boldsymbol{x}|}\right) &,\ |\boldsymbol{x}|> 1\end{cases},
    \end{aligned}
\end{equation}
where $B$ is the size of the extending boundary space and is defined as the same voxel size as the occupancy map.
The contracted samples are used to color the infinite space via volume rendering with false surfaces.

\section{Supplementary Experiments}


\subsection{Implementation Details}
We represent our neural fields following a similar architecture to InstantNGP \cite{muller2022instant}, utilizing a combination of multi-resolution hash encoding and a tiny MLP decoder.
The hash encoding resolution spans from $2^5$ to $2^{21}$ with 16 levels, and each level contains $2^{19}$  two-dimensional feature vectors. 
Given any position $x$, the hash encoding concatenates each level's interpolation features to form a feature vector of size 32.
One encoding feature is fed to a geometry MLP with 64-width and 3 hidden layers to obtain the SDF value and scale.
Another encoding feature is concatenated with the spherical harmonics encoding of view directions and fed to an appearance MLP with 64-width and 3 hidden layers to obtain the view-dependent color.
We sample 8192 rays for NDF training and follow InstantNGP\cite{muller2022instant} to fix the batch size of point samples as 256000 while the batch size of rays is dynamic per iteration.
We take 20000 iterations for training, with outlier removal performed every 2000 iterations. 
We implement our method using LibTorch and CUDA.
All experiments are conducted on a platform equipped with an Intel i7-13700K CPU and an NVIDIA RTX 4090 GPU.

\subsection{FAST-LIVO2 Dataset}

\begin{figure}
    \centering
    \setcounter{subfigure}{0}
    \subfigure[InstantNGP]{
      \centering
      \includegraphics[width=0.23\textwidth]{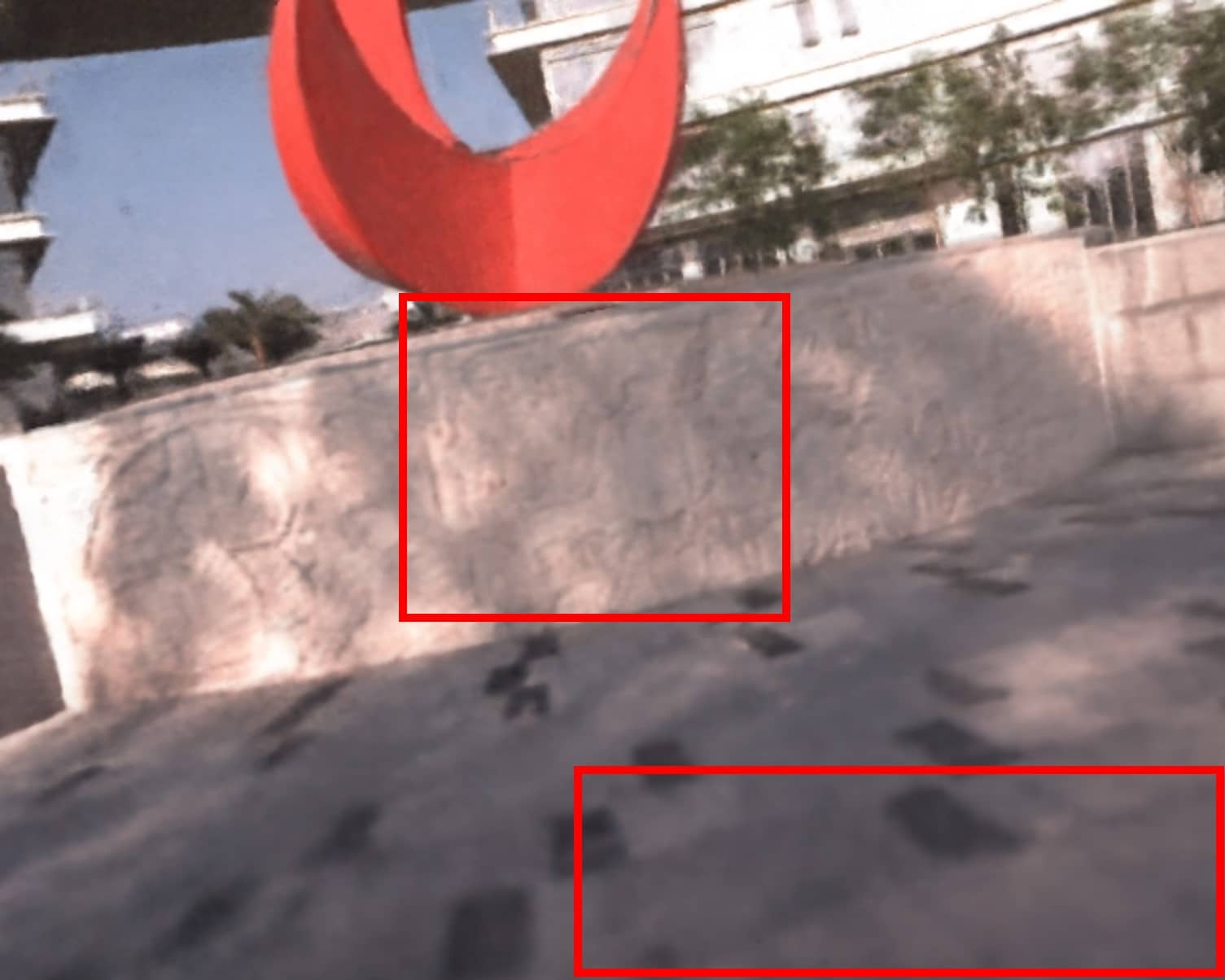}
  }
  \hspace{-12pt}
  \subfigure[3DGS]{
      \centering
      \includegraphics[width=0.23\textwidth]{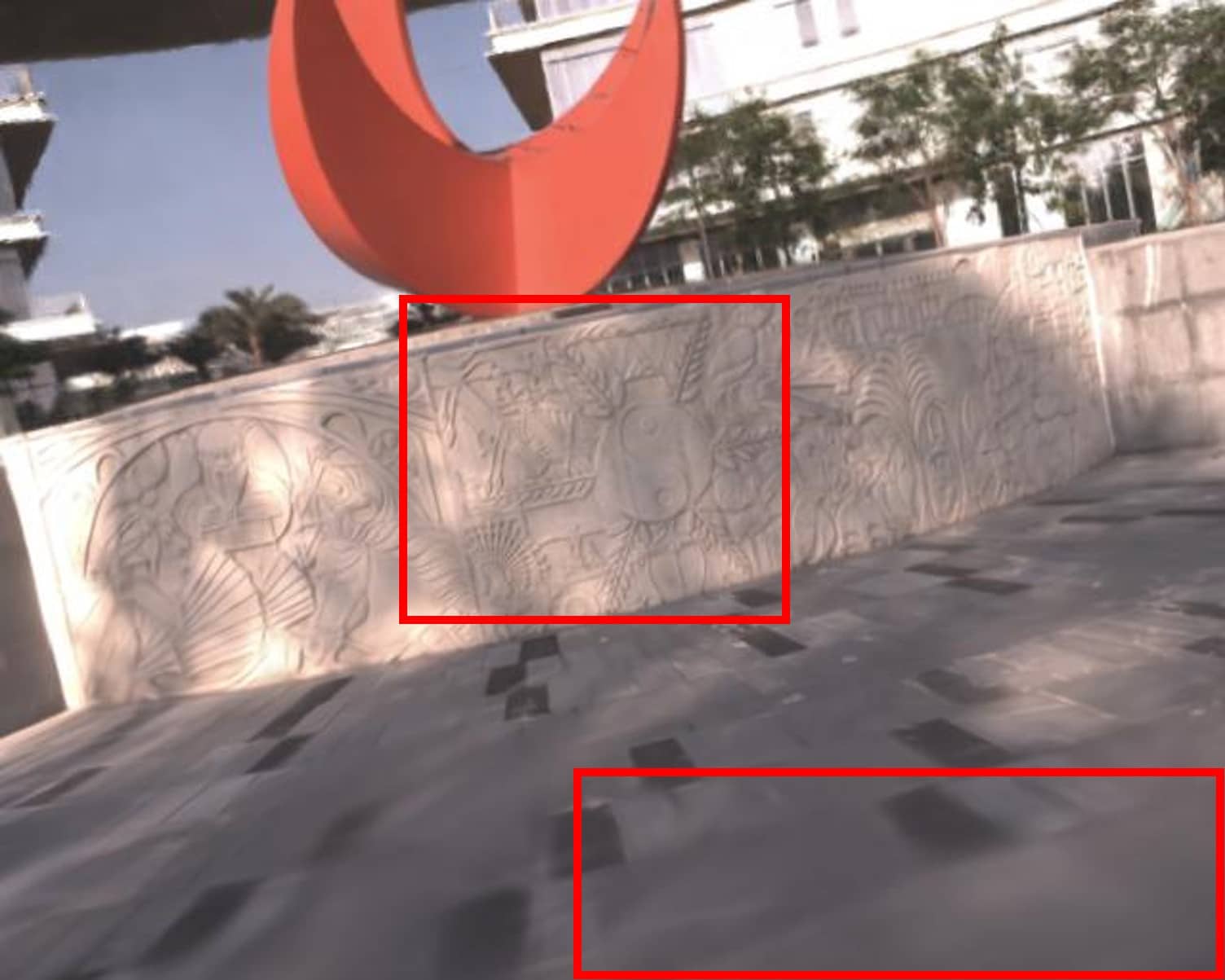}
  }

  \vspace{-4pt}
  
  \subfigure[Ours]{
    \centering
    \includegraphics[width=0.23\textwidth]{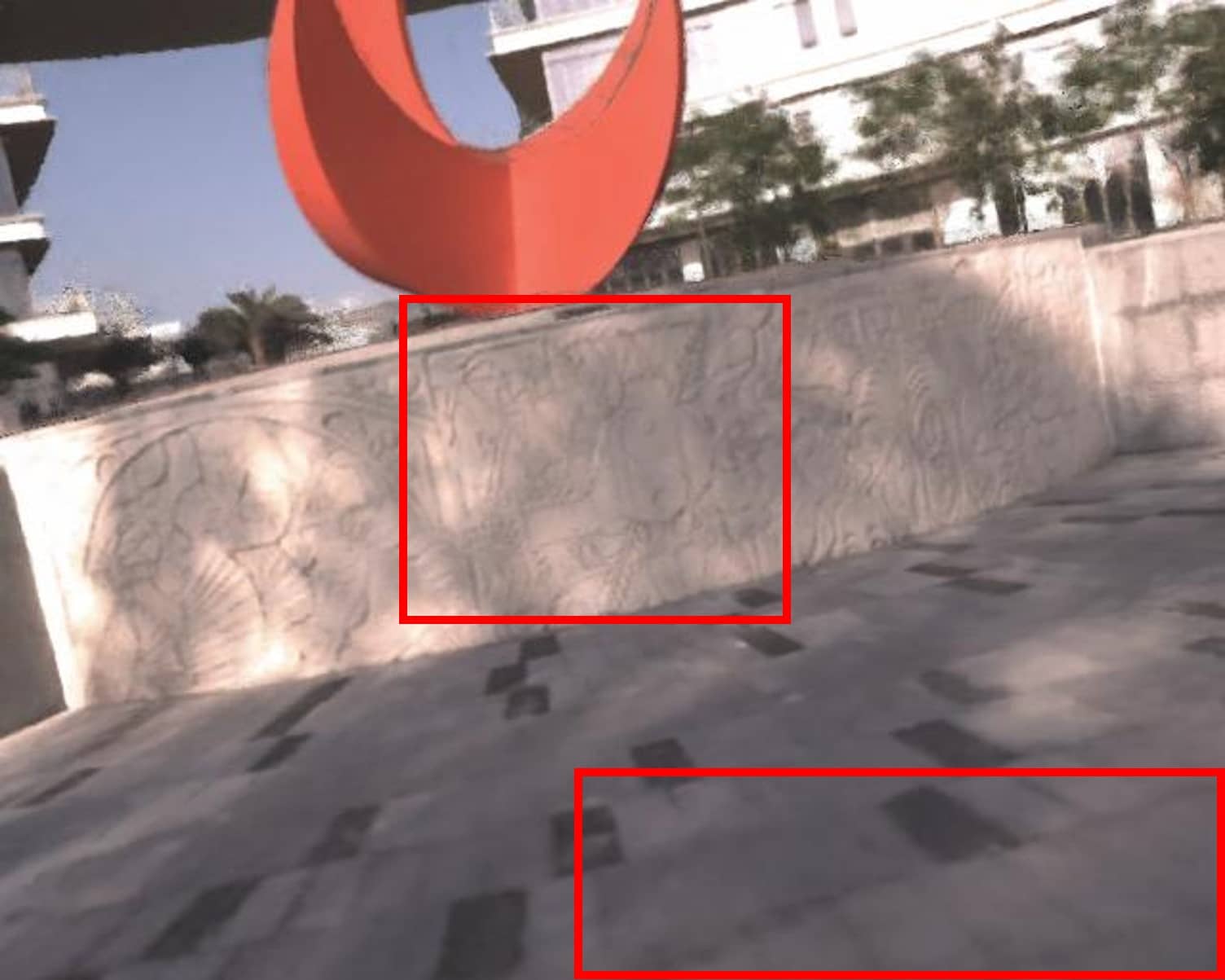}
}
\hspace{-12pt}
\subfigure[Ground Truth]{
    \centering
    \includegraphics[width=0.23\textwidth]{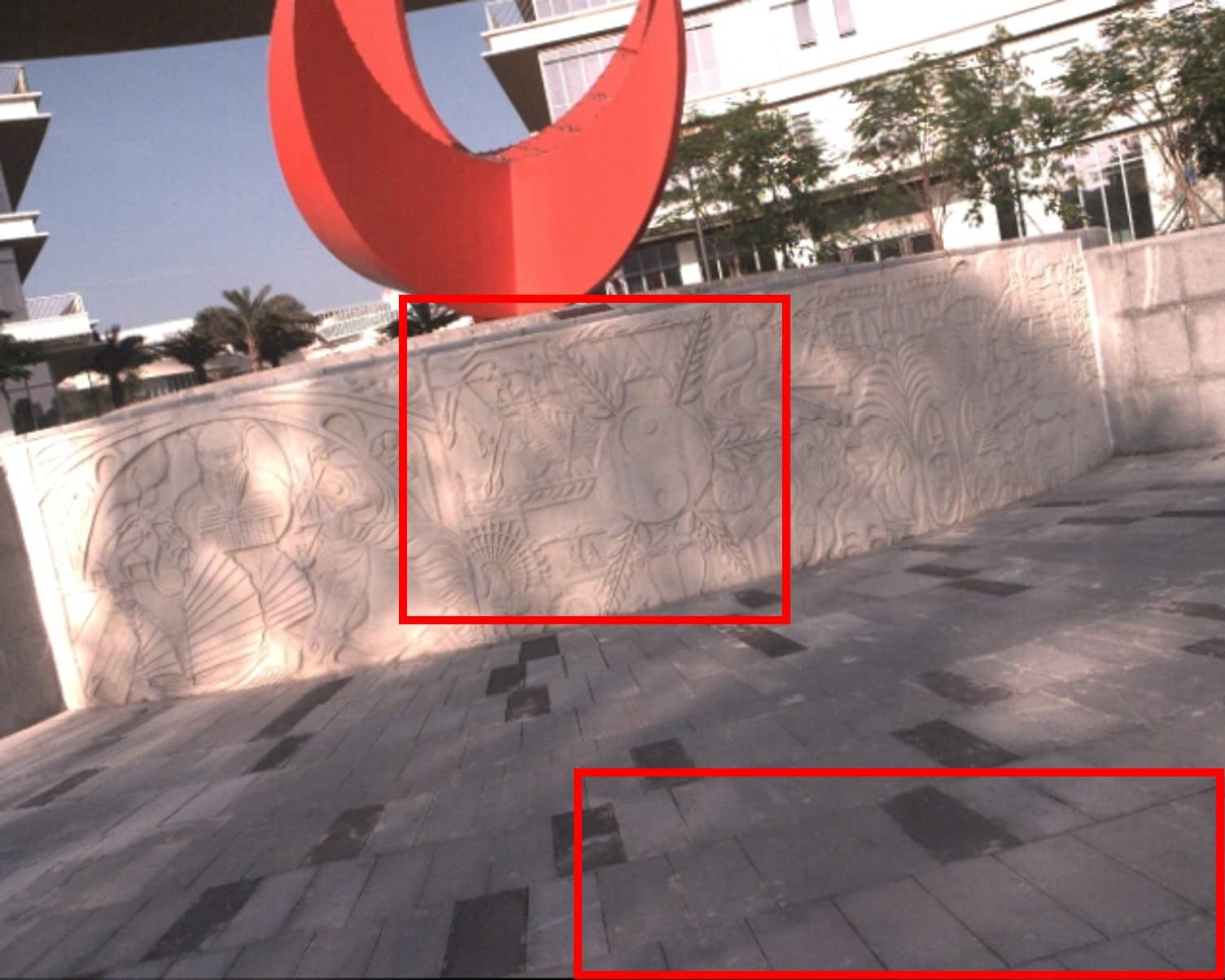}
}

\caption{We show the performance differences between volume-rendering-based methods (InstantNGP and Ours) and rasterization-based method (3DGS).}
\label{fig:performance_rendering}
\end{figure}

We present more surface reconstruction results and extrapolation rendering results on the FAST-LIVO2 datasets, as shown in Fig~\ref{fig:fast_livo2_qual}.
We further dive into the performance differences between volume-rendering-based methods (InstantNGP and Ours) and rasterization-based methods (3DGS), as shown in Fig.~\ref{fig:performance_rendering}.
InstantNGP and Our methods show more reasonable rendering results and fewer artifacts overall, such as the ground in the down-left image.
While 3DGS shows powerful capability to capture high-frequency textures, such as the carved wall in the middle image.

\begin{figure*}[!t]
    \centering
    \subfigure{
        \centering
        \includegraphics[width=0.155\textwidth]{figures/sampling/ingp.jpeg}
    }
    \hspace{-11pt}
    \subfigure{
    \centering
    \includegraphics[width=0.155\textwidth]{figures/sampling/3dgs.jpeg}
    }
    \hspace{-11pt}
    \subfigure{
    \centering
    \includegraphics[width=0.155\textwidth]{figures/sampling/ours.jpg}
    }
    \vline
    \hspace{1pt}
    \subfigure{
    \centering
    \includegraphics[width=0.155\textwidth]{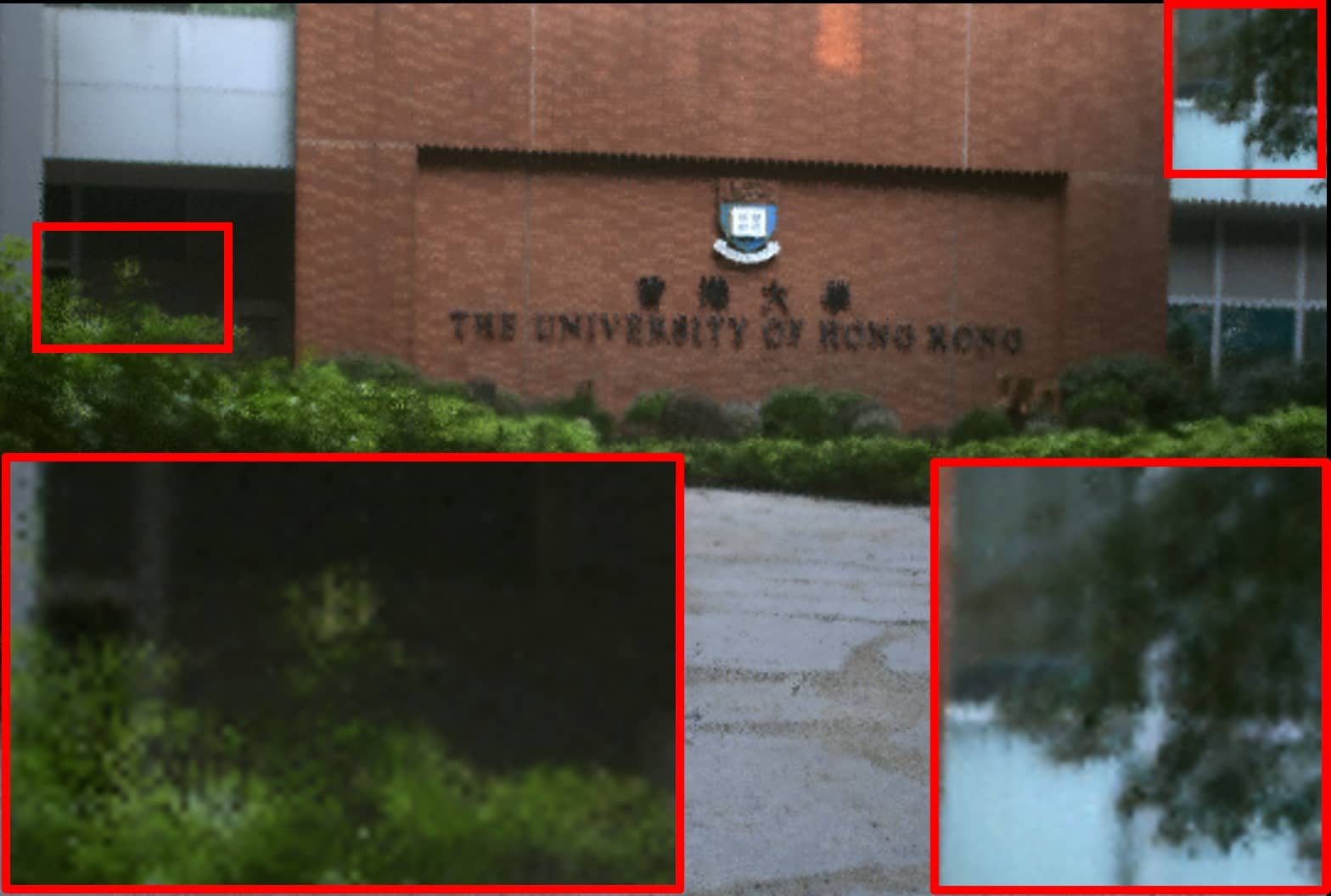}
    }
    \hspace{-11pt}
    \subfigure{
    \centering
    \includegraphics[width=0.155\textwidth]{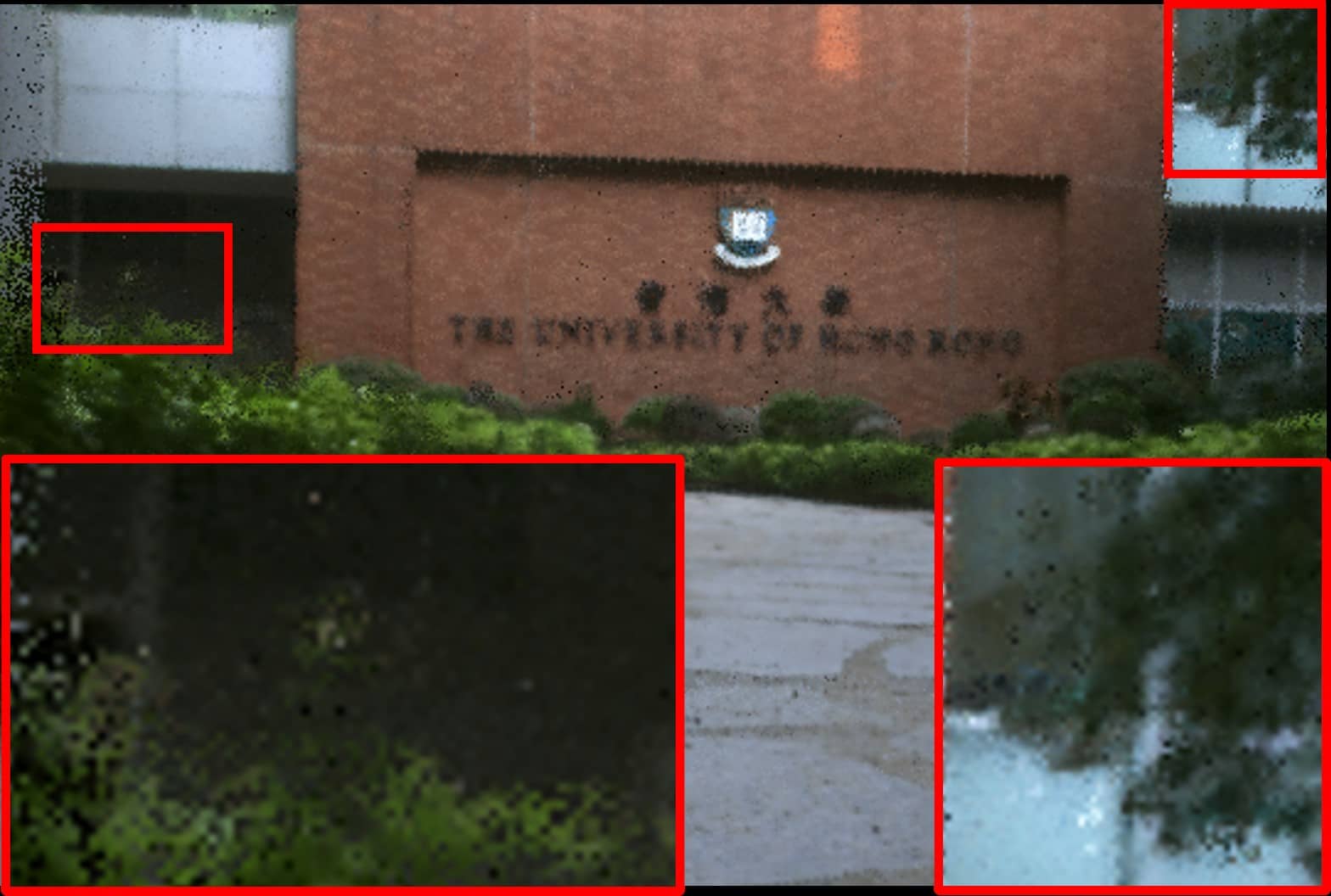}
    }
    \hspace{-11pt}
    \subfigure{
    \centering
    \includegraphics[width=0.155\textwidth]{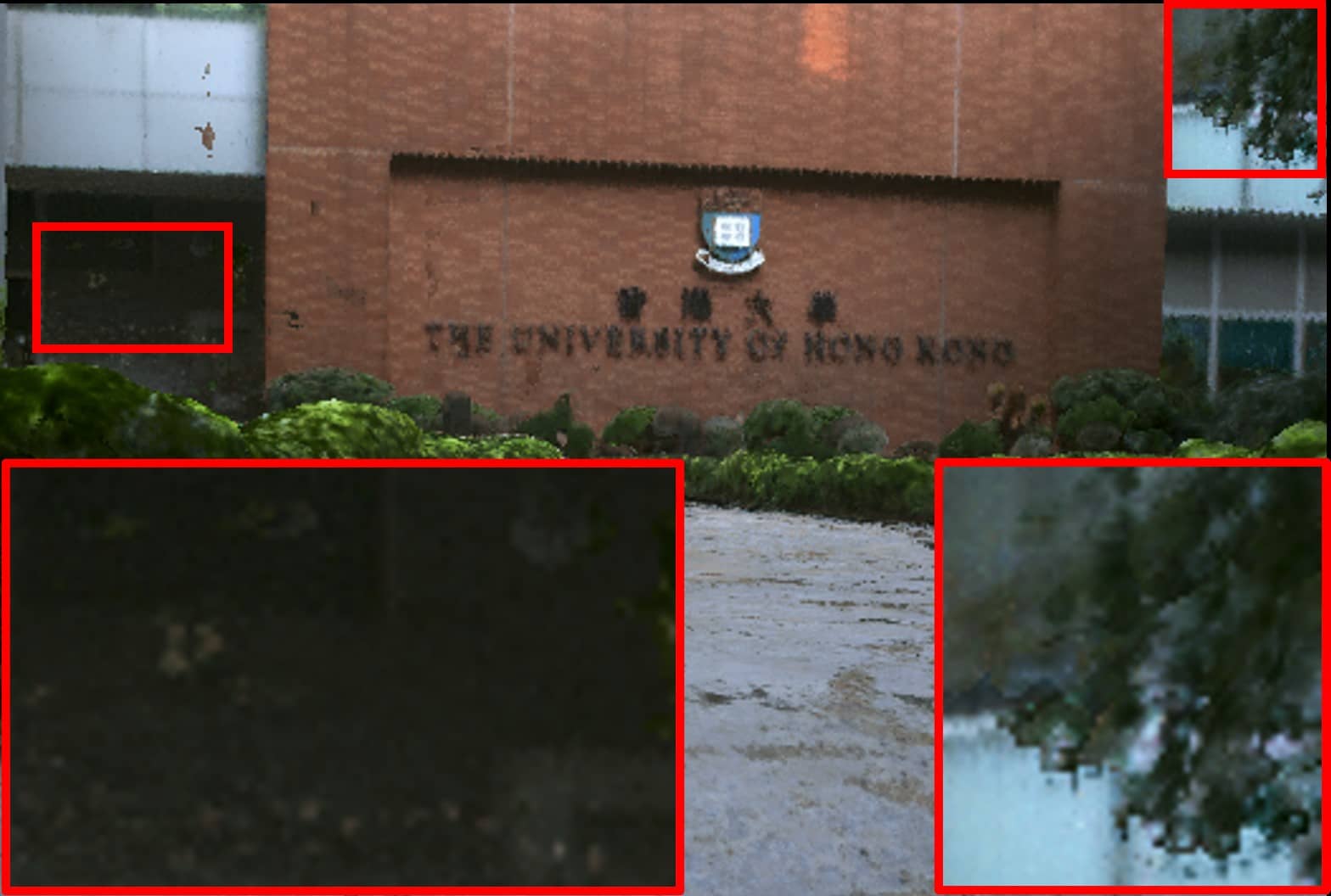}
    }

    \vspace{-8pt}
    
    \setcounter{subfigure}{0}
    \subfigure[InstantNGP]{
        \centering
        \includegraphics[width=0.155\textwidth]{figures/sampling/ingp_depth.jpeg}
    }
    \hspace{-11pt}
    \subfigure[3DGS$^{\dagger}$]{
    \centering
    \includegraphics[width=0.155\textwidth]{figures/sampling/3dgs_depth.jpeg}
    }
    \hspace{-11pt}
    \subfigure[Ours]{
    \centering
    \includegraphics[width=0.155\textwidth]{figures/sampling/ours_depth.jpeg}
    }
    \vline
    \hspace{1pt}
    \subfigure[Occ. Sampling]{
    \centering
    \includegraphics[width=0.155\textwidth]{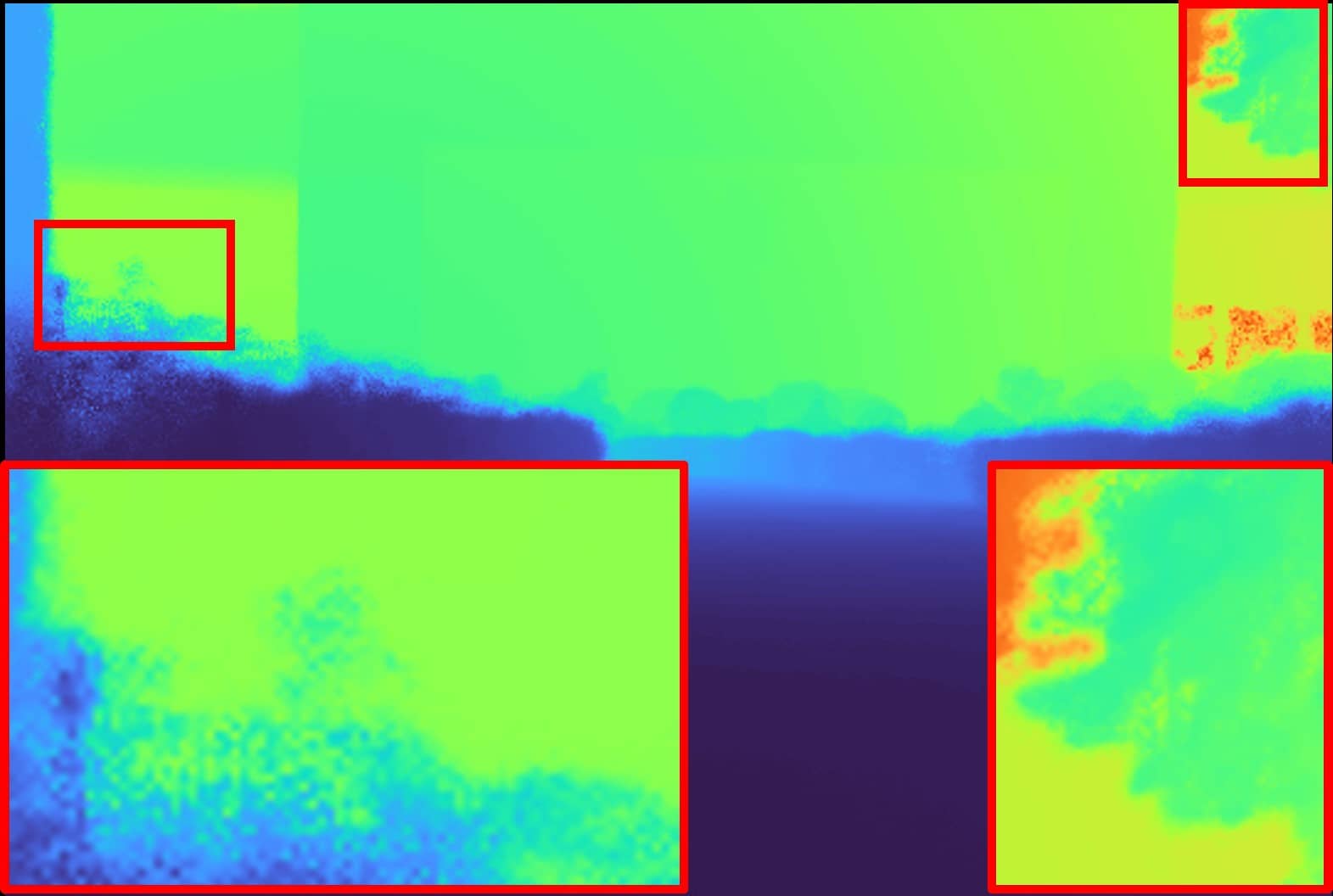}
    }
    \hspace{-11pt}
    \subfigure[PDF Sampling]{
    \centering
    \includegraphics[width=0.155\textwidth]{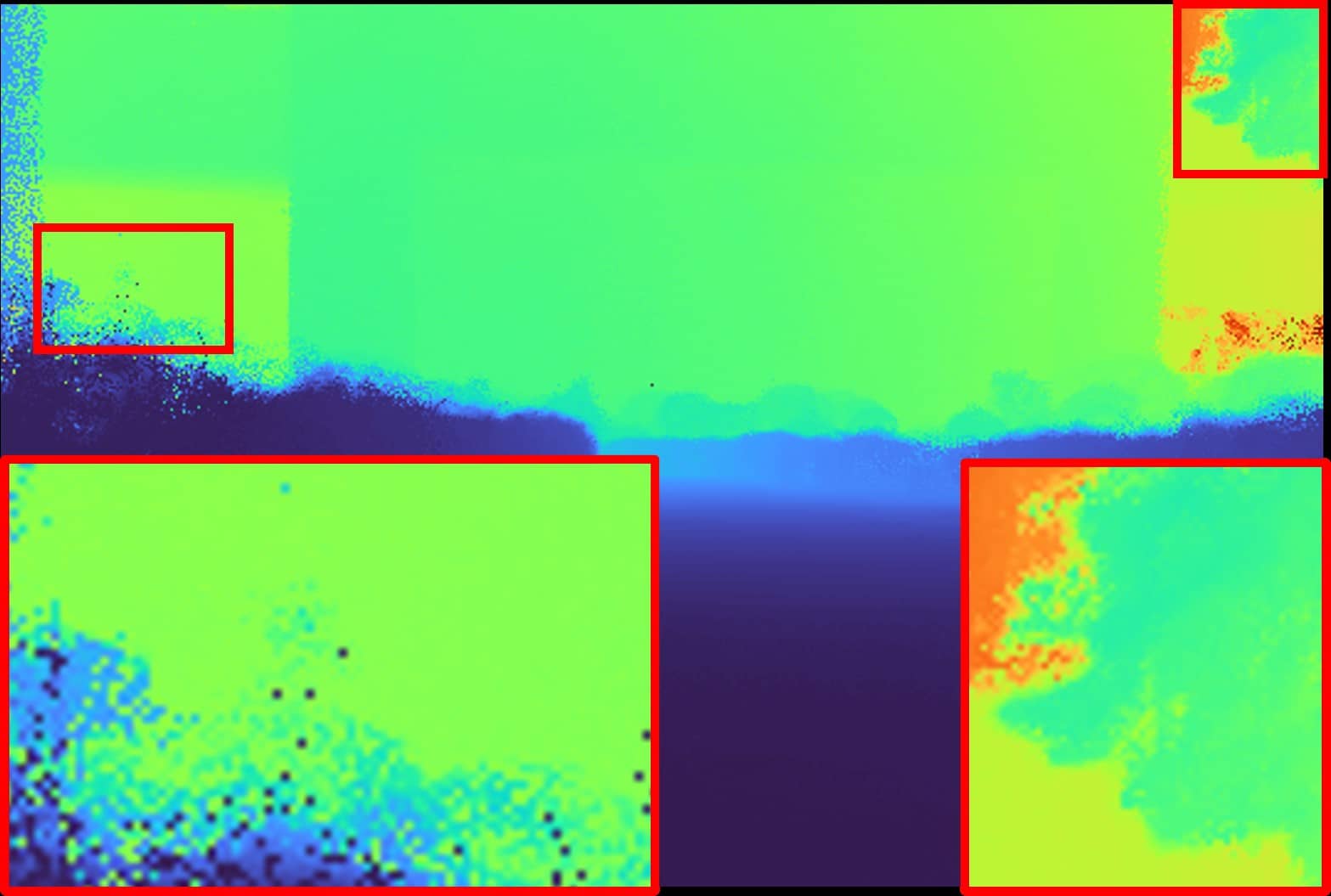}
    }
    \hspace{-11pt}
    \subfigure[Surface Rendering]{
    \centering
    \includegraphics[width=0.155\textwidth]{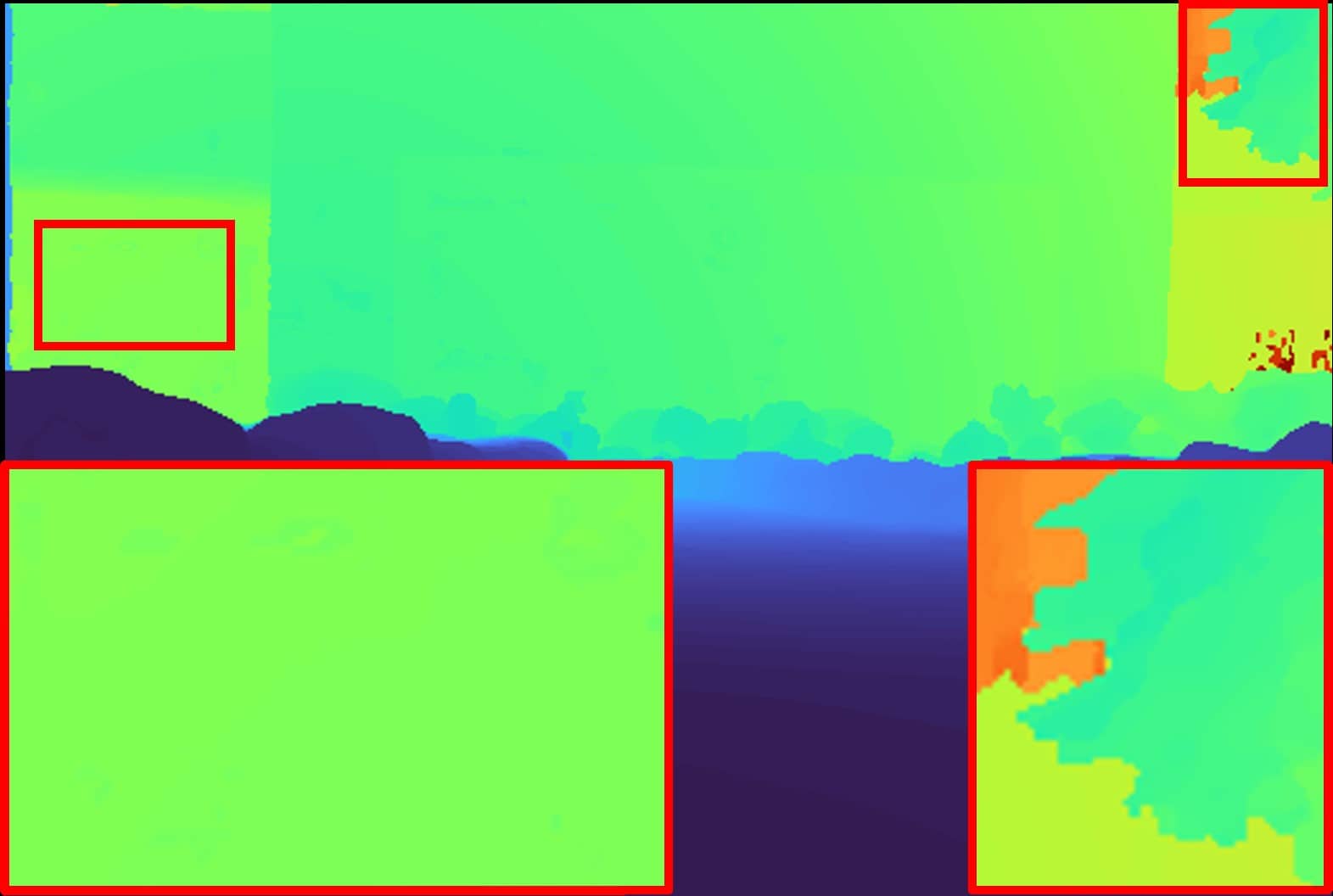}
    }
    
    \caption{
    (Left) Rendering results between different baselines on the FAST-LIVO dataset's Campus scene. 
    (Right) Rendering results with different sampling and rendering methods.}
    \label{fig:sampler_qual}
    \vspace{-8pt}
\end{figure*}

\subsection{Ablation Study}

\subsubsection{Spatial-varying scale}
\begin{figure}[!h]
    \centering
    \setcounter{subfigure}{0}
    \subfigure[$\beta = 5\times 10^{-2}$]{
        \label{fig:fix_scale_a}
      \centering
      \includegraphics[width=0.155\textwidth]{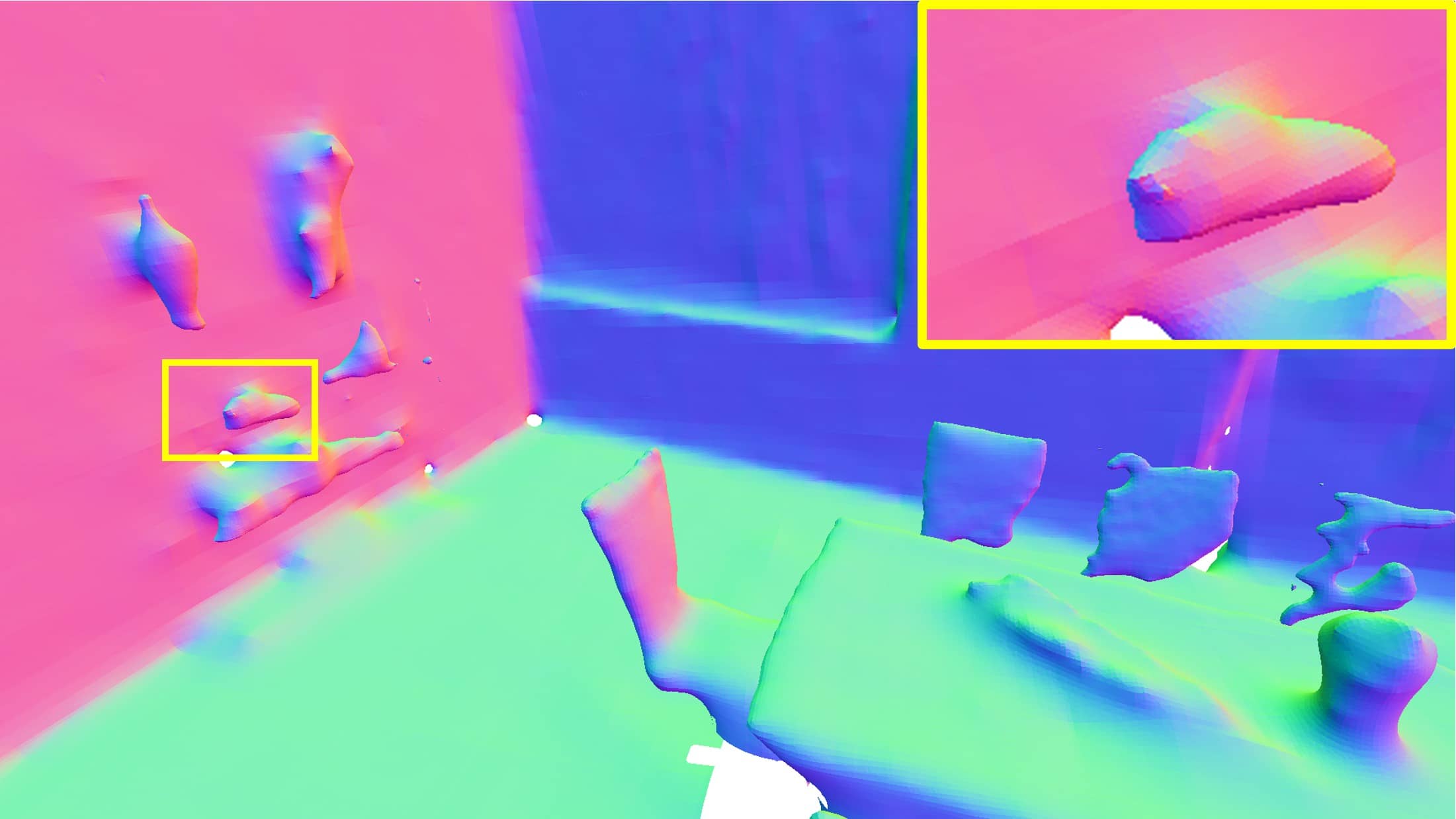}
  }
  \hspace{-12pt}
  \subfigure[$\beta = 10^{-4}$]{
      \centering
      \includegraphics[width=0.155\textwidth]{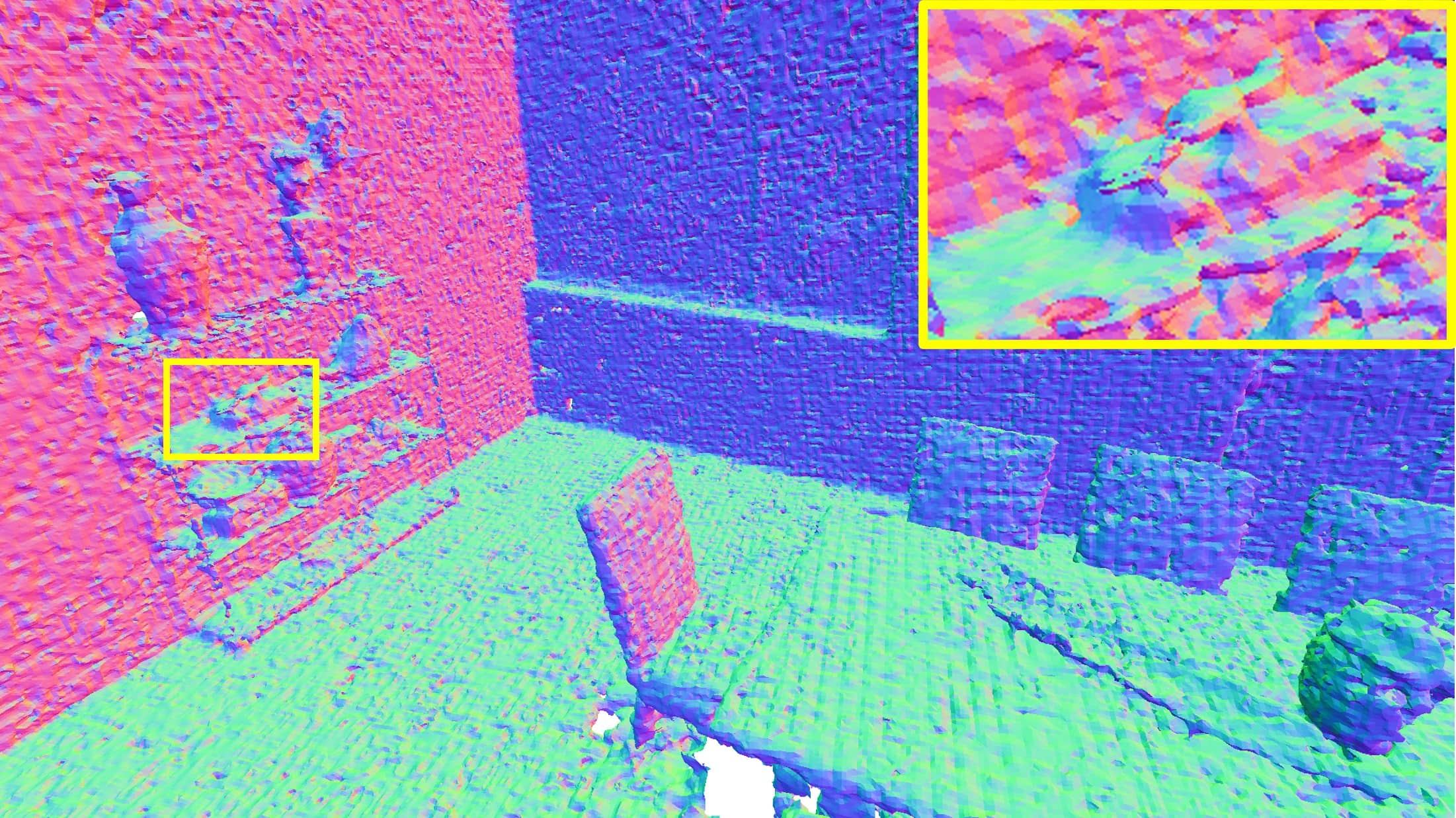}
  }
  \hspace{-12pt}
\subfigure[Spatial-varying ${\beta}$]{
    \label{fig:fix_scale_c}
      \centering
      \includegraphics[width=0.155\textwidth]{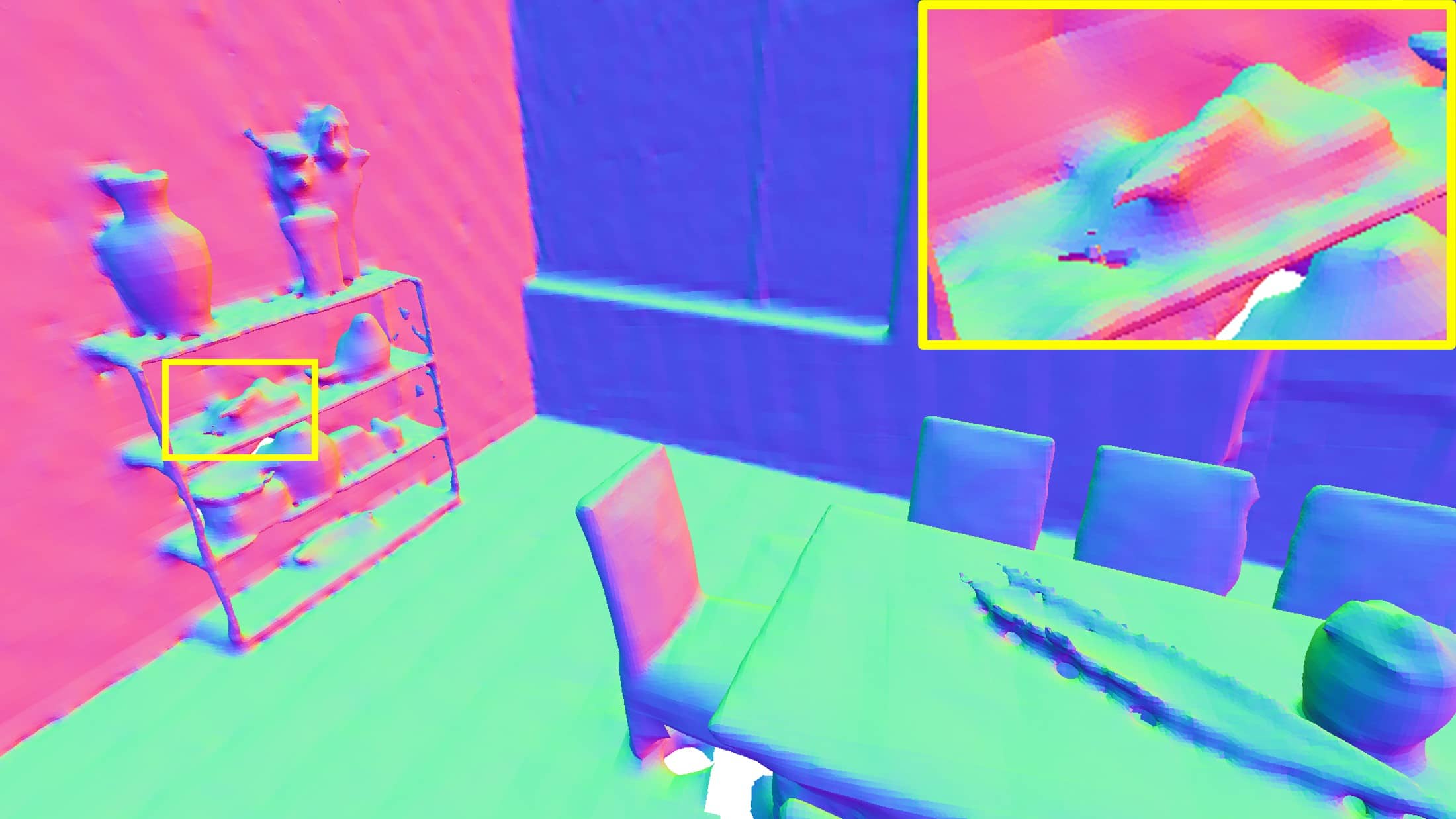}
  }

  \vspace{-4pt}
  
  \subfigure[Render Color]{
    \centering
    \includegraphics[width=0.155\textwidth]{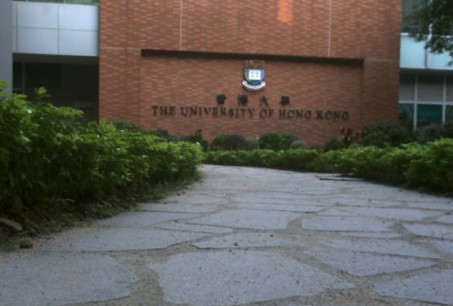}
}
\hspace{-12pt}
\subfigure[Render Scale $\beta$]{
    \centering
    \includegraphics[width=0.155\textwidth]{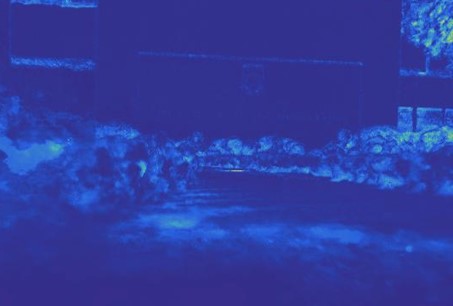}
}
\hspace{-12pt}
  \subfigure[Sample numer]{
    \centering
    \includegraphics[width=0.155\textwidth]{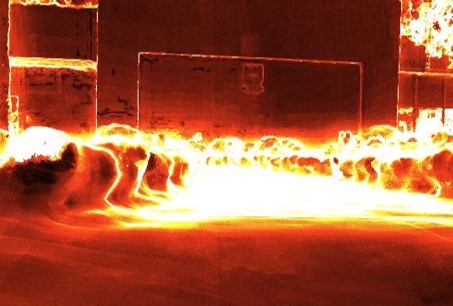}
}

\caption{In the first row, we show different scale settings in the Replica Room-2 scene. The average spatial-varying $\beta$ in the scene is $1.6\times 10^{-4}$. In the second row, we show the render scale (the lighter color corresponds to the bigger scale) and sample numer per ray (the lighter color corresponds to the bigger number) in Campus scenes.}
\label{fig:fix_scale}
\end{figure}

To demonstrate the necessity of spatial-varying scale, we conducted experiments on the Replica dataset's room-2 scene, where we applied $2cm$ normal noise to the ground truth depth.
As shown in Fig.~\ref{fig:fix_scale} (a-c), a large scale $\beta$ results in smoother surface reconstructions with a loss of details, while a small scale leads to overfitting and noisy surface reconstructions.
A spatial-varying scale is introduced to the neural distance field to adapt to the scene's various granularity and preserve levels of detail for objects of different scales.
In Fig.~\ref{fig:fix_scale} (d-e), the rendering scale that accumulates the scale along the rays, shows that rays traverse fuzzy geometries return larger scale $\beta$ indicating more uncertainty along these rays. 
Then the larger scale resulting in lower density (Eq.~3) makes the structure-aware sampling return more samples on these rays to meet the transmittance requirements.

\subsubsection{Structure-aware sampling}

To validate the proposed structure-aware sampling strategy, we compare the qualitative reconstruction results with occupancy sampling\cite{muller2022instant}, probabilistic density function (PDF) sampling\cite{mildenhall2021nerf} and surface rendering, as shown in Fig.~\ref{fig:sampler_qual} (Right).
The occupancy sampling takes uniform samples in every occupied grid.
The PDF sampling first takes uniform samples along the ray to obtains the cumulative distribution function, and then generates samples using inverse transform sampling.
Surface rendering renders the scene with the first intersecting surfaces' radiance.
The proposed structure-aware sampling strategy better adapts to the SDF prior and focuses more on the surface than previous sampling methods, avoiding the skipping of fuzzy geometries as seen in surface rendering for more accurate structure rendering.

\begin{figure*}[!t]
    \centering
    \subfigure{
        \centering
        \includegraphics[width=0.19\textwidth]{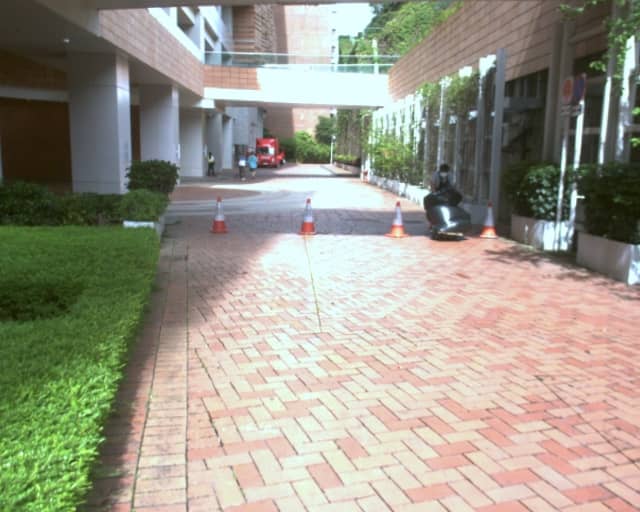}
    }
    \hspace{-11pt}
    \subfigure{
    \centering
    \includegraphics[width=0.19\textwidth]{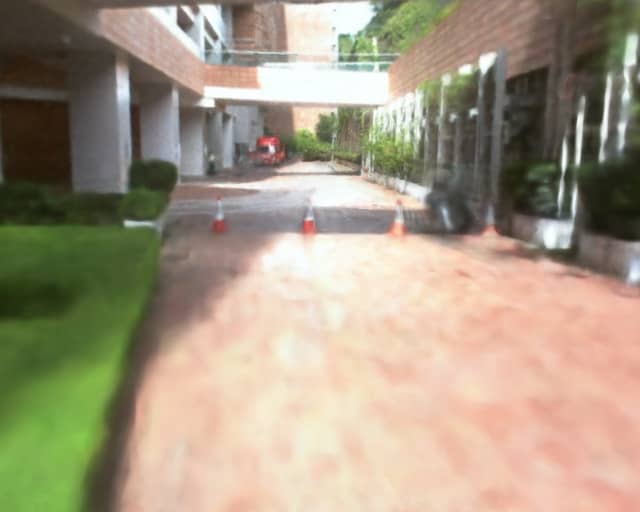}
    }
    \hspace{-11pt}
    \subfigure{
    \centering
    \includegraphics[width=0.19\textwidth]{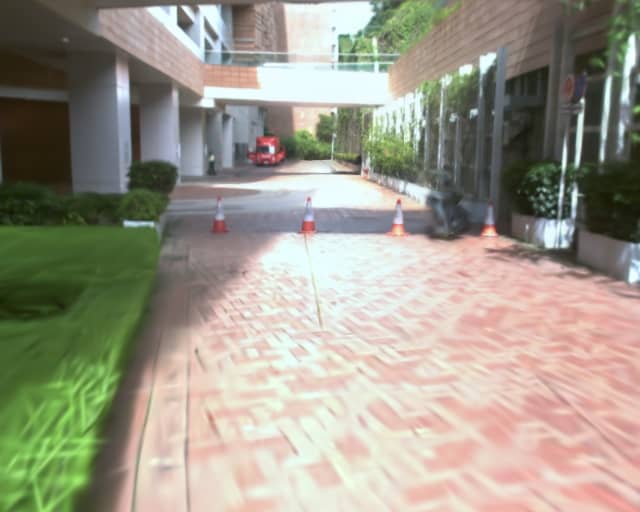}
    }
    \hspace{-11pt}
    \subfigure{
    \centering
    \includegraphics[width=0.19\textwidth]{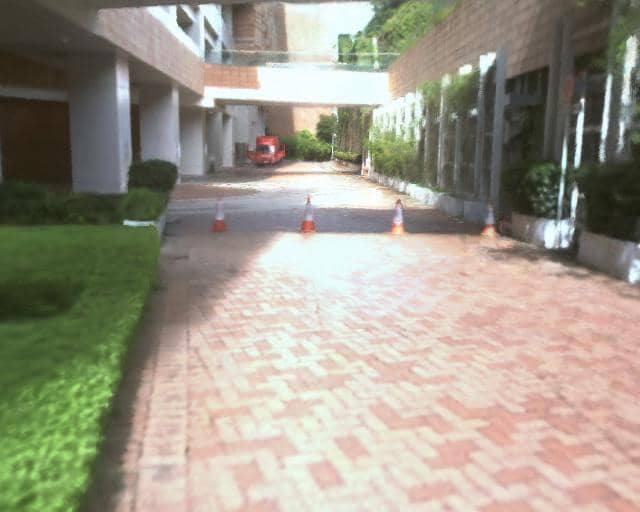}
    }
    \hspace{-11pt}
    \subfigure{
    \centering
    \includegraphics[width=0.19\textwidth]{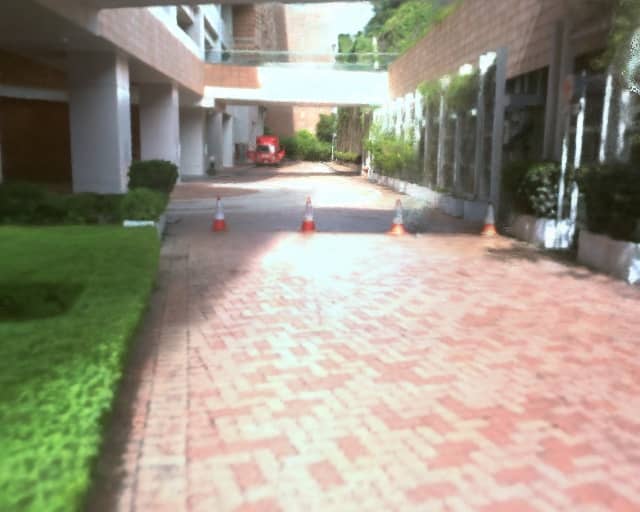}
    }

    \vspace{-8pt}
    
    \setcounter{subfigure}{0}
    \subfigure[Source Image]{
        \centering
        \includegraphics[width=0.19\textwidth]{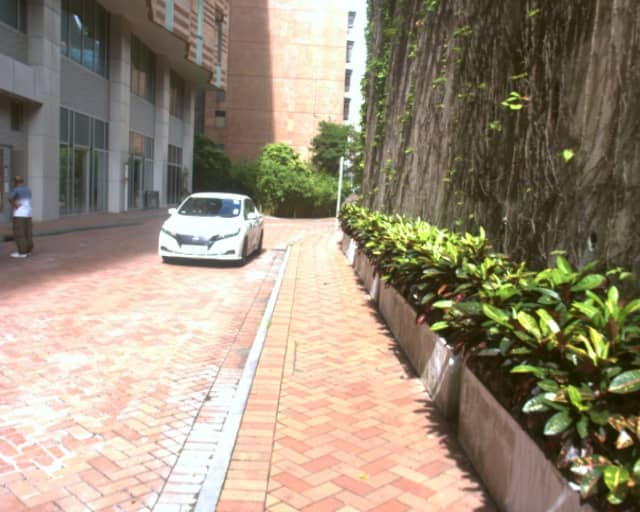}
    }
    \hspace{-11pt}
    \subfigure[InstantNGP]{
    \centering
    \includegraphics[width=0.19\textwidth]{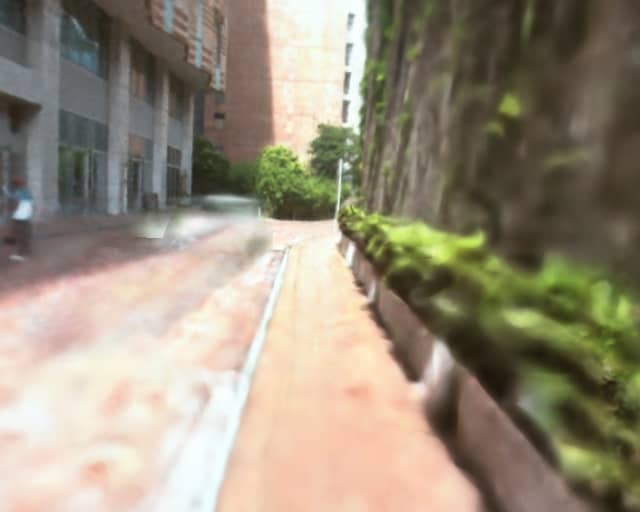}
    }
    \hspace{-11pt}
    \subfigure[3DGS$^{\dagger}$]{
    \centering
    \includegraphics[width=0.19\textwidth]{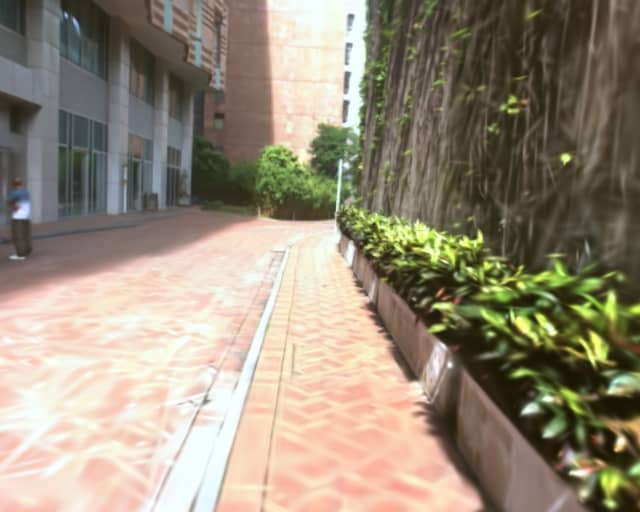}
    }
    \hspace{-11pt}
    \subfigure[Ours]{
    \centering
    \includegraphics[width=0.19\textwidth]{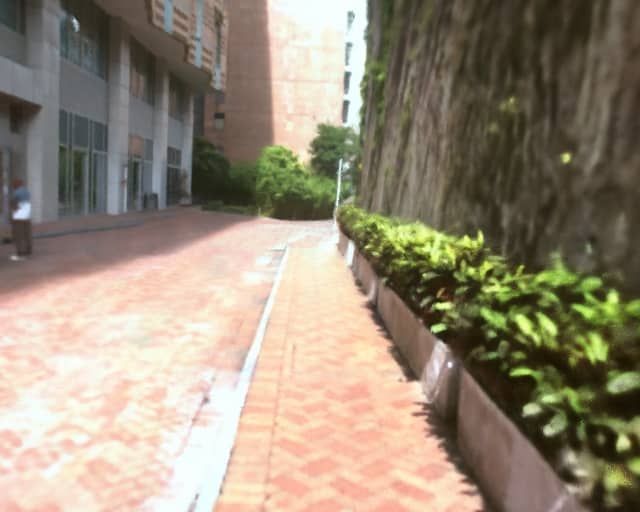}
    }
    \hspace{-11pt}
    \subfigure[wo outlier removal]{
    \centering
    \includegraphics[width=0.19\textwidth]{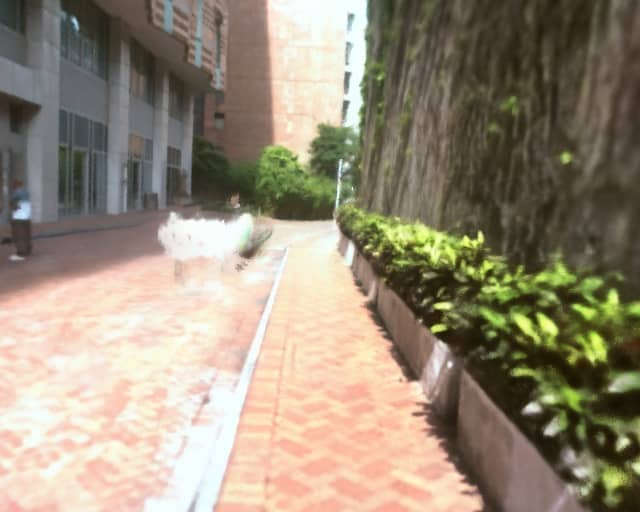}
    }
    
    \caption{
    Ablation study of outlier removal on the FAST-LIVO2 Drive dataset where dynamic objects (cart pusher and car) move in the scene.}
    \label{fig:outlier_removal}
\end{figure*}

\subsubsection{Outlier removal}

To validate the necessity of outlier removal (Sec.~\ref{sec:outlier_removal}) for real-world applications, we conducted experiments on the FAST-LIVO2 dataset's Drive scene, where dynamic objects (cart pusher and car) move in the scene.
As shown in Fig.~\ref{fig:outlier_removal}, the neural distance field with outlier removal can remove false surfaces caused by dynamic objects and regularize low density in the space traversed by the dynamic objects, and the more consistent background appearance filters out temporary dynamic objects in rendering.

\subsubsection{Directional embedding scheduler}

We propose to use a directional embedding scheduler (Sec.~\ref{sec:dir_scheduler}) to enhance the generalization of the neural radiance field in extrapolating views, which can be verified 
in Fig.~\ref{fig:fast_livo2_qual} (d). The rendering image with the directional embedding scheduler can better generalize the extrapolating views for more consistent color, especially in the free-view trajectory scenes, like Drive. Meanwhile, the directional embedding scheduler has little influence on the interpolation rendering results, as shown in Tab.~\ref{tab:fast_livo_quan} (wo dir. sched.).

\begin{table}
    \centering
    \caption{Quantitative results on the FAST-LIVO2 dataset.}
    \label{tab:fast_livo_quan}
    \resizebox{0.45\textwidth}{!}{
    \begin{tabular}{cccccccccc}
        \toprule
        \textbf{Metrics} & \textbf{Methods} & \textbf{Campus} & \textbf{Sculpture} & \textbf{Culture} & \textbf{Drive} & \textbf{Avg.} \\
        \midrule
  
        \multirow{4}*{SSIM$\uparrow$}  
        & {InstantNGP} 
        & 0.789 & 0.698 & 0.670 & 0.697 & 0.714
        \\
        
        & {3DGS$^{\dagger}$} 
        & \textbf{0.849} & \textbf{{0.769}} & \underline{{0.726}} & \textbf{0.778} & \textbf{0.780}
        \\

        & {Ours}
        & \underline{0.834} & \underline{0.729} & \textbf{0.727} & \underline{0.764} & \underline{0.764}
        \\
        \cmidrule{2-7}
        & {wo dir. sched.}
        & {0.833} & {0.726} & {{0.729}} & {0.767} & {0.764}
        \\
        \midrule
  
        \multirow{4}*{PSNR$\uparrow$}  
        & {InstantNGP} 
        & 28.880 & 22.356 & 21.563 & 24.145 & 24.236
        \\
  
        & {3DGS$^{\dagger}$} 
        & \textbf{31.310} & \textbf{{24.128}} & \underline{21.764} & \underline{25.837} & \underline{25.760}
        \\
        
        & {Ours}
        & \underline{30.681} & \underline{23.453} & \textbf{24.695} & \textbf{25.941} & \textbf{26.193}
        \\
        \cmidrule{2-7}
        & {wo dir. sched.}
        & {30.572} & {23.534} & {24.797} & {26.072} & {26.244}
        \\
        \midrule
  
        \multirow{4}*{LPIPS$\downarrow$}  
        
        & {InstantNGP} 
        & 0.255 & 0.376 & 0.428 & 0.416 & 0.369
        \\

        & {3DGS$^{\dagger}$} 
        & \textbf{0.182} & \textbf{{0.266}} & \underline{{{0.361}}} & \underline{0.296} & \textbf{0.276}
        \\
        & {Ours}
        & \underline{0.210} & \underline{0.321} & \textbf{0.350} & \textbf{0.293} & \underline{0.293}
        \\
        \cmidrule{2-7}
        & {wo dir. sched.}
        & {0.213} & {0.330} & {{0.351}} & {0.293} & {0.297}
        \\
        \bottomrule
    \end{tabular}
    }
\end{table}

\subsubsection{Visual-aided structure}

\begin{figure}
    \centering

  \subfigure[Point cloud]{
    \centering
    \includegraphics[width=0.115\textwidth]{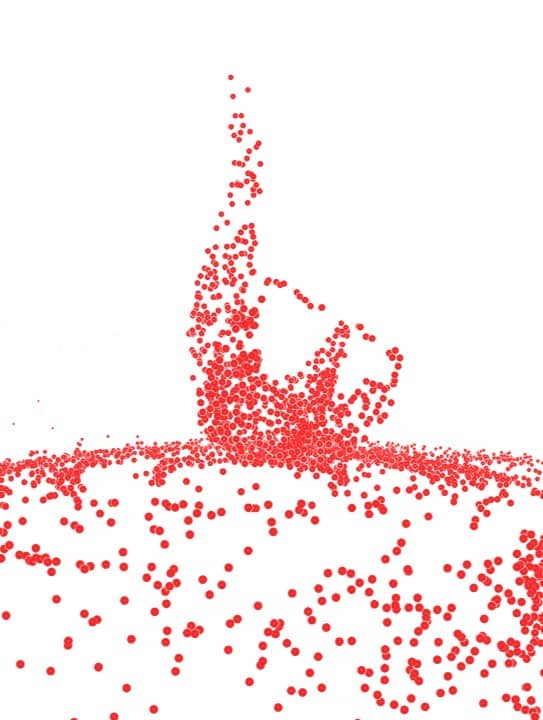}
    }
    \hspace{-12pt}
    \subfigure[wo visual]{
        \centering
        \includegraphics[width=0.115\textwidth]{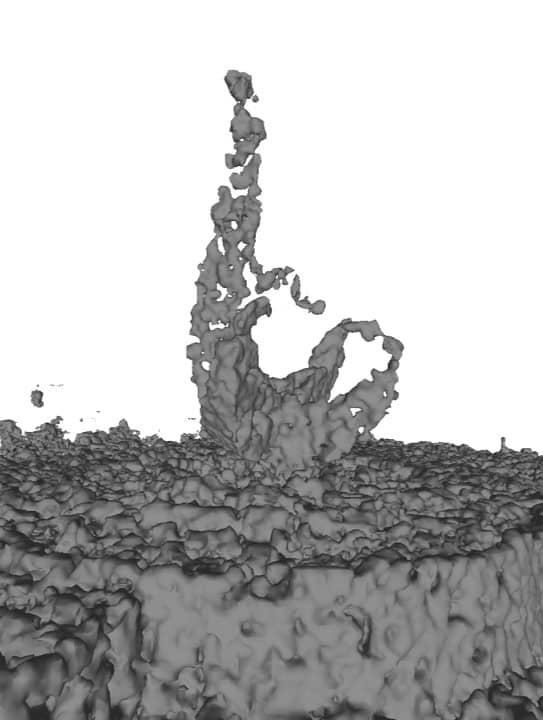}
    }
    \hspace{-12pt}
    \subfigure[Ours]{
      \centering
      \includegraphics[width=0.115\textwidth]{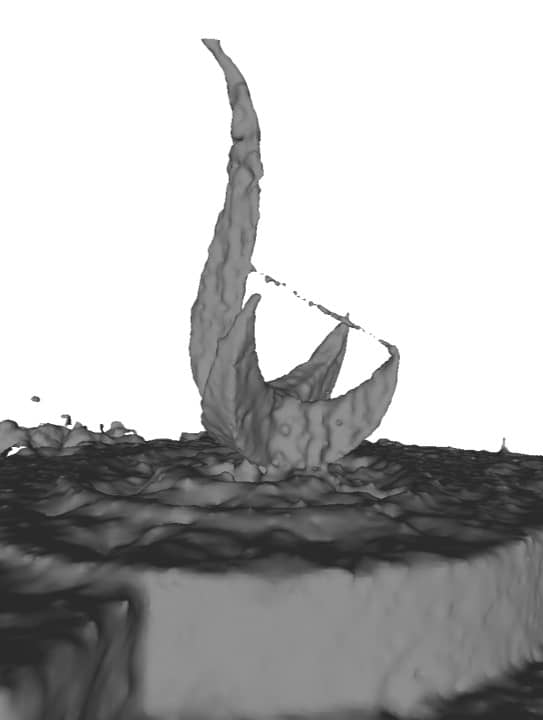}
  }
  \hspace{-12pt}
  \subfigure[Ours (Dense)]{
    \centering
    \includegraphics[width=0.115\textwidth]{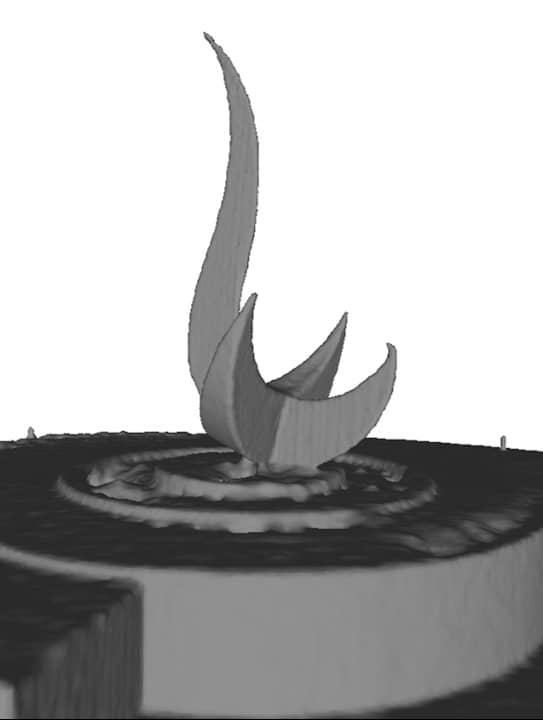}
}



\caption{We show the Sculpture scene's reconstruction results from downsampled point clouds (a) without (b) and with (c) visual supervision. And (d) shows the reconstruction result from dense point clouds.}
\label{fig:visual_completion}
\vspace{-16pt}
\end{figure}

To validate the influence of image measurements on surface reconstructions, we first downsampled the point cloud in the Sculpture scene and compared the reconstruction results with and without visual supervision $\mathcal{L}_{rgb}$.
As shown in Fig.~\ref{fig:visual_completion}, the neural radiance fields can complete the structure from sparse point clouds and avoid overfitting in scenes.

\subsection{Training and Rendering Efficiency}

So far the bottleneck of the efficiency is greatly impeded by the unbalanced sphere tracing's steps, once there is a ray that does not converge to the surface, the other converged rays need to wait until it is finished or reach the maximum steps.
The training time for a Replica scene takes about 20 minutes and the rendering time for a 1200x680 image takes about 0.07 seconds (13Hz).
In the future, it could be solved by conducting a ray-oriented rendering for ray rendering instead of batch rendering.

{
\balance
}

%% file: main.bbl
\begin{thebibliography}{10}
\providecommand{\url}[1]{#1}
\csname url@rmstyle\endcsname
\providecommand{\newblock}{\relax}
\providecommand{\bibinfo}[2]{#2}
\providecommand\BIBentrySTDinterwordspacing{\spaceskip=0pt\relax}
\providecommand\BIBentryALTinterwordstretchfactor{4}
\providecommand\BIBentryALTinterwordspacing{\spaceskip=\fontdimen2\font plus
\BIBentryALTinterwordstretchfactor\fontdimen3\font minus \fontdimen4\font\relax}
\providecommand\BIBforeignlanguage[2]{{%
\expandafter\ifx\csname l@#1\endcsname\relax
\typeout{** WARNING: IEEEtran.bst: No hyphenation pattern has been}%
\typeout{** loaded for the language `#1'. Using the pattern for}%
\typeout{** the default language instead.}%
\else
\language=\csname l@#1\endcsname
\fi
#2}}

\bibitem{wang2022survey}
Y.~Wang, Z.~Su, N.~Zhang, R.~Xing, D.~Liu, T.~H. Luan, and X.~Shen, ``A survey on metaverse: Fundamentals, security, and privacy,'' \emph{IEEE Communications Surveys \& Tutorials}, vol.~25, no.~1, pp. 319--352, 2022.

\bibitem{makoviychuk2021isaac}
V.~Makoviychuk, L.~Wawrzyniak, Y.~Guo, M.~Lu, K.~Storey, M.~Macklin, D.~Hoeller, N.~Rudin, A.~Allshire, A.~Handa, \emph{et~al.}, ``Isaac gym: High performance gpu-based physics simulation for robot learning,'' \emph{arXiv preprint arXiv:2108.10470}, 2021.

\bibitem{lvisam2021shan}
T.~Shan, B.~Englot, C.~Ratti, and R.~Daniela, ``Lvi-sam: Tightly-coupled lidar-visual-inertial odometry via smoothing and mapping,'' in \emph{IEEE International Conference on Robotics and Automation (ICRA)}.\hskip 1em plus 0.5em minus 0.4em\relax IEEE, 2021, pp. 5692--5698.

\bibitem{lin2022r}
J.~Lin and F.~Zhang, ``R 3 live: A robust, real-time, rgb-colored, lidar-inertial-visual tightly-coupled state estimation and mapping package,'' in \emph{2022 International Conference on Robotics and Automation (ICRA)}.\hskip 1em plus 0.5em minus 0.4em\relax IEEE, 2022, pp. 10\,672--10\,678.

\bibitem{zheng2024fast}
C.~Zheng, W.~Xu, Z.~Zou, T.~Hua, C.~Yuan, D.~He, B.~Zhou, Z.~Liu, J.~Lin, F.~Zhu, \emph{et~al.}, ``Fast-livo2: Fast, direct lidar-inertial-visual odometry,'' \emph{arXiv preprint arXiv:2408.14035}, 2024.

\bibitem{straub2019replica}
J.~Straub, T.~Whelan, L.~Ma, Y.~Chen, E.~Wijmans, S.~Green, J.~J. Engel, R.~Mur-Artal, C.~Ren, S.~Verma, \emph{et~al.}, ``The replica dataset: A digital replica of indoor spaces,'' \emph{arXiv preprint arXiv:1906.05797}, 2019.

\bibitem{lin2023immesh}
J.~Lin, C.~Yuan, Y.~Cai, H.~Li, Y.~Ren, Y.~Zou, X.~Hong, and F.~Zhang, ``Immesh: An immediate lidar localization and meshing framework,'' \emph{IEEE Transactions on Robotics}, 2023.

\bibitem{vizzo2021poisson}
I.~Vizzo, X.~Chen, N.~Chebrolu, J.~Behley, and C.~Stachniss, ``Poisson surface reconstruction for lidar odometry and mapping,'' in \emph{2021 IEEE International Conference on Robotics and Automation (ICRA)}.\hskip 1em plus 0.5em minus 0.4em\relax IEEE, 2021, pp. 5624--5630.

\bibitem{oleynikova2017voxblox}
H.~Oleynikova, Z.~Taylor, M.~Fehr, R.~Siegwart, and J.~Nieto, ``Voxblox: Incremental 3d euclidean signed distance fields for on-board mav planning,'' in \emph{2017 IEEE/RSJ International Conference on Intelligent Robots and Systems (IROS)}.\hskip 1em plus 0.5em minus 0.4em\relax IEEE, 2017, pp. 1366--1373.

\bibitem{vizzo2022vdbfusion}
I.~Vizzo, T.~Guadagnino, J.~Behley, and C.~Stachniss, ``Vdbfusion: Flexible and efficient tsdf integration of range sensor data,'' \emph{Sensors}, vol.~22, no.~3, p. 1296, 2022.

\bibitem{chibane2020neural}
J.~Chibane, G.~Pons-Moll, \emph{et~al.}, ``Neural unsigned distance fields for implicit function learning,'' \emph{Advances in Neural Information Processing Systems}, vol.~33, pp. 21\,638--21\,652, 2020.

\bibitem{ortiz2022isdf}
J.~Ortiz, A.~Clegg, J.~Dong, E.~Sucar, D.~Novotny, M.~Zollhoefer, and M.~Mukadam, ``isdf: Real-time neural signed distance fields for robot perception,'' \emph{arXiv preprint arXiv:2204.02296}, 2022.

\bibitem{zhong2022shine}
X.~Zhong, Y.~Pan, J.~Behley, and C.~Stachniss, ``Shine-mapping: Large-scale 3d mapping using sparse hierarchical implicit neural representations,'' \emph{arXiv preprint arXiv:2210.02299}, 2022.

\bibitem{gropp2020implicit}
A.~Gropp, L.~Yariv, N.~Haim, M.~Atzmon, and Y.~Lipman, ``Implicit geometric regularization for learning shapes,'' in \emph{Proceedings of the 37th International Conference on Machine Learning}, 2020, pp. 3789--3799.

\bibitem{wang2023adaptive}
Z.~Wang, T.~Shen, M.~Nimier-David, N.~Sharp, J.~Gao, A.~Keller, S.~Fidler, T.~M{\"u}ller, and Z.~Gojcic, ``Adaptive shells for efficient neural radiance field rendering,'' \emph{ACM Transactions on Graphics (TOG)}, vol.~42, no.~6, pp. 1--15, 2023.

\bibitem{yariv2021volume}
L.~Yariv, J.~Gu, Y.~Kasten, and Y.~Lipman, ``Volume rendering of neural implicit surfaces,'' \emph{Advances in Neural Information Processing Systems}, vol.~34, pp. 4805--4815, 2021.

\bibitem{wang2021neus}
P.~Wang, L.~Liu, Y.~Liu, C.~Theobalt, T.~Komura, and W.~Wang, ``Neus: Learning neural implicit surfaces by volume rendering for multi-view reconstruction,'' \emph{Advances in Neural Information Processing Systems}, vol.~34, pp. 27\,171--27\,183, 2021.

\bibitem{mildenhall2021nerf}
B.~Mildenhall, P.~P. Srinivasan, M.~Tancik, J.~T. Barron, R.~Ramamoorthi, and R.~Ng, ``Nerf: Representing scenes as neural radiance fields for view synthesis,'' \emph{Communications of the ACM}, vol.~65, no.~1, pp. 99--106, 2021.

\bibitem{oechsle2021unisurf}
M.~Oechsle, S.~Peng, and A.~Geiger, ``Unisurf: Unifying neural implicit surfaces and radiance fields for multi-view reconstruction,'' in \emph{Proceedings of the IEEE/CVF International Conference on Computer Vision}, 2021, pp. 5589--5599.

\bibitem{kazhdan2006poisson}
M.~Kazhdan, M.~Bolitho, and H.~Hoppe, ``Poisson surface reconstruction,'' in \emph{Proceedings of the fourth Eurographics symposium on Geometry processing}, vol.~7, no.~4, 2006.

\bibitem{liu2023towards}
J.~Liu and H.~Chen, ``Towards real-time scalable dense mapping using robot-centric implicit representation,'' in \emph{IEEE International Conference on Robotics and Automation (ICRA)}, 2024.

\bibitem{deng2022depth}
K.~Deng, A.~Liu, J.-Y. Zhu, and D.~Ramanan, ``Depth-supervised nerf: Fewer views and faster training for free,'' in \emph{Proceedings of the IEEE/CVF Conference on Computer Vision and Pattern Recognition}, 2022, pp. 12\,882--12\,891.

\bibitem{sucar2021imap}
E.~Sucar, S.~Liu, J.~Ortiz, and A.~J. Davison, ``imap: Implicit mapping and positioning in real-time,'' in \emph{Proceedings of the IEEE/CVF International Conference on Computer Vision}, 2021, pp. 6229--6238.

\bibitem{jiang2023h}
C.~Jiang, H.~Zhang, P.~Liu, Z.~Yu, H.~Cheng, B.~Zhou, and S.~Shen, ``H $ \_ $\{$2$\}$ $-mapping: Real-time dense mapping using hierarchical hybrid representation,'' \emph{IEEE Robotics and Automation Letters}, 2023.

\bibitem{ziyang2023snerf}
Z.~Xie, J.~Zhang, W.~Li, F.~Zhang, and L.~Zhang, ``S-nerf: Neural radiance fields for street views,'' in \emph{International Conference on Learning Representations (ICLR)}, 2023.

\bibitem{tao2024silvr}
Y.~Tao, Y.~Bhalgat, L.~F.~T. Fu, M.~Mattamala, N.~Chebrolu, and M.~Fallon, ``Silvr: Scalable lidar-visual reconstruction with neural radiance fields for robotic inspection,'' in \emph{IEEE International Conference on Robotics and Automation (ICRA)}, 2024.

\bibitem{kerbl20233d}
B.~Kerbl, G.~Kopanas, T.~Leimk{\"u}hler, and G.~Drettakis, ``3d gaussian splatting for real-time radiance field rendering,'' \emph{ACM Transactions on Graphics}, vol.~42, no.~4, pp. 1--14, 2023.

\bibitem{GSSLAM2024}
H.~Matsuki, R.~Murai, P.~H.~J. Kelly, and A.~J. Davison, ``{G}aussian {S}platting {SLAM},'' in \emph{Proceedings of the IEEE/CVF Conference on Computer Vision and Pattern Recognition}, 2024.

\bibitem{hong2024liv}
S.~Hong, J.~He, X.~Zheng, H.~Wang, H.~Fang, K.~Liu, C.~Zheng, and S.~Shen, ``Liv-gaussmap: Lidar-inertial-visual fusion for real-time 3d radiance field map rendering,'' \emph{arXiv preprint arXiv:2401.14857}, 2024.

\bibitem{ren2023rog}
Y.~Ren, Y.~Cai, F.~Zhu, S.~Liang, and F.~Zhang, ``Rog-map: An efficient robocentric occupancy grid map for large-scene and high-resolution lidar-based motion planning,'' \emph{arXiv preprint arXiv:2302.14819}, 2023.

\bibitem{muller2022instant}
T.~M{\"u}ller, A.~Evans, C.~Schied, and A.~Keller, ``Instant neural graphics primitives with a multiresolution hash encoding,'' \emph{arXiv preprint arXiv:2201.05989}, 2022.

\bibitem{verbin2022ref}
D.~Verbin, P.~Hedman, B.~Mildenhall, T.~Zickler, J.~T. Barron, and P.~P. Srinivasan, ``Ref-nerf: Structured view-dependent appearance for neural radiance fields,'' in \emph{2022 IEEE/CVF Conference on Computer Vision and Pattern Recognition (CVPR)}.\hskip 1em plus 0.5em minus 0.4em\relax IEEE, 2022, pp. 5481--5490.

\bibitem{wang2022hf}
Y.~Wang, I.~Skorokhodov, and P.~Wonka, ``Hf-neus: Improved surface reconstruction using high-frequency details,'' \emph{Advances in Neural Information Processing Systems}, vol.~35, pp. 1966--1978, 2022.

\bibitem{ban2023automatic}
R.~B{\'a}n and G.~Valasek, ``Automatic step size relaxation in sphere tracing,'' 2023.

\bibitem{yang2023steik}
H.~Yang, Y.~Sun, G.~Sundaramoorthi, and A.~Yezzi, ``Steik: Stabilizing the optimization of neural signed distance functions and finer shape representation,'' in \emph{Thirty-seventh Conference on Neural Information Processing Systems}, 2023.

\bibitem{lorensen1987marching}
W.~E. Lorensen and H.~E. Cline, ``Marching cubes: A high resolution 3d surface construction algorithm,'' \emph{ACM Siggraph Computer Graphics}, vol.~21, no.~4, pp. 163--169, 1987.

\bibitem{li2023neuralangelo}
Z.~Li, T.~M{\"u}ller, A.~Evans, R.~H. Taylor, M.~Unberath, M.-Y. Liu, and C.-H. Lin, ``Neuralangelo: High-fidelity neural surface reconstruction,'' in \emph{Proceedings of the IEEE/CVF Conference on Computer Vision and Pattern Recognition}, 2023, pp. 8456--8465.

\bibitem{barron2022mip}
J.~T. Barron, B.~Mildenhall, D.~Verbin, P.~P. Srinivasan, and P.~Hedman, ``Mip-nerf 360: Unbounded anti-aliased neural radiance fields,'' in \emph{Proceedings of the IEEE/CVF Conference on Computer Vision and Pattern Recognition}, 2022, pp. 5470--5479.

\end{thebibliography}
